\pdfoutput=1
\documentclass[12pt]{article}

\usepackage[margin=1in]{geometry}
\usepackage[T1]{fontenc}
\usepackage[utf8]{inputenc}
\usepackage{lmodern}
\usepackage{microtype}
\usepackage{amsmath,amssymb,amsthm,mathtools}
\usepackage{bm}
\usepackage{booktabs}
\usepackage{enumitem}
\usepackage{xcolor}
\usepackage{graphicx}
\usepackage[font=footnotesize]{caption}
\usepackage{tikz}
\usepackage{float}
\usepackage{placeins}
\usepackage[numbers,sort&compress]{natbib}
\usepackage[colorlinks=true,allcolors=blue]{hyperref}
\usepackage[nameinlink,noabbrev]{cleveref}
\usepackage{setspace}

\newcommand{\R}{\mathbb{R}}
\newcommand{\E}{\mathbb{E}}
\newcommand{\Pp}{\mathbb{P}}

\newcommand{\norm}[1]{\left\lVert #1\right\rVert}

\newtheorem{theorem}{Theorem}

\newtheorem{lemma}[theorem]{Lemma}
\newtheorem{corollary}[theorem]{Corollary}
\theoremstyle{definition}
\newtheorem{definition}{Definition}
\newtheorem{remark}{Remark}
\newtheorem{assumption}{Assumption}

\vspace{-2em}
\title{\bf In-Context Estimation and Prediction for Parametric ODEs}

\author{
Abdullah Ahmed\textsuperscript{$\ast$}
and
Yun Yang\textsuperscript{$\dagger$}
}

\date{\small 
Department of Mathematics, University of Maryland, College Park
\\
\textsuperscript{$\ast$}\texttt{abdullah@umd.edu}
\qquad
\textsuperscript{$\dagger$}\texttt{yy84@umd.edu}
}

\setlist{itemsep=2pt,topsep=4pt}
\newcommand{\appendixentry}[2]{\noindent\hyperref[#1]{\textbf{\ref*{#1}\quad #2}}\dotfill\hyperref[#1]{\pageref*{#1}}\par}
\DeclareMathOperator*{\argmin}{arg\,min}
\newcommand{\cellstack}[1]{\begin{tabular}[t]{@{}l@{}}#1\end{tabular}}

\begin{document}
\maketitle
\begin{abstract}
Transformers have demonstrated the ability to perform in-context learning, that is, to adapt to new tasks from a few examples without updating their parameters. Existing theory has largely focused on emulating gradient-based algorithms for convex statistical problems or establishing general function approximation guarantees without exploiting the underlying statistical structure. We study in-context parameter estimation and trajectory prediction for parametric ordinary differential equations (ODEs), where estimation is generally nonconvex. Under suitable identifiability and regularity conditions, we construct a transformer that estimates unknown system parameters from noisy trajectory observations through a two-stage approximate maximum likelihood procedure: a coarse search provides an initialization within a locally strongly convex region, followed by gradient descent with local linear contraction. For prediction, we compose this estimator with a neural approximation of the ODE solution map. We further show that, through parameter sharing, a single transformer can perform both parameter estimation and trajectory prediction simultaneously. For these constructions, we establish nonasymptotic guarantees and finite-pretraining oracle inequalities that separate errors due to finite pretraining data, finite context length, and finite-sized transformer approximation. Numerical experiments on linear and nonlinear ODEs further show that pretrained transformers can benefit from longer inference contexts without updating their weights.
\end{abstract}


\setstretch{1.25}
\setlength{\abovedisplayskip}{8pt plus 2pt minus 2pt}
\setlength{\belowdisplayskip}{8pt plus 2pt minus 2pt}
\setlength{\abovedisplayshortskip}{4pt plus 2pt minus 1pt}
\setlength{\belowdisplayshortskip}{6pt plus 2pt minus 2pt}

\section{Introduction}
\label{sec:introduction}

Ordinary differential equations (ODEs) are widely used to model dynamical systems in science and engineering~\cite{strogatz2015nonlinear,hirsch2013differential}. In many applications, the form of the governing equation is known, often derived from physical laws, while the initial conditions and system-specific parameters, such as physical coefficients, reaction rates, or environmental parameters, vary across systems or experimental settings. This naturally gives rise to families of ODEs indexed by these unknown quantities, commonly referred to as \emph{parametric ODEs}. Given a limited number of noisy observations from a new system, the goal is to estimate its unknown parameters and predict its state at unobserved or future times. Classical approaches estimate the parameters from the observations and then solve the corresponding ODE numerically for prediction~\cite{bates1988,ramsay2007,brunel2008parameter}. Although statistically and physically interpretable, these approaches can be computationally expensive when applied to many related systems, as parameter estimation generally involves solving a nonconvex inverse problem.

\emph{In-context learning} (ICL) is a learning paradigm in which a trained machine-learning model adapts to new tasks using examples or other task-specific information provided as input, without updating its parameters~\cite{brown2020,garg2022}. For statistical inference, ICL can be formulated as a \emph{meta-learning} problem, where an algorithm is learned from data across related tasks to map observations from a new task to an estimate or prediction of the quantity of interest. By meta-learning this mapping in advance, ICL offers a way to amortize the computational cost of inference across tasks, reducing the need for repeated task-specific optimization. The success of ICL in large language models~\cite{brown2020} has motivated growing interest in developing foundation models for scientific problems and dynamical systems~\cite{yang2023icon,herde2024poseidon,mauel2026fimode}.

Alongside these developments, recent theoretical work has investigated how transformers perform in-context learning. One line of work shows that transformers can emulate optimization and estimation algorithms, particularly gradient-based methods for convex statistical problems, where convergence to a global optimum can be guaranteed under suitable conditions~\cite{akyurek2023,vonoswald2023,bai2023,goel2024kalman}. However, these results do not directly address parameter estimation for nonlinear ODEs, where the optimization problem is generally nonconvex and gradient descent may fail to recover the true parameters without a suitable initialization. Another line establishes general function approximation guarantees for mappings from observed examples to predictions, without explicitly exploiting the structure of the underlying statistical problems~\cite{li2025icl,furuya2024universal}. This raises two natural questions:
\emph{Can pretrained transformers accurately estimate unknown parameters and predict trajectories for a new ODE system from finitely many noisy observations? How do the resulting errors depend on the number of pretraining tasks, the context length of a new task, and the size of the transformer?}

To address these questions, we consider a parametric family of ODEs of the form
\begin{equation}
    \dot{x}(t)
    =
    f_{\omega}(x(t),t),
    \qquad
    x(0)=x_0,
    \qquad
    \theta=(x_0,\omega)\in\Theta\subset\mathbb{R}^p,
\end{equation}
where $x(t)\in\mathbb{R}^m$ denotes the system state, $x_0\in\mathbb{R}^m$ is the initial condition, $\omega\in\mathbb{R}^d$ represents the system-specific parameters, and $\Theta$ is the parameter space, with $p=m+d$. The governing equation determines the known functional form of $f_{\omega}$, while each choice of $\theta$ specifies a particular system within the ODE family, which we regard as a distinct task. For a new task, the transformer uses finitely many noisy trajectory observations to estimate the unknown parameters or predict the system's state at specified query times. The precise observation model is introduced later in~\eqref{eq:observations}.

The ideas underlying our analysis may also apply to more general inverse problems with known parametric forward maps~\cite{tarantola2005}, as discussed in Section~\ref{subsec:general-inverse-problems}.

\subsection{Main contributions}
\label{sec:contributions}

Our main contributions are summarized as follows:

\noindent {\bf 1. Nonasymptotic guarantees for parameter estimation.}
Under local sensitivity and global identifiability assumptions, we analyze a two-stage approximate maximum likelihood procedure for recovering unknown ODE parameters from noisy trajectory observations. The procedure first performs a coarse search over the parameter space to obtain an initialization within a locally strongly convex region, followed by gradient descent that contracts linearly toward the empirical minimizer up to approximation errors. The resulting parameter recovery bound separates errors due to finite observations, numerical approximation, and optimization.

\noindent {\bf 2. Explicit transformer construction for estimation and prediction.}
We construct a transformer that implements the two-stage estimation procedure. For trajectory prediction, we compose this estimator with a neural approximation of the ODE solution map. The construction combines an explicit implementation of the estimation algorithm with neural network approximations of the trajectory and sensitivity functions. We quantify the required transformer depth, width, and number of parameters in terms of the approximation accuracy and the local curvature of the estimation problem.

\noindent {\bf 3. Joint estimation and prediction.}
We further show that a single transformer can perform parameter estimation and trajectory prediction simultaneously by exploiting their shared structure. In particular, the joint construction reuses the two-stage estimation procedure and its neural network components for both tasks, avoiding the need for separate transformers. We establish that this unified transformer achieves guarantees for both tasks with essentially the same order of model size as the individual constructions, demonstrating the benefit of parameter sharing without a substantial increase in model complexity.

\noindent {\bf 4. Finite-pretraining guarantees for the meta-learning task.}
We establish oracle inequalities for approximate empirical risk minimizers over suitable transformer classes for parameter estimation, trajectory prediction, and the joint task. Combined with our explicit constructions, these results give risk bounds that distinguish errors due to the finite number of pretraining tasks, the number of observations available for a new task, and transformer approximation, while also accounting for pretraining optimization error.

\noindent {\bf 5. Numerical experiments.}
We conduct experiments on linear and nonlinear ODE systems, including a non-identifiable parameterization, to investigate how estimation and prediction accuracy depend on pretraining data size and context length. In particular, we find that pretrained transformers can benefit from longer inference contexts without updating their weights. We also compare their performance with classical statistical estimators.

Our theoretical guarantees concern the existence of suitable transformer constructions and the statistical properties of approximate empirical pretraining risk minimizers. Explaining why transformers trained at a fixed context length can generalize to longer contexts, as observed in our experiments, remains an open question.

\subsection{Related work}
\label{sec:related-work}

Our work is closely related to recent developments in pretrained models for differential equations. Transformer-based methods have been developed for predicting ODE trajectories~\cite{becker2023predicting} and identifying governing equations from data~\cite{dascoli2024}. Foundation inference models (FIMs) perform zero-shot imputation and dynamical inference from sparse and noisy observations~\cite{seifner2025fim,mauel2026fimode}, while ICON studies in-context operator learning for differential equation problems~\cite{yang2023icon}. Related developments for partial differential equations (PDEs) include neural operators and foundation models that learn solution maps across different parameter regimes and physical systems~\cite{lu2021deeponet,li2021fno,herde2024poseidon}. While these studies demonstrate the potential of pretrained models for scientific problems, our focus is on establishing statistical guarantees for in-context parameter estimation and trajectory prediction, explicitly accounting for dynamical identifiability, finite pretraining data, inference context length, and transformer architecture.

Our theoretical analysis builds on previous work showing that transformers can implement gradient-based learning algorithms for statistical estimation problems, including least squares, ridge regression, and Lasso~\cite{akyurek2023,vonoswald2023,bai2023}. More recent work extends this perspective to nonlinear function learning through functional gradient descent~\cite{cheng2024functional}. Another line of research establishes general function approximation and nonparametric regression guarantees, including approaches based on kernel methods and structured function classes~\cite{furuya2024universal,li2025icl,shi2026crosstask,ching2026nonparametric,shen2026manifolds}. These results provide complementary insights into the in-context learning capabilities of transformers, but do not directly establish the combination of nonconvex parameter recovery, explicit transformer complexity bounds, and finite-pretraining guarantees considered here. Our construction uses the structure of the underlying inverse problem, rather than emulating generic function regression.

\paragraph{Organization}
The remainder of the paper is organized as follows.
Section~\ref{sec:preliminaries} introduces the ODE model, the estimation and prediction tasks, and the pretraining framework.
Section~\ref{sec:main-results} presents the assumptions and theoretical guarantees for transformer construction and pretraining.
Section~\ref{sec:experiments} presents the numerical experiments, and Section~\ref{sec:conclusion} discusses directions for future work. 
Proofs of all theoretical results, further details of the transformer constructions and their implementation, and additional numerical results are provided in the appendix.

\section{Preliminaries and Problem Formulation}
\label{sec:preliminaries}

In this section, we formulate in-context parameter estimation and trajectory prediction for parametric ODEs as meta-learning problems. We first specify the ODE model, observation scheme, and the estimation and prediction maps to be learned. We then introduce the pretraining objectives and the transformer function class used to approximate these maps.

\medskip
\paragraph{Notation}
For a matrix $A$, $\|A\|_{\mathrm{op}}:=\sup_{\|v\|_2=1}\|Av\|_2$ and $\|A\|_F:=(\sum_{i,j}A_{ij}^2)^{1/2}$ denote its operator and Frobenius norms, respectively. For a third-order tensor $\mathcal T=(T_1,\ldots,T_q)$ with matrix slices $T_j\in\mathbb R^{p\times p}$, we define its operator norm as $\|\mathcal T\|_{\mathrm{op}}:=\sup_{\|u\|_2=\|v\|_2=1}\bigl(\sum_{j=1}^q|u^\top T_jv|^2\bigr)^{1/2}$. For a symmetric matrix $A$, let $\lambda_{\min}(A)$ denote its smallest eigenvalue. We use $\operatorname{vect}(A)$ to denote the vector obtained by stacking the columns of $A$, and $\|\cdot\|_2$ for the Euclidean norm of a vector.
For a vector-valued map $h(\theta,t):\mathbb R^{p+1}\to\mathbb R^q$, where $\theta\in\R^p$ and $t\in\R$, $\nabla_\theta h$ denotes its $p\times q$ Jacobian with respect to $\theta$, whose $(k,j)$ entry is $\partial h_j/\partial\theta_k$, and $\nabla_\theta^2 h$ denotes its $q\times p\times p$ Hessian tensor, whose $(i,j,k)$ entry is $\partial^2 h_i/\partial\theta_j\partial\theta_k$. We say that $h$ is $t$-uniformly $C^{2,1}$ in $\theta$ if it is twice continuously differentiable in $\theta$ and its Hessian $\nabla_\theta^2 h(\theta,t)$ is Lipschitz continuous in $\theta$, with a Lipschitz constant independent of $t$.
We also use $\|\cdot\|_{L^\infty(D)}$ to denote the supremum of the pointwise norm of a function over its domain $D$, using the corresponding operator norm for matrix- or tensor-valued functions.

\subsection{ODE and meta-learning problem setup}
\label{subsec:problem-setup}

We consider the parametric ODE family introduced in Section~\ref{sec:introduction}, writing its solution as $x(\theta,t)$ to make the dependence on the system parameters explicit:
\begin{equation}
    \dot{x}(\theta,t)
    =
    f_\omega(x(\theta,t),t),
    \qquad
    x(\theta,0)=x_0,
    \qquad t\in[0,T_{\mathrm{pred}}].
    \label{eq:ode}
\end{equation}
Here, $T_{\mathrm{pred}}>0$ denotes the prediction horizon. Recall that $\theta=(x_0,\omega)\in\mathbb{R}^p$, with $p=m+d$, collects the initial condition $x_0\in\mathbb{R}^m$ and the system-specific parameters $\omega\in\mathbb{R}^d$. For technical convenience, we take the parameter space to be
\[
    \Theta=\bar B(0,D/2)\subset\mathbb{R}^p,
    \qquad D>0,
\]
where $\bar B(0,D/2)$ denotes the closed Euclidean ball of radius $D/2$ centered at the origin. More general compact convex parameter spaces with nonempty interior can also be accommodated with minor modifications.
We assume that the ODE~\eqref{eq:ode} admits a unique solution throughout $[0,T_{\mathrm{pred}}]$ for every parameter in an open neighborhood of $\Theta$. This condition holds, for example, under \eqref{eq:existence}.

\paragraph{Observation model}
We allow the observation horizon $T_{\mathrm{obs}}$ to be shorter than the prediction horizon $T_{\mathrm{pred}}$, with $0<T_{\mathrm{obs}}\leq T_{\mathrm{pred}}$, so that prediction may extend beyond the observed time interval. We assume that the true parameter $\theta^*$ belongs to a compact subset $\Theta_0\Subset\operatorname{int}(\Theta)$, with a positive distance from the boundary of $\Theta$:
\[
    \rho_{\partial}
    :=
    \inf_{\vartheta\in\Theta_0}
    \operatorname{dist}(\vartheta,\Theta^c)>0.
\]
For a fixed true parameter $\theta^*\in\Theta_0$, we observe $n$ noisy measurements of the corresponding ODE trajectory at randomly sampled time points:
\begin{equation}
\begin{aligned}
    \mathcal D^{(n)}
    &=
    \{(t_i,Y(\theta^*,t_i))\}_{i=1}^n
    \sim\mathcal P_{\theta^*}^{(n)},\\
    Y(\theta^*,t_i)
    &=
    x(\theta^*,t_i)+\sigma\xi_i,
    \qquad
    t_i\overset{\mathrm{i.i.d.}}{\sim}\nu,
    \qquad
    \xi_i\overset{\mathrm{i.i.d.}}{\sim}N(0,I_m),
\end{aligned}
\label{eq:observations}
\end{equation}
where $\nu$ is a probability distribution supported on $[0,T_{\mathrm{obs}}]$, $\sigma>0$ is the noise standard deviation, and the observation times $\{t_i\}_{i=1}^n$ and noise variables $\{\xi_i\}_{i=1}^n$ are mutually independent. Our analysis can also be extended to sub-Gaussian noise with minor modifications, but we focus on Gaussian noise for simplicity. Here, $\mathcal P_{\theta^*}^{(n)}$ denotes the joint sampling distribution of the observations under $\theta^*$. 

\paragraph{Meta-learning framework}
In scientific computing with parametric ODEs, two common tasks arise when only noisy trajectory observations are available: (i) \emph{parameter estimation}, which aims to recover the unknown initial conditions and system parameters $\theta^*$ from $\mathcal D^{(n)}$, and (ii) \emph{trajectory prediction}, which aims to predict the system state $x(\theta^*,\tau)$ at a query time $\tau\in[0,T_{\mathrm{pred}}]$, possibly beyond the observation interval. A conventional approach is to first estimate $\theta^*$ by solving a statistical inverse problem, for example by maximum likelihood, and then obtain trajectory predictions by numerically solving the ODE with the estimated parameters.

Our goal, however, is to avoid solving these problems separately for each dataset by meta-learning two maps across related ODE tasks: an estimator $g_{\mathrm{est}}$ and a trajectory predictor $g_{\mathrm{pred}}$, such that
\begin{equation*}
    \widehat\theta
    =
    g_{\mathrm{est}}(\mathcal D^{(n)}),
    \qquad
    \widehat x(\tau)
    =
    g_{\mathrm{pred}}(\mathcal D^{(n)},\tau),
\end{equation*}
and also possibly a joint map $g_{\mathrm{joint}}(\mathcal D^{(n)},\tau)=\big(\hat\theta, \widehat x(\tau)\big)$ that outputs both the estimated parameters and predicted trajectory.
We learn these maps by pretraining on data from multiple ODE tasks. Once pretrained, they can perform in-context parameter estimation and trajectory prediction using observations from a new task as context, without updating their learned parameters. In particular, we seek to learn the trajectory predictor $g_{\mathrm{pred}}$ end-to-end, directly mapping the observations and query time to the predicted state. Rather than explicitly separating statistical parameter estimation and numerical solution of the ODE, we aim to learn a predictor that handles both components implicitly. We next specify the pretraining procedure for learning these maps.

\subsection{Pretraining setup}
\label{subsec:pretraining-setup}

We assume that the system parameters vary across tasks according to a probability distribution $\Pi$ supported on $\Theta_0$. Each task is generated by sampling $\vartheta\sim\Pi$, followed by $n$ noisy trajectory observations $\mathcal D^{(n)}\mid\vartheta\sim\mathcal P_\vartheta^{(n)}$ according to~\eqref{eq:observations}. We use $M$ independently sampled tasks for pretraining, with the goal of learning maps that generalize to new tasks drawn from the same distribution. For trajectory prediction, we also introduce a distribution $\nu'$ supported on $[0,T_{\mathrm{pred}}]$ for sampling query times. Throughout this subsection, we write $\mathcal D=\mathcal D^{(n)}$ for simplicity.

\paragraph{Pretraining for parameter estimation}
For parameter estimation, the pretraining data consist of $M$ independent pairs $\{(\vartheta_j,\mathcal D_j)\}_{j=1}^M$, where $\vartheta_j\sim\Pi$ and $\mathcal D_j\mid\vartheta_j\sim\mathcal P_{\vartheta_j}^{(n)}$. For a measurable estimator $g_{\mathrm{est}}$ taking values in $\mathbb R^p$, we define the population and empirical pretraining risks under the squared-error loss as
\begin{align}
    \mathcal R_{\mathrm{est}}(g_{\mathrm{est}})
    &:=
    \E\big[\|g_{\mathrm{est}}(\mathcal D)-\vartheta\|_2^2\big],
    &
    \mathcal R_{\mathrm{est}}^{(M)}(g_{\mathrm{est}})
    &:=
    \frac{1}{M}\sum_{j=1}^M
    \|g_{\mathrm{est}}(\mathcal D_j)-\vartheta_j\|_2^2.
    \label{eq:pretraining-risks}
\end{align}
The population expectation is taken over both $\vartheta\sim\Pi$ and $\mathcal D\mid\vartheta\sim\mathcal P_\vartheta^{(n)}$. Under squared-error loss, the optimal population estimator is the Bayes estimator
$g_{\mathrm{est}}^{\mathrm{bayes}}(\mathcal D)
    :=
    \E[\vartheta\mid\mathcal D]$.
By the orthogonality property of conditional expectation,
\begin{equation}
\begin{aligned}
    \E[\vartheta-g_{\mathrm{est}}^{\mathrm{bayes}}(\mathcal D)\mid\mathcal D]
    =0,\quad \mbox{and} \quad
    \mathcal R_{\mathrm{est}}(g_{\mathrm{est}})
    -
    \mathcal R_{\mathrm{est}}(g_{\mathrm{est}}^{\mathrm{bayes}})
    =
    \E\big[\|g_{\mathrm{est}}(\mathcal D)-g_{\mathrm{est}}^{\mathrm{bayes}}(\mathcal D)\|_2^2\big].
\end{aligned}
\label{eq:pretraining-bayes-identity}
\end{equation}
Thus, the excess population risk measures the mean squared difference between an estimator $g_{\mathrm{est}}$ and the Bayes estimator $g_{\mathrm{est}}^{\mathrm{bayes}}$.

\paragraph{Pretraining for trajectory prediction}
For trajectory prediction, each pretraining task additionally includes a query time $\tau\sim\nu'$, sampled independently of $\vartheta$ and $\mathcal D$. The corresponding prediction target is the state $x(\vartheta,\tau)$. We obtain $M$ independent samples $\{(\vartheta_j,\tau_j,\mathcal D_j)\}_{j=1}^M$ and define the population and empirical pretraining risks for a measurable predictor $g_{\mathrm{pred}}$ taking values in $\mathbb R^m$ as
\begin{align}
    \mathcal R_{\mathrm{pred}}(g_{\mathrm{pred}})
    &:=
    \E\big[\|g_{\mathrm{pred}}(\mathcal D,\tau)-x(\vartheta,\tau)\|_2^2\big],
    \nonumber\\
    \mathcal R_{\mathrm{pred}}^{(M)}(g_{\mathrm{pred}})
    &:=
    \frac{1}{M}\sum_{j=1}^M
    \|g_{\mathrm{pred}}(\mathcal D_j,\tau_j)-x(\vartheta_j,\tau_j)\|_2^2,
    \label{eq:traj-risks}
\end{align}

with an analogous Bayes predictor $g_{\mathrm{pred}}^{\mathrm{bayes}}$. 

\paragraph{Pretraining for the joint task}
We also analyze pretraining for the joint task of simultaneously performing parameter estimation and trajectory prediction. Thus for the same input data as the prediction task, the corresponding target is both $(\vartheta, x(\vartheta,\tau))$. We obtain $M$ independent samples $\{(\vartheta_j,\tau_j,\mathcal D_j)\}_{j=1}^M$ and define the population and empirical pretraining risks for a measurable predictor $g_{\mathrm{joint}}$ taking values in $\mathbb R^{m+p}$ as
\begin{align}
    \mathcal R_{\mathrm{joint}}(g_{\mathrm{joint}})
    &:=
    \E\big[\|g_{\mathrm{joint}}(\mathcal D,\tau)-(\vartheta,x(\vartheta,\tau))\|_2^2\big],
    \nonumber\\
    \mathcal R_{\mathrm{joint}}^{(M)}(g_{\mathrm{joint}})
    &:=
    \frac{1}{M}\sum_{j=1}^M
    \|g_{\mathrm{joint}}(\mathcal D_j,\tau_j)-(\vartheta_j,x(\vartheta_j,\tau_j))\|_2^2,
    \label{eq:comb-risks}
\end{align}
with an analogous Bayes predictor $g_{\mathrm{joint}}^{\mathrm{bayes}}$. 

We now specify the transformer architecture that will help define our pretraining function class for all three pretraining tasks.
\subsection{Transformer architecture}
\label{subsec:transformer-architecture}

Our transformer architecture uses multi-head attention and token-wise feedforward networks as its basic building blocks, following Vaswani et al.~\cite{Vaswani2017}.

\begin{definition}[Multi-head attention layer]
\label{def:square-attention}
Let $q,H\geq1$ be integers denoting the token dimension and the number of attention heads, respectively, and write $[H]=\{1,\ldots,H\}$. For each $h\in[H]$, let $Q_h,K_h,V_h\in\mathbb R^{q\times q}$ denote the query, key, and value matrices. Given an input sequence $X=[z_1\ \cdots\ z_n]\in\mathbb R^{q\times n}$, the multi-head attention layer $\mathcal A^{H,q,\phi}:\mathbb R^{q\times n}\to\mathbb R^{q\times n}$ is defined by
\[
    \mathcal A^{H,q,\phi}(X)
    :=
    X+
    \sum_{h=1}^H
    V_hX\,
    \operatorname{softmax}_{\mathrm{col}}
    \big((K_hX)^\top(Q_hX)\big),
\]
where $\operatorname{softmax}_{\mathrm{col}}$ applies the softmax function to each column of its argument, and $\phi=\{Q_h,K_h,V_h\}_{h=1}^H$ collects the trainable parameters.
\end{definition}
In our setting, each observation $(t_i,Y(\theta^*,t_i))$ can be represented as an input token, with attention allowing information to be combined across those observations.

\begin{definition}[Two-layer squared-ReLU network]
\label{def:positionwise-ffn}
Let $q_{\mathrm{in}},r,q_{\mathrm{out}}\ge 1$ be integers, and let
$\sigma_2(s)=(s_+)^2$ denote the squared-ReLU activation, applied coordinate-wise.
A two-layer feedforward network with input dimension $q_{\mathrm{in}}$,
hidden width $r$, and output dimension $q_{\mathrm{out}}$ is a map
$\mathcal F^{q_{\mathrm{in}},r,q_{\mathrm{out}},\phi}:
\mathbb R^{q_{\mathrm{in}}}\to\mathbb R^{q_{\mathrm{out}}}$ defined by
\[
\mathcal F^{q_{\mathrm{in}},r,q_{\mathrm{out}},\phi}(z)
=W_2\sigma_2(W_1z+b_1)+b_2,
\]
where $\phi=\{W_1,W_2,b_1,b_2\}$ consists of the trainable parameters,
with $W_1\in\mathbb R^{r\times q_{\mathrm{in}}}$,
$W_2\in\mathbb R^{q_{\mathrm{out}}\times r}$,
$b_1\in\mathbb R^r$, and $b_2\in\mathbb R^{q_{\mathrm{out}}}$.
When applied to an input sequence, the same network is applied
independently to each token. If
$q_{\mathrm{in}}=r=q_{\mathrm{out}}=q$, we write
$\mathcal F^{q,\phi}:=\mathcal F^{q,q,q,\phi}$.
\end{definition}

\noindent We use the squared-ReLU activation mainly for technical convenience, as it allows exact implementation of multiplication. Our results can also be extended to other activation functions, such as ReLU, through standard multiplication approximation arguments.

\begin{definition}[$L$-layer transformer with squared-ReLU activation]
\label{def:transformer-composition}
Let $L,H\ge 1$ be integers, and let $q_0,\ldots,q_L$ and
$r_1,\ldots,r_L$ denote the token dimensions and feedforward hidden
dimensions, respectively. For each layer $\ell=1,\ldots,L$, let
$\mathcal F_\ell
:=\mathcal F^{q_{\ell-1},r_\ell,q_\ell,\phi_\ell^{\mathrm{ffn}}}$,
and $\mathcal A_\ell
:=\mathcal A^{H,q_\ell,\phi_\ell^{\mathrm{attn}}}$.
An $L$-layer transformer with squared-ReLU activation is defined as
the composition
\[
\mathcal T
:=\mathcal A_L\circ\mathcal F_L\circ\cdots
\circ\mathcal A_1\circ\mathcal F_1,
\]
which maps an input sequence in $\mathbb R^{q_0\times n}$ to an
output sequence in $\mathbb R^{q_L\times n}$.
We characterize the complexity of $\mathcal T$ by
$\operatorname{depth}(\mathcal T):=L$, 
$\operatorname{width}(\mathcal T):=\max\{q_0,\ldots,q_L,r_1,\ldots,r_L\}$,
and total number of trainable parameters
$\#\operatorname{params}(\mathcal T)
:=\sum_{\ell=1}^L
\left(\dim(\phi_\ell^{\mathrm{ffn}})
+\dim(\phi_\ell^{\mathrm{attn}})\right).$

\end{definition}

\section{Theoretical results}
\label{sec:main-results}

Recall the general pretraining approach. For a class $\mathcal A$ of predictors for a pretraining task, let $\widehat a_M$ be an approximate empirical pretraining risk minimizer satisfying
\[
\mathcal R_{\mathrm{pre}}^{(M)}(\widehat a_M) \
\leq \ \inf_{a\in\mathcal A}\mathcal R_{\mathrm{pre}}^{(M)}(a)
+\varepsilon_{\mathrm{opt}},
\]
where $\varepsilon_{\mathrm{opt}}\geq 0$ denotes the empirical optimization error. Usual analysis aims to establish an oracle inequality of the form
\begin{equation}
\begin{aligned}
\mathcal{R}_{\mathrm{pre}}(\widehat a_M)
-\mathcal{R}_{\mathrm{pre}}(a^{\mathrm{bayes}})\
\lesssim  \ \
\text{within-task error} +\text{pretraining estimation error}
\ +\ \varepsilon_{\mathrm{opt}}.
\end{aligned}
\label{eq:pretraining-oracle-inequality}
\end{equation}
The within-task error has two sources in our case: an estimation error from recovering the ODE parameters from $n$ noisy observations, and an approximation error from implementing the estimation procedure by a finite transformer. The pretraining estimation error arises from learning the transformer predictor using $M$ tasks. We include a nonzero training error to make our bounds more flexible, and do not address the optimization procedure in this work.

To address the within-task error, we first construct transformers that implement a two-stage approximate maximum likelihood procedure for parameter recovery, and then extend the construction to trajectory prediction. We next use these transformers as comparators in the inequality above to derive pretraining risk bounds for both tasks. We also discuss an immediate possible extension to more general inverse problems. Proof details are provided in Appendix~\ref{sec:proof-ideas}.

\subsection{Assumptions}

Our results rely on the regularity of the ODE solution map and both local and global identifiability of the ODE parameters. Regularity allows us to control numerical and statistical approximation errors, while local identifiability ensures sufficient curvature of the loss near the true parameter, and global identifiability rules out other parameters that produce the same trajectory.

\begin{assumption}[ODE regularity]
\label{ass:ode-regularity}
In addition to uniqueness of solutions, assume that the solution map $x(\theta,t)$ is $t$-uniformly $C^{2,1}$ in $\theta$, with $x(\theta,t)$, $\nabla_\theta x(\theta,t)$, and $\nabla_\theta^2 x(\theta,t)$ jointly continuous in $(\theta,t)$. In particular, the Hessian is uniformly Lipschitz in $\theta$, with
\[
L_H:=\sup_{t\in[0,T_{\mathrm{pred}}]}
\sup_{\theta\ne\vartheta}
\frac{\|\nabla_\theta^2 x(\theta,t)-\nabla_\theta^2 x(\vartheta,t)\|_{\mathrm{op}}}
{\|\theta-\vartheta\|_2}<\infty.
\]
\end{assumption}

\begin{assumption}[Global and local identifiability]
\label{ass:quantitative-identifiability}
There exist constants $\mu,\gamma_\rho>0$ and $\rho\in(0,\rho_\partial]$ such that, for every $\vartheta\in\Theta_0$, the following conditions hold:
\begin{enumerate}
\item[(i)] \emph{Local curvature:}
$\lambda_{\min}\left(
\nabla_\theta^2
\E_{t\sim\nu}\big[\|x(\theta,t)-x(\vartheta,t)\|_2^2\big]
\right)\ge\mu$, for
$\theta\in B(\vartheta,\rho)$.

\item[(ii)] \emph{Global separation:}
$\E_{t\sim\nu}\big[\|x(\theta,t)-x(\vartheta,t)\|_2^2\big]
\ge\gamma_\rho$, for $\theta\in\Theta$ and  
$\|\theta-\vartheta\|_2\ge\rho/4$.
\end{enumerate}
\end{assumption}

\begin{remark}
The regularity condition in Assumption~\ref{ass:ode-regularity} can be verified through the sensitivity equations obtained by differentiating the ODE with respect to $\theta$. In particular, it follows under suitable conditions if $f$ is $t$-uniformly $C^{2,1}$ in $(x,\omega)$, with its derivatives jointly continuous in $(x,\omega,t)$. Existence and uniqueness of solutions over $[0,T_{\mathrm{pred}}]$ are also guaranteed under standard conditions, such as local Lipschitz continuity in $x$ together with the linear growth bound
\begin{equation}
\|f(x,\omega,t)\|_2\le a+b\|x\|_2,
\qquad x\in\mathbb R^m,\quad t\in[0,T_{\mathrm{pred}}],
\label{eq:existence}
\end{equation}
for some finite constants $a,b\ge0$ uniformly over admissible $\omega$.
Under the above regularity conditions, the quantitative identifiability conditions in Assumption~\ref{ass:quantitative-identifiability} can also be derived from more standard conditions. Specifically, consider
$\inf_{\theta\in\Theta_0}
\lambda_{\min}\big(\mathcal I_\nu(\theta)\big)\ge\mu$, where
$\mathcal I_\nu(\theta)
:=\E_{t\sim\nu}\big[
\nabla_\theta x(\theta,t)
\nabla_\theta x(\theta,t)^\top
\big]$
is the sensitivity matrix, together with the global identifiability condition
\[
\E_{t\sim\nu}\big[
\|x(\theta,t)-x(\vartheta,t)\|_2^2
\big]>0,
\qquad
\vartheta\in\Theta_0,\quad
\theta\in\Theta\setminus\{\vartheta\}.
\]
Under compactness and continuity, these conditions imply the local curvature and global separation bounds for sufficiently small $\rho$, with constants uniform over $\Theta_0$.
Furthermore, when $\operatorname{supp}(\nu)=[0,T_{\mathrm{obs}}]$, a sufficient condition for the positive definiteness of $\mathcal I_\nu(\theta)$, at the ODE level, is
\[
\E_{t\sim\nu}\big[
\|\nabla_\omega f(x(\theta,t),\omega,t)^\top v\|_2^2
\big]>0,
\qquad v\in\mathbb R^d\setminus\{0\}.
\]
We discuss these sufficient conditions in more detail in Appendix~\ref{app:local-conditions}.
\end{remark}

\subsection{Transformer construction for estimation and prediction}
\label{subsec:inverse-transformer}

We first show that the two-stage approximate maximum likelihood procedure can be implemented by a transformer, and then extend the construction to trajectory prediction. Rather than specifying a particular ODE solver, we characterize the approximations of the solution map and the sensitivity-related quantities through their accuracy and network complexity. This allows the component networks to exploit the specific structure of the ODE family.

\medskip
\paragraph{Parameter estimation}
We define two types of squared-ReLU networks used in our construction. Firstly, for any $T\in[0,T_{\mathrm{pred}}]$ and $\varepsilon_{\mathrm{sol}}>0$ define the network $\mathrm{NN}^{\varepsilon_{\mathrm{sol}}}_{\mathrm{sol},T}:\Theta\times[0,T]\to\R^m$, that approximates the solution map $x(\theta,t)$ with accuracy
$\|\mathrm{NN}^{\varepsilon_{\mathrm{sol}}}_{\mathrm{sol},T}-x\|_{L^\infty(\Theta\times[0,T])}\le\varepsilon_{\mathrm{sol}}$.
We denote its depth, width, and number of parameters by $L_{\mathrm{sol},T}(\varepsilon_{\mathrm{sol}})$, $W_{\mathrm{sol},T}(\varepsilon_{\mathrm{sol}})$, and $P_{\mathrm{sol},T}(\varepsilon_{\mathrm{sol}})$, respectively. Secondly, for any $\varepsilon_{\mathrm{grad}}>0$ define $\mathrm{NN}^{\varepsilon_{\mathrm{grad}}}_{\mathrm{grad}}:\Theta\times[0,T_{\mathrm{obs}}]\to\R^{pm+p}$ as an approximation of sensitivity-related quantities such that
\[
\left\|
\mathrm{NN}^{\varepsilon_{\mathrm{grad}}}_{\mathrm{grad}}(\theta,t)
-2\begin{bmatrix}
\operatorname{vect}(\nabla_\theta x(\theta,t))\\
\nabla_\theta x(\theta,t)x(\theta,t)
\end{bmatrix}
\right\|_{L^\infty(\Theta\times[0,T_{\mathrm{obs}}])}
\le\varepsilon_{\mathrm{grad}}.
\]
Its depth, width, and number of parameters are denoted by $L_{\mathrm{grad}}(\varepsilon_{\mathrm{grad}})$, $W_{\mathrm{grad}}(\varepsilon_{\mathrm{grad}})$, and $P_{\mathrm{grad}}(\varepsilon_{\mathrm{grad}})$, respectively. Such choices exist by standard neural network approximation results \cite{yarotsky2017,li2020powernet}.

For brevity, let
$M_j:=\|\nabla_\theta^j x\|_{L^\infty(\Theta\times[0,T_{\mathrm{obs}}])}$
for $j=0,1,2$, with $\nabla_\theta^0 x=x$. We also define
\[
N:=
\mathcal N\left(
\Theta,\,
\min\left\{
\frac{\rho}{4},
\left(\frac{\gamma_\rho}{4M_1^2+8M_0M_2}\right)^{1/2}
\right\};
\|\cdot\|_2
\right),
\]
where $\mathcal N(\Theta,\delta;\|\cdot\|_2)$ denotes the covering number of $\Theta$ at scale $\delta>0$ under the Euclidean norm. Since $\Theta$ is compact, $N$ is finite and depends on the geometry of the parameter space.

Each observation $(t_i,Y_i)$ is represented by one token, so the data from a task with true parameter $\theta^*$ form the input sequence
\[
\mathcal D^{(n)}
:=
\begin{pmatrix}
t_1 & \cdots & t_n\\
Y_1 & \cdots & Y_n
\end{pmatrix}
\in\mathbb R^{(m+1)\times n}.
\]
The following theorem establishes the existence of a transformer that implements the two-stage estimation procedure, together with explicit bounds on its complexity and parameter recovery error.

\begin{theorem}[Transformer implements parameter estimation]
\label{thm:inverse-transformer-realization}
Let approximation tolerances $\varepsilon_{\mathrm{sol}}^{(\mathrm{obs})},\varepsilon_{\mathrm{grad}}>0$ satisfy
\begin{equation}
\varepsilon_{\mathrm{sol}}^{(\mathrm{obs})}
=2\left(\sqrt{M_0^2+\frac{\gamma_\rho}{32}}-M_0\right),
\qquad
\varepsilon_{\mathrm{grad}}
<\frac{3\rho\mu}
{8(1+M_0+\sigma(\sqrt m+1))}.
\label{eq:transformer-approximation-budgets}
\end{equation}
Then, for any $\varepsilon_{\mathrm{gd}}\in(0,1)$, there exists a single-head transformer $\mathcal T_{\mathrm{est}}$ with squared-ReLU activations whose depth, width, and number of parameters satisfy
\begin{equation}
\begin{aligned}
\operatorname{depth}(\mathcal T_{\mathrm{est}})
&=O\left(
L_{\mathrm{sol},T_{\mathrm{obs}}}(\varepsilon_{\mathrm{sol}}^{(\mathrm{obs})})+\log(N)
+\log(e/\varepsilon_{\mathrm{gd}})L_{\mathrm{grad}}(\varepsilon_{\mathrm{grad}})
\right),\\
\operatorname{width}(\mathcal T_{\mathrm{est}})
&=O\left(
NW_{\mathrm{sol},T_{\mathrm{obs}}}(\varepsilon_{\mathrm{sol}}^{(\mathrm{obs})})+W_{\mathrm{grad}}(\varepsilon_{\mathrm{grad}})
\right),
\\
\#\operatorname{params}(\mathcal T_{\mathrm{est}})
&=O\left(
N^2P_{\mathrm{sol},T_{\mathrm{obs}}}(\varepsilon_{\mathrm{sol}}^{(\mathrm{obs})})
+\log(e/\varepsilon_{\mathrm{gd}})P_{\mathrm{grad}}(\varepsilon_{\mathrm{grad}})
\right),
\end{aligned}
\label{eq:transformer-component-size}
\end{equation}
and, for every $\alpha\in(0,1/7)$, $n>\mathfrak N(\alpha)$, and $\theta^*\in\Theta_0$, the transformer outputs
$\mathcal T_{\mathrm{est}}(\mathcal D^{(n)})
=\widehat\theta_n^{\mathrm{out}}\mathbf1_n^\top$,
where the parameter estimate satisfies
\begin{equation}
\begin{aligned}
\|\widehat\theta_n^{\mathrm{out}}-\theta^*\|_2
\lesssim\;&
C(\sigma,D,M_0,M_1)
\sqrt{
\frac{
(2p+1)\log n
+(2p+1)\log\sqrt m
+\log(1/\alpha)}
{\mu n}
}\\
&+
(1+M_0+\sigma(\sqrt m+1))
\frac{\varepsilon_{\mathrm{grad}}}{\mu}
+\varepsilon_{\mathrm{gd}}
\end{aligned}
\label{eq:transformer-component-recovery}
\end{equation}
with probability at least $1-7\alpha$. Here $\mathfrak N(\alpha)$ is a sample size threshold depending on $\alpha$ and the ODE family, and the constant $C(\sigma,D,M_0,M_1)$ is of linear order in $\sigma$ and $M_0$, and square-root logarithmic order in $M_1$ and $D$.
\end{theorem}

The recovery bound separates three sources of error respectively: statistical error due to finite observations, approximation error in the sensitivity-related quantities $\varepsilon_{\mathrm{grad}}$, and within-task optimization error $\varepsilon_{\mathrm{gd}}$ from the finite number of GD steps emulated by the transformer. The solution approximation accuracy over the observation interval $\varepsilon_{\mathrm{sol}}^{(\mathrm{obs})}$ enters through the construction and its complexity requirements, rather than appearing explicitly in the recovery bound. For fixed problem dimensions and constants, the leading statistical error is of order $\sqrt{\log n/n}$, provided the approximation and optimization errors are sufficiently small.

\begin{remark}[Complexity of the component networks]
The complexity functions $L_{(\cdot)}(\varepsilon_{(\cdot)})$, $W_{(\cdot)}(\varepsilon_{(\cdot)})$, and $P_{(\cdot)}(\varepsilon_{(\cdot)})$ can be specified using existing neural network approximation results. For example, Li, Tang, and Yu~\cite{li2020powernet} show that sparse $s$-power-ReLU networks can represent a $d$-dimensional polynomial of degree $K$ exactly with depth $O(d\log_s K)$, width $O(K^d)$, and $O(K^d)$ unique nonzero parameters. With additional uniform Lipschitz regularity in time, such results can yield networks whose depths grow logarithmically and widths grow polynomially in the inverse approximation accuracy. That said, this additional time regularity is not required for the existence of finite networks for arbitrarily small approximation errors \cite{yarotsky2017}.
\end{remark}

\begin{remark}[Solver-based component networks]
The approximating networks $\mathrm{NN}^{\varepsilon_{\mathrm{sol}}}_{\mathrm{sol},T}$ and $\mathrm{NN}^{\varepsilon_{\mathrm{grad}}}_{\mathrm{grad}}$ can also be constructed by emulating numerical ODE solvers when the drift admits a convenient representation. For example, for a fixed exact squared-ReLU drift network, unrolling $K$ Euler steps with a query-time-dependent step size $t/K$ yields uniform trajectory approximation error $\varepsilon$ with depth $O(\varepsilon^{-1})$ and constant width, under suitable uniform regularity conditions. If the drift itself is approximated, the network complexity also depends on the accuracy and cost of this approximation. Such solver-based constructions can be preferable to generic approximation bounds when the drift is easy to represent, since their approximation exponents need not depend on the parameter dimension as above. We provide an Euler-based construction in Appendix~\ref{subsec:euler-nn1}.
\end{remark}

\begin{remark}[Parameter sharing and attention heads]
The transformer construction uses repeated sparse subnetworks, whose weights can be shared to reduce the number of distinct trainable parameters. By tying their weights and counting only nonzero parameters, the parameter count can be reduced to
\[
O\left(
(W_{\mathrm{sol},T_{\mathrm{obs}}})^{\beta_1}(\varepsilon_{\mathrm{sol}}^{(\mathrm{obs})})L_{\mathrm{sol},T_{\mathrm{obs}}}(\varepsilon_{\mathrm{sol}}^{(\mathrm{obs})})
+W_{\mathrm{grad}}^{\beta_2}(\varepsilon_{\mathrm{grad}})L_{\mathrm{grad}}(\varepsilon_{\mathrm{grad}})
+Np
\right),
\]
for some $\beta_1,\beta_2\in[1,2]$, with the worst-case bound corresponding to $\beta_1=\beta_2=2$. Moreover, although our construction uses only one attention head, the same estimation procedure can also be implemented with an arbitrary number of heads without changing the error bound.
\end{remark}

\medskip
\paragraph{Trajectory prediction} We next extend the construction to trajectory prediction. Given the parameter estimate produced by $\mathcal T_{\mathrm{est}}$, prediction at a query time $\tau$ can be obtained by evaluating an approximation of the ODE solution map at the estimated parameter. The prediction error is therefore controlled by the parameter recovery error and the approximation accuracy of the solution map over the prediction horizon.

Let $e_n^{\mathrm{est}}(\alpha)$ denote the right-hand side of the recovery bound \eqref{eq:transformer-component-recovery}. To incorporate the query time $\tau$ into the transformer input, we append it to every observation token, so the sequence length remains $n$. Specifically, define
\[
\mathcal D_\tau^{(n)}
:=
\begin{pmatrix}
t_1 & \cdots & t_n\\
Y_1 & \cdots & Y_n\\
\tau & \cdots & \tau
\end{pmatrix}
\in\mathbb R^{(m+2)\times n}.
\]
The following corollary establishes the complexity and prediction accuracy of the resulting transformer.

\begin{corollary}[Transformer implements trajectory prediction]
\label{cor:prediction-realization}
Under the conditions of Theorem~\ref{thm:inverse-transformer-realization}, for any $\varepsilon_{\mathrm{sol}}^{(\mathrm{pred})}>0$, there exists a single-head transformer $\mathcal T_{\mathrm{pred}}$ with squared-ReLU activations whose depth, width, and number of parameters satisfy
\begin{equation}
\begin{aligned}
\operatorname{depth}(\mathcal T_{\mathrm{pred}})
&=O\left(
\operatorname{depth}(\mathcal T_{\mathrm{est}})
+L_{\mathrm{sol},T_{\mathrm{pred}}}(\varepsilon_{\mathrm{sol}}^{(\mathrm{pred})})
\right)\\
\operatorname{width}(\mathcal T_{\mathrm{pred}})
&=O\left(
\operatorname{width}(\mathcal T_{\mathrm{est}})
+W_{\mathrm{sol},T_{\mathrm{pred}}}(\varepsilon_{\mathrm{sol}}^{(\mathrm{pred})})
\right),
\\
\#\operatorname{params}(\mathcal T_{\mathrm{pred}})
&=O\left(
\#\operatorname{params}(\mathcal T_{\mathrm{est}})
+P_{\mathrm{sol},T_{\mathrm{pred}}}(\varepsilon_{\mathrm{sol}}^{(\mathrm{pred})})
\right),
\end{aligned}
\label{eq:trajectory-component-size}
\end{equation}
and, for every $\alpha\in(0,1/7)$, $n>\mathfrak N(\alpha)$, $\theta^*\in\Theta_0$, and $\tau\in[0,T_{\mathrm{pred}}]$, the transformer outputs
$\mathcal T_{\mathrm{pred}}(\mathcal D_\tau^{(n)})
=\widehat x_n(\tau)\mathbf1_n^\top$,
where the prediction satisfies
\begin{equation}
\sup_{\tau\in[0,T_{\mathrm{pred}}]}
\|\widehat x_n(\tau)-x(\theta^*,\tau)\|_2
\lesssim
\varepsilon_{\mathrm{sol}}^{(\mathrm{pred})}+
\|\nabla_\theta x\|_{L^\infty(\Theta\times[0,T_{\mathrm{pred}}])}
e_n^{\mathrm{est}}(\alpha)
\label{eq:prediction-error}
\end{equation}
with probability at least $1-7\alpha$.
\end{corollary}

The prediction error consists of two terms. The first, $\varepsilon_{\mathrm{sol}}^{(\mathrm{pred})}$, accounts for the approximation of the solution map over the prediction horizon $[0,T_{\mathrm{pred}}]$, while the second propagates the parameter recovery error through the sensitivity of the solution map to the ODE parameters. In particular, the bound holds uniformly over the entire prediction horizon $[0,T_{\mathrm{pred}}]$, including times beyond the observation interval $[0,T_{\mathrm{obs}}]$.

\medskip
\paragraph{Joint estimation and prediction} An important feature of our prediction construction is that it already contains the two-stage parameter estimation procedure as an intermediate computation. By retaining the estimated parameter as an additional output, we can construct a single transformer that performs both estimation and prediction without duplicating the estimation network. The following corollary formalizes this observation.

\begin{corollary}[Joint estimation and prediction]
\label{cor:joint-realization}
Under the conditions of Corollary~\ref{cor:prediction-realization}, there exists a single-head transformer $\mathcal T_{\mathrm{joint}}$ with squared-ReLU activations whose depth, width, and number of parameters are of the same order as those of $\mathcal T_{\mathrm{pred}}$ in \eqref{eq:trajectory-component-size}.
Moreover, for every $\alpha\in(0,1/7)$, $n>\mathfrak N(\alpha)$, $\theta^*\in\Theta_0$, and $\tau\in[0,T_{\mathrm{pred}}]$, the transformer produces
\[
\mathcal T_{\mathrm{joint}}(\mathcal D_\tau^{(n)})
=
\begin{pmatrix}
\widehat\theta_n^{\mathrm{out}}\\
\widehat x_n(\tau)
\end{pmatrix}
\mathbf1_n^\top,
\]
where the parameter recovery and prediction bounds \eqref{eq:transformer-component-recovery} and \eqref{eq:prediction-error} hold simultaneously with probability at least $1-7\alpha$.
\end{corollary}

This result shows that the joint transformer achieves both statistical guarantees without a substantial increase in model size. This provides a theoretical justification for exploiting the shared structure of closely related statistical tasks in transformer architectures.

\subsection{Pretraining guarantees}
\label{subsec:pretraining-comparator}

We now use the transformer constructions for parameter estimation and trajectory prediction as comparators in the oracle inequality \eqref{eq:pretraining-oracle-inequality}. To this end, we introduce bounded transformer classes that contain these constructions and establish pretraining risk bounds based on $M$ independently sampled tasks.

We first modify the transformer architecture to accommodate the ODE estimation setting and facilitate the theoretical analysis. Specifically, we introduce input and output transformations to encode the observations into a common hidden representation and extract the desired estimates. We also impose boundedness constraints on the network parameters and intermediate representations to control the complexity of the resulting function class.
Let $\operatorname{Clip}_R$ denote coordinate-wise clipping to $[-R,R]$, $\operatorname{Proj}_R$ denote Euclidean projection onto the closed ball $\bar B(0,R)$, and $\operatorname{Read}_j$ denote the operation that extracts the $j$th token from an output sequence. When applied to a sequence, the projection acts independently on each token.

Fix constants $R_1,R_2,R_3,M_A,M_b>0$, input and output dimensions $d_{\mathrm{in}},d_{\mathrm{out}}\ge1$, depth $L\ge2$, hidden width $W\ge\max\{d_{\mathrm{in}},d_{\mathrm{out}}\}$, number of attention heads $H\ge1$, and readout position $j\in\{1,\ldots,n\}$. For each layer $\ell=1,\ldots,L$, define the modified transformer block
\[
\mathcal B_{\ell,\phi}
:=
\operatorname{Proj}_{R_2}
\circ\mathcal A^{H,W,\phi_\ell^{\mathrm{attn}}}
\circ\operatorname{Proj}_{R_2}
\circ\mathcal F^{W,\phi_\ell^{\mathrm{ffn}}}.
\]
Here we use a common token and feedforward hidden width $W$, corresponding to $q_\ell=r_\ell=W$, to simplify the analysis. Compared with the transformer blocks defined earlier, we introduce projections after the feedforward and attention layers to keep the intermediate token representations bounded.

Let $P_{\mathrm{in}}\in\mathbb R^{W\times d_{\mathrm{in}}}$ and $P_{\mathrm{out}}\in\mathbb R^{d_{\mathrm{out}}\times W}$ be trainable linear maps that encode the input tokens into the hidden space and map the hidden representations to the output space, respectively. Both maps act independently on each token. We define the modified transformer by
\[
\mathcal T_\phi^{\mathrm{bounded}}
:=
\operatorname{Read}_j
\circ\operatorname{Proj}_{R_3}
\circ P_{\mathrm{out}}
\circ\mathcal B_{L,\phi}
\circ\cdots\circ\mathcal B_{1,\phi}
\circ P_{\mathrm{in}}
\circ\operatorname{Clip}_{R_1},
\]
where $\phi\in\mathbb R^P$ collects all trainable parameters in the feedforward and attention layers, together with $P_{\mathrm{in}}$ and $P_{\mathrm{out}}$.
Let $\Phi_{M_A,M_b}\subset\mathbb R^P$ denote the set of parameter vectors satisfying $\|A\|_{\mathrm{op}}\le M_A$ and $\|b\|_2\le M_b$ for every trainable weight matrix $A$ and bias vector $b$. We define the corresponding bounded transformer class as
\begin{equation}
\mathfrak T^{R_1,R_2,R_3,M_A,M_b}_{d_{\mathrm{in}},d_{\mathrm{out}},W,L,H,j}
:=
\left\{
\mathcal T_\phi^{\mathrm{bounded}}:
\phi\in\Phi_{M_A,M_b}
\right\}.
\label{eq:pretraining-class}
\end{equation}
Each transformer in this class maps an input sequence in $\mathbb R^{d_{\mathrm{in}}\times n}$ to an output in $\mathbb R^{d_{\mathrm{out}}}$, with at most $P=O(HW^2L)$ trainable parameters.

The input clipping handles the unbounded Gaussian observations, while the intermediate projections and parameter bounds allow us to control the Lipschitz properties of the transformer. In Lemma~\ref{lem:transformer-parameter-bounds} of Appendix~\ref{app:background}, we characterize the radius $R_\Phi$ of the parameter domain and the parameter-Lipschitz constant $\mathfrak L_\Phi$ of this function class. Moreover, the final projection ensures bounded outputs, which is useful for establishing the pretraining risk guarantees. The class also accommodates networks with smaller intermediate widths through zero-padding, while certain affine operations can be absorbed into adjacent feedforward layers. These properties allow us to incorporate the explicit transformer constructions into the bounded class.

\medskip
\paragraph{Joint estimation and prediction}
We focus on a pretraining result for the joint task using the construction from Corollary~\ref{cor:joint-realization}. Fix tolerances $\varepsilon_{\mathrm{sol}}^{(\mathrm{obs})},\varepsilon_{\mathrm{grad}},\varepsilon_{\mathrm{gd}},\varepsilon_{\mathrm{sol}}^{(\mathrm{pred})}$ satisfying the conditions of that theorem, and let $\mathcal T_{\mathrm{joint}}^{\varepsilon_{\mathrm{sol}}^{(\mathrm{obs})},\varepsilon_{\mathrm{grad}},\varepsilon_{\mathrm{gd}},\varepsilon_{\mathrm{sol}}^{(\mathrm{pred})}}$ denote the resulting construction. Since this transformer has finitely many parameters, we can choose $M_A^{\mathrm{joint}}$ and $M_b^{\mathrm{joint}}$ sufficiently large to accommodate all its weights and biases.

We choose the depth and width of the bounded transformer class to match those of the constructed joint transformer, so that
\begin{align*}
L_{\mathrm{joint}}
&=O\left(
L_{\mathrm{sol},T_{\mathrm{obs}}}(\varepsilon_{\mathrm{sol}}^{(\mathrm{obs})})+\log(2N)
+\log(e/\varepsilon_{\mathrm{gd}})L_{\mathrm{grad}}(\varepsilon_{\mathrm{grad}})+L_{\mathrm{sol},T_{\mathrm{pred}}}(\varepsilon_{\mathrm{sol}}^{(\mathrm{pred})})
\right),\\
W_{\mathrm{joint}}
&=O\left(
NW_{\mathrm{sol},T_{\mathrm{obs}}}(\varepsilon_{\mathrm{sol}}^{(\mathrm{obs})})+W_{\mathrm{grad}}(\varepsilon_{\mathrm{grad}})+W_{\mathrm{sol},T_{\mathrm{pred}}}(\varepsilon_{\mathrm{sol}}^{(\mathrm{pred})})
\right),
\end{align*}
where the dependence of $L_{\mathrm{joint}}$ and $W_{\mathrm{joint}}$ on the approximation tolerances is suppressed for brevity.
To incorporate the clipping and projections without affecting the comparator error bound, we choose suitable radii. For $\eta\in(0,1)$, set
\[
R_1^{\mathrm{joint}}(\eta)
:=
\max\left\{
T_{\mathrm{pred}},\,
\|x\|_{L^\infty(\Theta\times[0,T_{\mathrm{obs}}])}
+\sigma\sqrt{2\log(2mn/\eta)}
\right\},
\]

The input clipping then leaves the observations unchanged with probability at least $1-\eta$. Since the clipped inputs are bounded and the transformer has finitely many layers, we can choose $R_2^{\mathrm{joint}}$ sufficiently large so that the intermediate projections do not alter the constructed computation. For our fixed transformer, a worst-case bound on $R_2^{\mathrm{joint}}$ can be obtained by propagating the input magnitude through its layers, yielding polynomial dependence on $(R_1^{\mathrm{joint}})^L$, with degree independent of $n$.
Finally, set $R_3^{\mathrm{joint}}:=D_{\mathrm{joint}}/2$ where 
\[
D_{\mathrm{joint}}
:=
2\sqrt{2}\max\left\{
1,\frac D2,
\|x\|_{L^\infty(\Theta\times[0,T_{\mathrm{pred}}])}
\right\},
\]
We therefore define the joint class
\[
\mathcal A_{\mathrm{joint}}
:=
\mathfrak T^{R_1^{\mathrm{joint}}(\eta),R_2^{\mathrm{joint}},R_3^{\mathrm{joint}},
M_A^{\mathrm{joint}},M_b^{\mathrm{joint}}}_{m+2,m+p,W_{\mathrm{joint}},L_{\mathrm{joint}},H,j},
\]
where $H\ge1$ and $j\in\{1,\ldots,n\}$. This class contains the joint construction with the corresponding clipping, projections, and readout. Let $P_{\mathrm{joint}}$, $R_\Phi^{\mathrm{joint}}$, and $\mathfrak L_\Phi^{\mathrm{joint}}$ denote its number of trainable parameters, parameter-domain radius, and parameter-Lipschitz constant, respectively, as characterized in Lemma~\ref{lem:transformer-parameter-bounds}. In particular, only $\mathfrak{L}_\Phi^{\mathrm{joint}}$ depends on $n$ through $R_1^{\mathrm{joint}}(\eta)$.

Let $e_n^{\mathrm{pred}}(\alpha)$ denote the right-hand side of the prediction error bound \eqref{eq:prediction-error}. The following theorem combines the joint construction's error guarantee with the pretraining oracle inequality.

\begin{theorem}[Pretraining guarantee for joint estimation and prediction]
\label{thm:joint-pretraining}
Under the conditions of Corollary~\ref{cor:joint-realization}, fix $\eta\in(0,1)$, $\alpha\in(0,1/7)$, and $n>\mathfrak N(\alpha)$. For any $\delta\in(0,1)$ and any $\varepsilon_{\mathrm{opt}}$-approximate empirical pretraining risk minimizer $\widehat g_{\mathrm{joint}}$ over $\mathcal A_{\mathrm{joint}}$, we have
\begin{align}
\mathcal R_{\mathrm{joint}}(\widehat g_{\mathrm{joint}})
-\mathcal R_{\mathrm{joint}}(g_{\mathrm{joint}}^{\mathrm{bayes}})\notag
&\quad\lesssim
e_n^{\mathrm{est}}(\alpha)^2+
e_n^{\mathrm{pred}}(\alpha)^2
+D_{\mathrm{joint}}^2(7\alpha+\eta)+\varepsilon_{\mathrm{opt}}\notag\\
&\quad+\frac{D_{\mathrm{joint}}^2}{M}
\left[
P_{\mathrm{joint}}\log\left(
e+\frac{
M\mathfrak L_\Phi^{\mathrm{joint}}
R_\Phi^{\mathrm{joint}}}{D_{\mathrm{joint}}}
\right)
+\log\frac{2}{\delta}
\right]
\label{eq:joint-pretraining}
\end{align}
with probability at least $1-\delta$ over the pretraining tasks, where the implied constant is universal and $\mathfrak L_\Phi^{\mathrm{joint}}$ is polylogarithmic in $n/\eta$.
\end{theorem}

The bound separates errors from finite context length, finite pretraining data, and pretraining optimization. Specifically, the first three terms account for the comparator error, including the finite-context estimation error and the small-probability events excluded by the recovery and clipping guarantees. The term involving $M$ controls the statistical error due to finite pretraining data, while $\varepsilon_{\mathrm{opt}}$ measures the pretraining optimization error. For fixed problem dimensions and transformer complexity, the bound gives a finite-context contribution of order $\widetilde O(1/n)$ and a pretraining estimation contribution of order $\widetilde O(P_{\mathrm{joint}}/M)$, up to the approximation, optimization, and failure-probability terms.

Exactly analogous results hold for the estimation and prediction tasks (Corollaries~\ref{cor:recovery-pretraining} and~\ref{cor:prediction-pretraining}). That said, the joint-task guarantee controls both parameter estimation and trajectory prediction errors under the same function class complexity as the prediction transformer.

\subsection{Extensions to general inverse problems}
\label{subsec:general-inverse-problems}

Our transformer constructions can be extended beyond ODEs to general inverse problems with a well-defined forward map. Consider observations of the form
\[
Y_i=G(\theta,Z_i)+\sigma\xi_i,\qquad
Z_i\overset{\mathrm{i.i.d.}}{\sim}\nu,\quad
\xi_i\overset{\mathrm{i.i.d.}}{\sim}N(0,I_m),
\]
where $\nu$ is supported on a compact set $\mathcal Z\subset\R^s$, and the design and noise variables are mutually independent. The ODE setting corresponds to $s=1$ and $G(\theta,Z)=x(\theta,t)$, with the observation time as the design variable.
Imposing conditions exactly analogous to Assumptions~\ref{ass:ode-regularity} and~\ref{ass:quantitative-identifiability} by replacing $x(\theta,t)$ with $G(\theta,Z)$ and their corresponding domains, the same two-stage estimation procedure applies to the least-squares loss associated with $G$, using a coarse search followed by approximate gradient descent. Since our constructions and pretraining results only use the forward map, it is not hard to see that exactly analogous results can be proven. We leave a formal treatment of these extensions to future work.

\section{Numerical experiments}
\label{sec:experiments}

We evaluate pretrained transformers for each task across several parametric ODE families. Our experiments examine how performance depends on the number of pretraining tasks $M$ and the number of observations $n$ available within each task. We also investigate whether pretrained transformers can benefit from inference contexts of different lengths without updating their weights, and compare their performance with classical statistical estimators that use the known ODE solution. 

\medskip
\paragraph{Code availability}
Code for data generation, training, evaluation, and figure reproduction is at \url{https://github.com/abdullah5523p/In-context-estimation-and-prediction-for-ODEs-study}. An archived snapshot of the code used to produce all reported results, together with the trained model weights and evaluation caches, is available at \url{https://doi.org/10.5281/zenodo.24263698}.

\subsection{Experimental setup}
\label{subsec:experiments-setup}

We consider four ODE families with known explicit solutions, summarized in Table~\ref{tab:ode-families}, with parameter coordinates sampled independently and uniformly over the intervals specified there. During practical training, we normalize the input parameters and state to a common scale. Observation times lie in $[0,3.2]$, while prediction queries lie in $[0,4]$. We use the exact ODE solutions for data generation.

\begin{table}[htbp]
\footnotesize
\caption{Parametric ODE families used in the experiments, with state dimension $m$ and parameter dimension $p$.}
\label{tab:ode-families}
\centering
\renewcommand{\arraystretch}{1.2}
\begin{tabular}{@{}llllc@{}}
\toprule
ODE & Equation & Parameter range & $(m,p)$ & Main test characteristic\\
\midrule
\cellstack{Forced linear\\(FL)}
 & \cellstack{$\dot x=-x+\omega_1+\omega_2t$\\$x(0)=x_0$}
 & $x_0,\omega_1,\omega_2\in[-0.8,0.8]$
 & $(1,3)$ & Convex\\
\midrule
\cellstack{Logistic growth\\(LG)}
 & \cellstack{$\dot x=rx(1-x/K)$\\$x(0)=x_0$}
 & \cellstack{$x_0\in[0.1,0.3]$, $r\in[0.6,1.8]$\\$K\in[0.8,1.6]$}
 & $(1,3)$ & Nonconvex\\
\midrule
\cellstack{Birth-death\\(BD)}
 & \cellstack{$\dot x=(b-d)x-cx^2$\\$x(0)=0.2$}
 & \cellstack{$b\in[0.8,1.8]$, $d\in[0.1,0.6]$\\$c\in[0.2,0.8]$}
 & $(1,3)$ & Non-identifiable\\
\midrule
\cellstack{Stuart-Landau\\(SL)}
 & \cellstack{$\dot x=(a-x^2-y^2)x-\omega y$\\$\dot y=\omega x+(a-x^2-y^2)y$\\$(x(0),y(0))=(x_0,y_0)$}
 & \cellstack{$x_0,y_0\in[0.4,0.6]$\\$a\in[0.8,1.2]$, $\omega\in[2,3]$}
 & $(2,4)$ & Higher dimension\\
\bottomrule
\end{tabular}
\end{table}

The FL system is our main example for extensive scaling studies and comparisons, since its solution
is linear in $\theta$, and thus Gaussian maximum likelihood reduces to a convex least-squares problem with tolerance conditions \eqref{eq:transformer-approximation-budgets} easily satisfied. This allows us to study how estimation and prediction accuracy scale with $M$ and $n$ without the additional difficulty of nonconvex optimization. LG serves as a non-convex example.

For BD, the trajectory depends on $b$ and $d$ only through their difference $b-d$. Thus, while $(b-d,c)$ is identifiable, the individual parameters $b$ and $d$ are not. We train the parameter transformer to estimate all three coordinates but evaluate recovery using the identifiable pair $(b-d,c)$. Although this setting falls outside our identifiability assumptions, it allows us to examine whether the transformer can recover identifiable parameter combinations when the original parameterization is non-identifiable. Trajectory prediction remains well-defined.

For SL, the dynamics has limit cycles as the radius approaches $\sqrt a$, while the phase rotates at rate $\omega$. It serves as a more difficult test due to oscillatory dynamics.

\medskip
\paragraph{Architecture and training loss}
Our default squared-ReLU transformer consists of 16 layers with token width 64, feedforward hidden width 256, and two full-width attention heads per layer, totaling approximately \(0.92\)M trainable parameters. We use variable hidden widths, pre-RMSNorm, residual connections, scaled attention, bias-free linear maps,and dropout of $0.1$ on residual updates. Additional architecture details and ablations are provided in Appendix~\ref{app:exp-architecture}.

The implementation is closely related to our theoretical architecture. Residual connections, scaled attention, bias-free linear maps, and variable hidden widths can be accommodated by the theoretical construction. The main additional operation at inference is RMSNorm, while dropout is used only during training. These architectural choices are motivated by their benefits for training stability~\cite{Vaswani2017,zhang2019rmsnorm,xiong2020layernorm}. Furthermore, we define
\begin{equation}
L_{\mathrm{mean}}
:=\frac{1}{d_z}
\left\|
\frac{1}{n}\sum_{i=1}^n\widehat z_i-z
\right\|_2^2
\le \frac{1}{nd_z}\sum_{i=1}^n\|\widehat z_i-z\|_2^2 =:L_{\mathrm{all}}.
\label{eq:exp-all-token}
\end{equation}
where $d_z$ is the matching dimension of the output tokens and the target $z$. We supervise the training using $L_{\mathrm{all}}$,
and unless otherwise specified, test errors are measured the same way. For comparisons with statistical estimators, where a single output is needed, we use $L_{\mathrm{mean}}$. Our experiments show that the output tokens are practically identical after training, so $L_{\mathrm{mean}}\approx L_{\mathrm{all}}$.
Although our main theoretical formulation uses a single-token position-based readout, it is easy to see that the same guarantees extend to supervision using either $L_{\mathrm{mean}}$ or $L_{\mathrm{all}}$. This is because both preserve the comparator error, bounded outputs, and Lipschitz properties needed for the pretraining analysis.

\subsection{Experimental results}

\begin{figure}[!tbp]
\centering
\includegraphics[width=\linewidth,trim=0bp 20.7bp 0bp 0bp,clip]{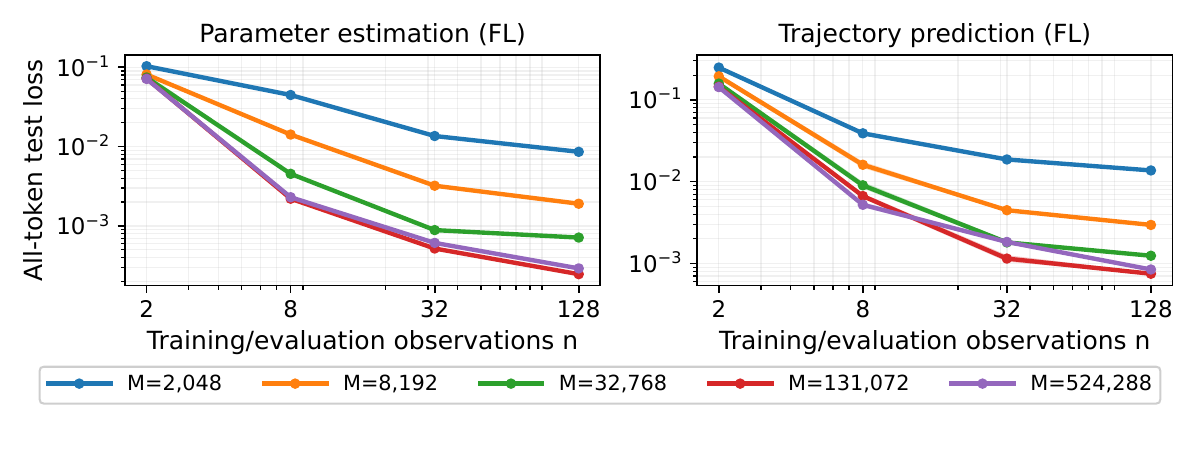}\par\medskip
\includegraphics[width=\linewidth,trim=0bp 20.7bp 0bp 0bp,clip]{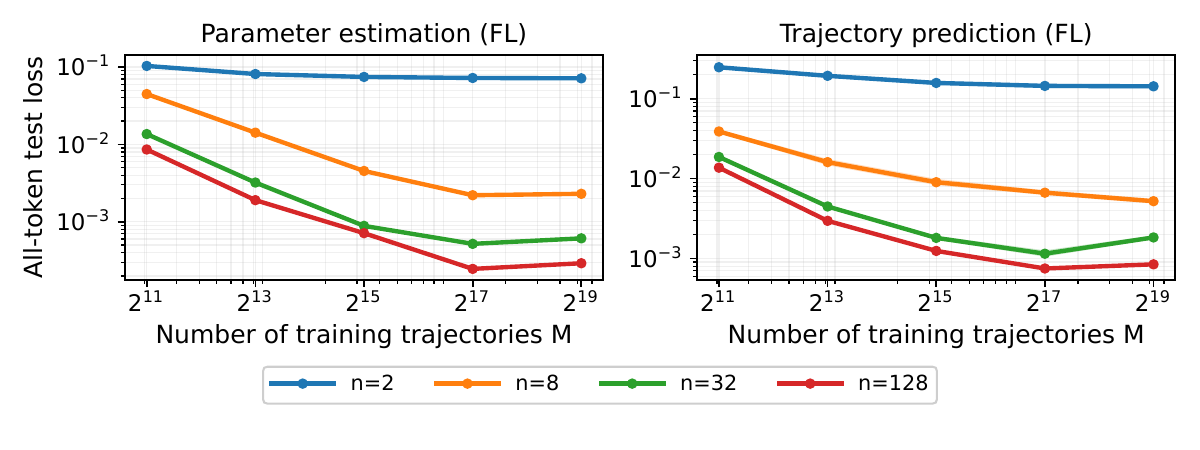}%
\caption{Matched train and inference context-size test error for parameter estimation and trajectory prediction in the FL system, plotted against $n$ (top) and against $M$ (bottom).}
\label{fig:exp-n}
\end{figure}

\medskip
\paragraph{Scaling with pretraining tasks and within-task observations}
\label{subsec:experiments-M}
For both parameter estimation and trajectory prediction, increasing either the number of within-task observations or the number of pretraining tasks generally seems to reduce the test error (Figure~\ref{fig:exp-n}). The curves eventually level off, suggesting that increasing $M$ or $n$ alone may not eliminate the remaining error for a fixed trained architecture. These trends are consistent with the error decomposition in our theory, although they do not isolate the individual error terms or establish the predicted rates. We provide descriptive fits of the scaling exponents with respect to $M$ in Appendix~\ref{app:experiments}. The case $n=2$ is particularly interesting, since there are fewer observations than unknown parameters. The relatively low estimation error in this setting suggests that the transformer benefits from information learned about the parameter prior during pretraining.

\begin{figure}[!tbp]
\centering
\includegraphics[width=\linewidth,trim=0bp 20.7bp 0bp 0bp,clip]{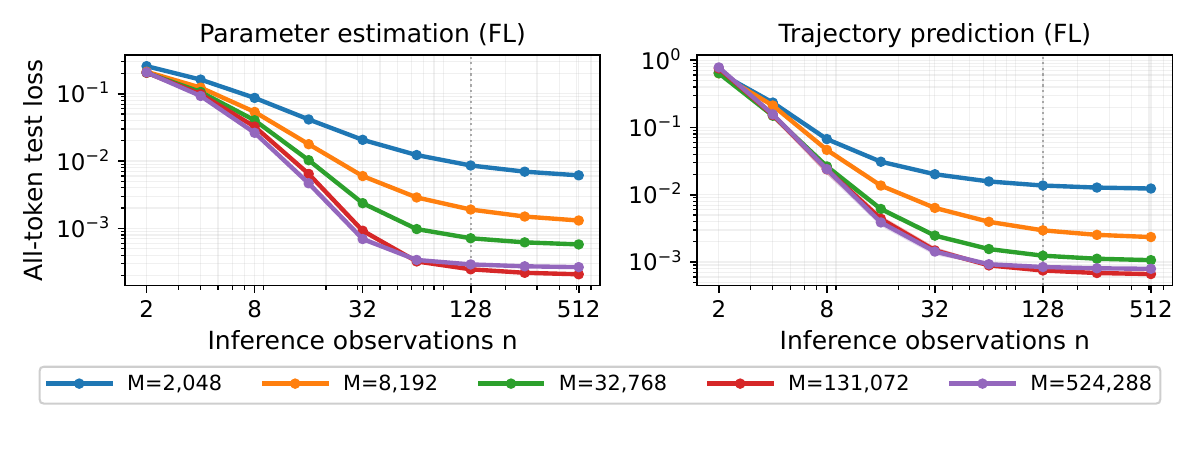}%
\caption{Test error versus inference context length for fixed networks trained at $n_{\mathrm{train}}=128$ and different values of $M$. The dotted line marks the training context.}
\label{fig:exp-parameter-training-n}
\end{figure}

\paragraph{Generalization across inference context lengths}
\label{subsec:experiments-training-n} 
Figure~\ref{fig:exp-parameter-training-n} shows that pretrained transformers trained at $n_{\mathrm{train}}=128$ can generalize to different inference context lengths beyond $n=128$ without adapting their weights. For both estimation and prediction, errors generally decrease as more inference observations become available, including beyond the training context length, before approaching a plateau. Increasing $M$ also tends to lower this plateau, suggesting that more extensive pretraining improves the ability to use additional observations at inference. While our realization theorem establishes the existence of transformers with such properties, it does not directly guarantee such generalization for empirically trained networks.

\begin{figure}[!tbp]
\centering
\includegraphics[width=\linewidth,trim=0bp 20.7bp 0bp 0bp,clip]{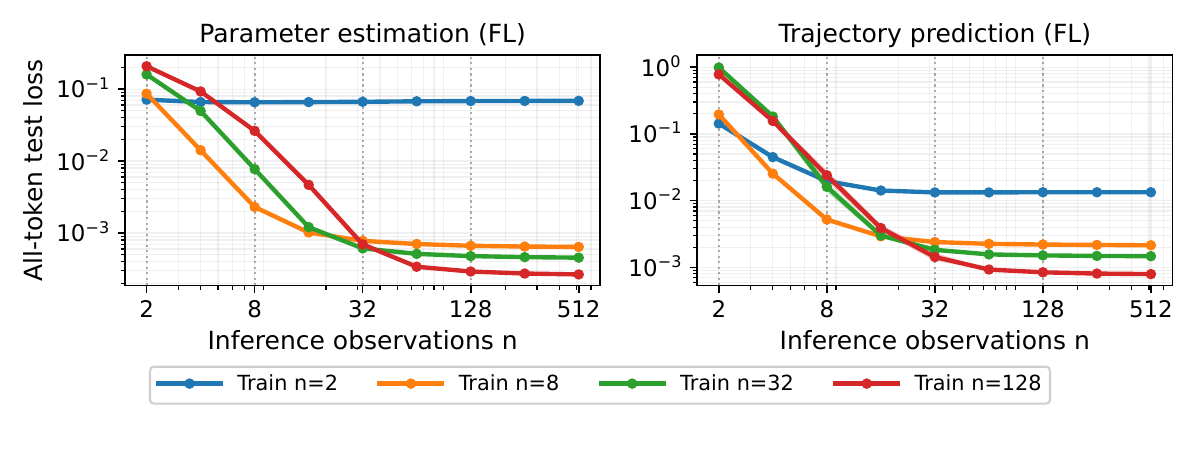}%
\caption{Inference-context scaling at fixed $M=2^{19}$ for networks trained at $n_{\mathrm{train}}\in\{2,8,32,128\}$.}
\label{fig:n}
\end{figure}

The choice of training context length also affects performance. At fixed $M=2^{19}$, Figure~\ref{fig:n} shows that networks trained with longer contexts generally perform better when many inference observations are available. That said, those trained at $n_{\mathrm{train}}=8$ can achieve lower errors for short inference contexts, while those trained at $n_{\mathrm{train}}=2$ stagnate. Thus, the training context length influences performance across different inference settings, even though the architecture can process variable-length inputs.

\medskip

\begin{figure}[!tbp]
\centering
\includegraphics[width=.95\linewidth,trim=0bp 20.2bp 0bp 0bp,clip]{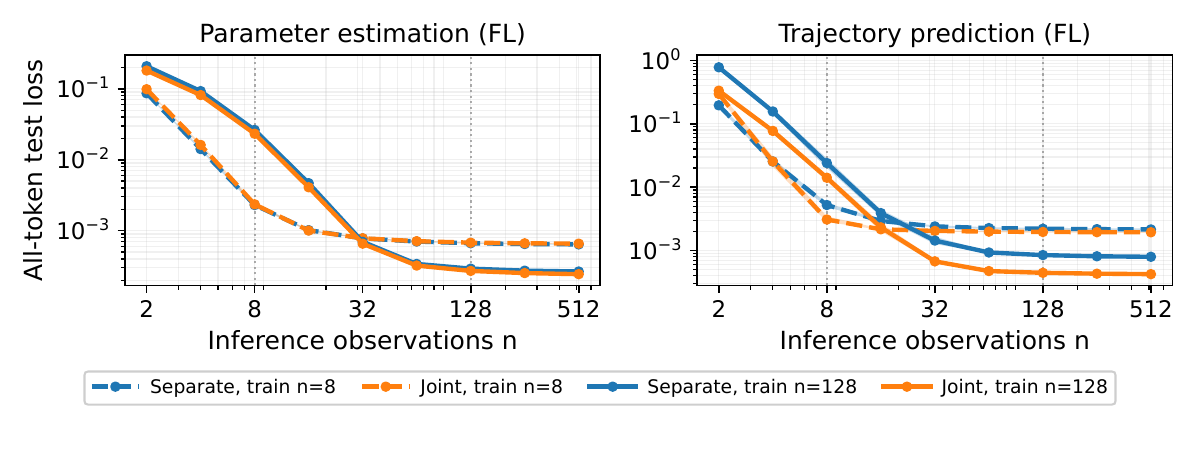}%
\caption{Comparison of joint and separately trained transformers for the forced-linear system at $M=2^{19}$ and $n_{\mathrm{train}}=8,128$.}
\label{fig:exp-forced-linear-joint}
\end{figure}

\paragraph{Joint estimation and prediction}
We also train a single transformer to perform parameter estimation and trajectory prediction simultaneously using the same-sized architecture with output dimension $p+m$ and all-token loss normalized accordingly. Since $p>m$ in our examples, the parameter estimation task contributes more output coordinates to the joint loss, giving it a larger aggregate weight when the per-coordinate errors are comparable.
Figure~\ref{fig:exp-forced-linear-joint} compares the joint transformer with separately trained models at $M=2^{19}$ and $n_{\mathrm{train}}\in\{8,128\}$. The joint model achieves comparable performance on both tasks without a substantial increase in model size. This is consistent with our theoretical construction, which shows that the two tasks can share the underlying parameter estimation computation rather than requiring separate networks. That said, the experiments do not identify the computations learned by the transformer, but provide empirical support for shared computation.

\medskip

\begin{figure}[!tbp]
\centering
\includegraphics[width=\linewidth,trim=0bp 25bp 0bp 0bp,clip]{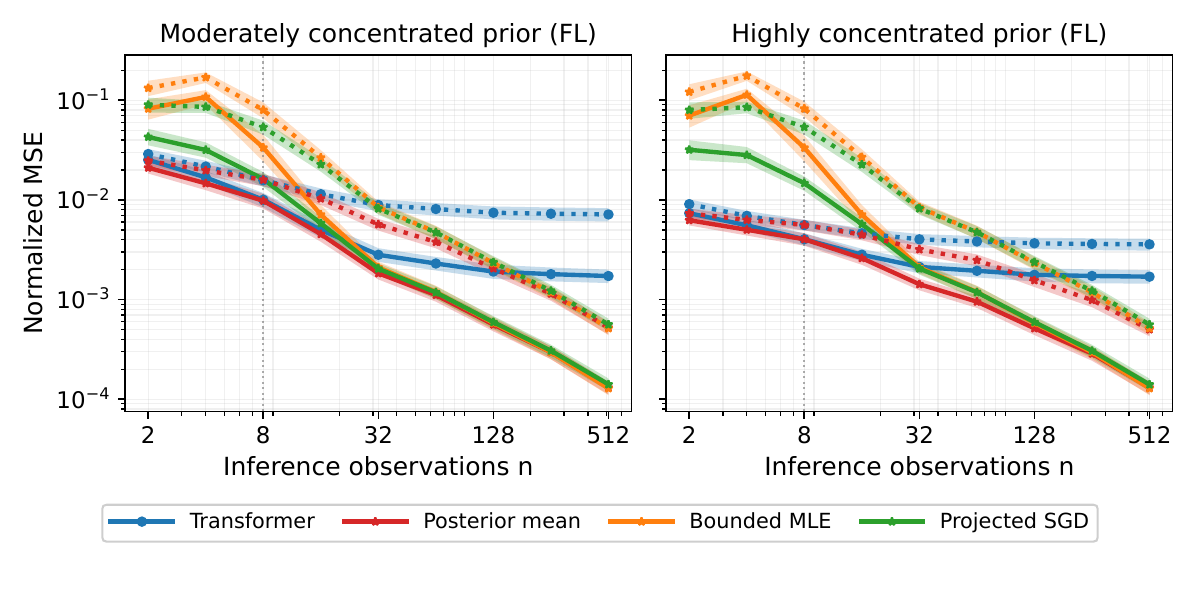}%
\caption{Comparison of parameter-only transformer with statistical estimators for a truncated Gaussian prior with increasing variance (left to right) and observation noise (solid to dotted lines).}
\label{fig:exp-statistics}
\end{figure}

\paragraph{Comparison with statistical estimators} We compare our parameter-only transformer trained at $M=2^{19}$ and $n_{\mathrm{train}}=8$ with classical statistical estimators for a non-trivial parameter prior, i.e., a truncated Gaussian. In particular, we include bounded MLE computed by linear least squares, projected SGD on the same exact trajectory-matching loss, and the posterior mean under the training prior. 
Figure~\ref{fig:exp-statistics} shows that near the training context length till around $n=16$, the transformer closely matches the posterior mean (as expected theoretically under sufficiently good training), and outperforms bounded MLE and projected SGD, illustrating the benefit of the training
prior when observations are limited. In particular, the difference in performance is larger for more concentrated priors, and increasing observation noise (solid to dotted lines). As the inference context grows, the trend reverses.

\medskip
\begin{figure}[!tbp]
\centering
\includegraphics[width=\linewidth,trim=0bp 5bp 0bp 0bp,clip]{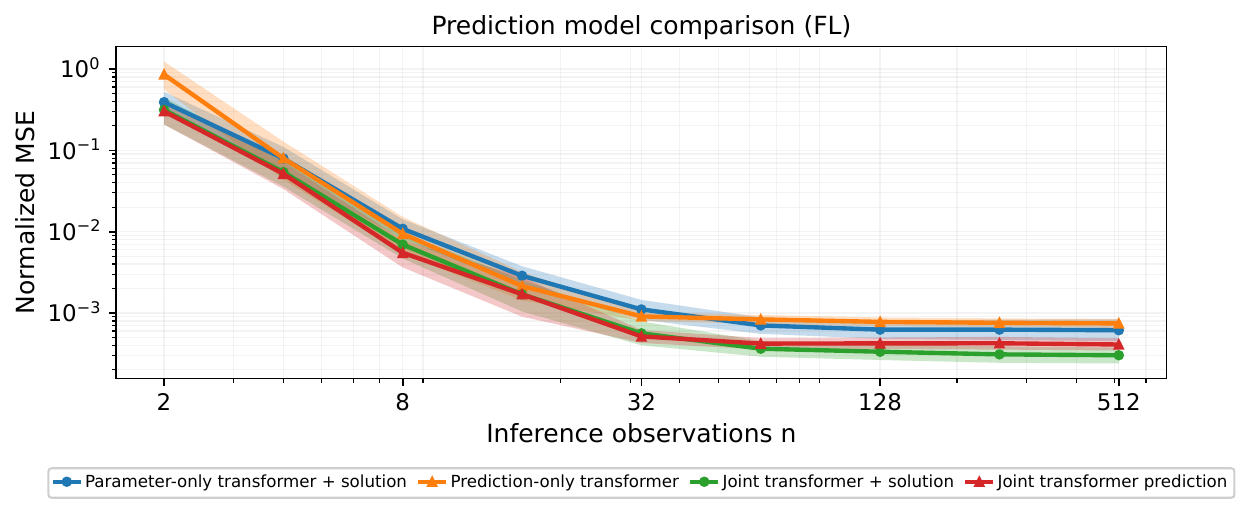}%
\caption{Comparison between transformer models for prediction tasks.}
\label{fig:exp-prediction}
\label{fig:prediction}
\end{figure}

\paragraph{Comparison between prediction approaches} For trajectory prediction, we compare direct transformer predictions with plug-in predictions obtained by first estimating the parameters and then evaluating the exact ODE solution at the query time. For the latter, we average the parameter transformer's output tokens and transform the estimate back to the original parameter scale before the plug-in. Direct prediction is also evaluated using the average token output. 
Figure~\ref{fig:exp-prediction} shows how different transformer approaches perform similarly across a range of inference context lengths. Direct transformer prediction generally has slightly higher error than the corresponding plug-in approach using the exact ODE solution. This is consistent with our theoretical error decomposition, where direct prediction involves an additional approximation error from the solution map. However, the comparison alone does not reveal if the transformers learned such a modular procedure.

\medskip
\paragraph{Results for the other ODE families}
Additional results for the logistic growth, birth-death, and Stuart-Landau systems are provided in Appendix~\ref{app:exp-statistics}, including inference-context scaling, comparisons with statistical estimators, and conditional-variance analyses. The benefits of additional inference observations generally persist, although parameter estimation can be more challenging than in the forced-linear case. In particular, the birth-death example illustrates the distinction between recovering identifiable parameter combinations and predicting trajectories when the original parameters are not uniquely determined. Since these ODE families differ in several respects, the experiments are not intended to isolate the individual effects of nonconvexity or identifiability.

\section{Discussion}
\label{sec:conclusion}
In this paper, we studied in-context parameter estimation and trajectory prediction for parametric ODE families, distinguishing statistical estimation within a new task from learning a shared inference procedure across tasks. We showed that a transformer can implement a two-stage approximate maximum likelihood procedure, combining a coarse global search with local gradient descent, and established nonasymptotic parameter recovery guarantees from noisy observations. We further derived pretraining risk bounds that separate within-task estimation and approximation errors from errors due to learning across finitely many tasks. Compared with existing studies of transformers emulating optimization algorithms, our analysis extends this perspective to nonconvex inverse problems, where a suitable initialization is needed before local optimization. For trajectory prediction, our construction provides an interpretable approach that first recovers the governing parameters and then evaluates the corresponding ODE solution. Moreover, by exploiting the shared computations between estimation and prediction, we show that a single transformer can perform both tasks with essentially the same order of model complexity as prediction alone. Together, these results suggest a possible route toward foundation models for physical systems that incorporate the underlying dynamics into prediction, rather than relying solely on black-box mappings~\cite{bonavita2024limitations,banomedina2025physics,selz2026resolution,chu2026generalizable,craig2026physics}.

One question for future work is how the theoretical construction relates to the behavior of transformers learned through pretraining. Although our constructed transformer can operate across different context lengths, our pretraining guarantees are established for a fixed number of within-task observations $n$. Our numerical experiments suggest that transformers trained at a fixed context length can nevertheless benefit from additional observations at inference, even beyond the training context. Understanding how this ability emerges through pretraining, and whether the learned transformers implement inference procedures resembling our construction, would help connect the theory with empirical observations. Another direction is to develop more efficient initialization procedures, as the finite parameter net used in our construction can become computationally expensive in high dimensions.

The framework could also be extended to settings beyond the ODE families considered here. These include partially observed systems, more general observation noise, stochastic dynamics, transfer to previously unseen dynamical systems or equation discovery. Moreover, fully nonparametric learning of unknown governing equations~\cite{heinonen2018,schoetz2026} could be another challenge. In particular, stochastic dynamics may benefit from the modular nature of our construction, where parameter estimation and subsequent prediction are treated as separate steps. Another direction is to study non-identifiable systems, where the goal may be to recover identifiable parameter combinations or equivalence classes rather than the full parameter vector. Our birth--death experiments provide a preliminary illustration of this setting, while a corresponding theoretical analysis remains for future work.

\bibliographystyle{plainnat}
\bibliography{ICL_ODE}

\begin{thebibliography}{43}
\providecommand{\natexlab}[1]{#1}
\providecommand{\url}[1]{\texttt{#1}}
\expandafter\ifx\csname urlstyle\endcsname\relax
  \providecommand{\doi}[1]{doi: #1}\else
  \providecommand{\doi}{doi: \begingroup \urlstyle{rm}\Url}\fi

\bibitem[Aky{\"u}rek et~al.(2023)Aky{\"u}rek, Schuurmans, Andreas, Ma, and Zhou]{akyurek2023}
E.~Aky{\"u}rek, D.~Schuurmans, J.~Andreas, T.~Ma, and D.~Zhou.
\newblock What learning algorithm is in-context learning? {I}nvestigations with linear models.
\newblock In \emph{International Conference on Learning Representations}, 2023.

\bibitem[Bai et~al.(2023)Bai, Chen, Wang, Xiong, and Mei]{bai2023}
Y.~Bai, F.~Chen, H.~Wang, C.~Xiong, and S.~Mei.
\newblock Transformers as statisticians: Provable in-context learning with in-context algorithm selection.
\newblock In \emph{Advances in Neural Information Processing Systems}, 2023.

\bibitem[Ba{\~n}o-Medina et~al.(2025)Ba{\~n}o-Medina, Sengupta, Doyle, Reynolds, Watson-Parris, and Delle~Monache]{banomedina2025physics}
J.~Ba{\~n}o-Medina, A.~Sengupta, J.~D. Doyle, C.~A. Reynolds, D.~Watson-Parris, and L.~Delle~Monache.
\newblock Are {AI} weather models learning atmospheric physics? {A} sensitivity analysis of cyclone {X}ynthia.
\newblock \emph{npj Climate and Atmospheric Science}, 8:\penalty0 92, 2025.
\newblock \doi{10.1038/s41612-025-00949-6}.

\bibitem[Bates and Watts(1988)]{bates1988}
D.~M. Bates and D.~G. Watts.
\newblock \emph{Nonlinear Regression Analysis and Its Applications}.
\newblock Wiley, 1988.

\bibitem[Becker et~al.(2023)Becker, Klein, Neitz, Parascandolo, and Kilbertus]{becker2023predicting}
S.~Becker, M.~Klein, A.~Neitz, G.~Parascandolo, and N.~Kilbertus.
\newblock Predicting ordinary differential equations with transformers.
\newblock In \emph{International Conference on Machine Learning}, volume 202 of \emph{Proceedings of Machine Learning Research}, pages 1978--2002, 2023.
\newblock URL \url{https://proceedings.mlr.press/v202/becker23a.html}.

\bibitem[Bonavita(2024)]{bonavita2024limitations}
M.~Bonavita.
\newblock On some limitations of current machine learning weather prediction models.
\newblock \emph{Geophysical Research Letters}, 51:\penalty0 e2023GL107377, 2024.
\newblock \doi{10.1029/2023GL107377}.

\bibitem[Boucheron et~al.(2013)Boucheron, Lugosi, and Massart]{boucheron2013}
S.~Boucheron, G.~Lugosi, and P.~Massart.
\newblock \emph{Concentration Inequalities: A Nonasymptotic Theory of Independence}.
\newblock Oxford University Press, 2013.

\bibitem[Brown et~al.(2020)Brown, Mann, Ryder, Subbiah, Kaplan, Dhariwal, Neelakantan, Shyam, Sastry, Askell, Agarwal, Herbert-Voss, Krueger, Henighan, Child, Ramesh, Ziegler, Wu, Winter, Hesse, Chen, Sigler, Litwin, Gray, Chess, Clark, Berner, McCandlish, Radford, Sutskever, and Amodei]{brown2020}
T.~B. Brown, B.~Mann, N.~Ryder, M.~Subbiah, J.~Kaplan, P.~Dhariwal, A.~Neelakantan, P.~Shyam, G.~Sastry, A.~Askell, S.~Agarwal, A.~Herbert-Voss, G.~Krueger, T.~Henighan, R.~Child, A.~Ramesh, D.~M. Ziegler, J.~Wu, C.~Winter, C.~Hesse, M.~Chen, E.~Sigler, M.~Litwin, S.~Gray, B.~Chess, J.~Clark, C.~Berner, S.~McCandlish, A.~Radford, I.~Sutskever, and D.~Amodei.
\newblock Language models are few-shot learners.
\newblock In \emph{Advances in Neural Information Processing Systems}, volume~33, pages 1877--1901, 2020.

\bibitem[Brunel(2008)]{brunel2008parameter}
N.~J.-B. Brunel.
\newblock Parameter estimation of {ODE}'s via nonparametric estimators.
\newblock \emph{Electronic Journal of Statistics}, 2:\penalty0 1242--1267, 2008.
\newblock \doi{10.1214/07-EJS132}.

\bibitem[Cheng et~al.(2024)Cheng, Chen, and Sra]{cheng2024functional}
X.~Cheng, Y.~Chen, and S.~Sra.
\newblock Transformers implement functional gradient descent to learn non-linear functions in context.
\newblock In \emph{International Conference on Machine Learning}, volume 235 of \emph{Proceedings of Machine Learning Research}, pages 8002--8037, 2024.
\newblock URL \url{https://proceedings.mlr.press/v235/cheng24a.html}.

\bibitem[Ching et~al.(2026)Ching, Popescu, Smith, Ma, Underwood, and Samworth]{ching2026nonparametric}
M.~Ching, I.~Popescu, N.~Smith, T.~Ma, W.~G. Underwood, and R.~J. Samworth.
\newblock Efficient and minimax-optimal in-context nonparametric regression with transformers.
\newblock arXiv preprint arXiv:2601.15014, 2026.
\newblock URL \url{https://arxiv.org/abs/2601.15014}.

\bibitem[Chu et~al.(2026)Chu, Liu, Biswas, and Shen]{chu2026generalizable}
M.~Chu, Y.~Liu, A.~Biswas, and H.-W. Shen.
\newblock Do physics foundation models learn generalizable physics? {A} bias-aware benchmark across physical regimes and distribution shifts.
\newblock arXiv preprint arXiv:2605.29283, 2026.
\newblock URL \url{https://arxiv.org/abs/2605.29283}.

\bibitem[Craig et~al.(2026)Craig, Selz, Beylich, and Tempest]{craig2026physics}
G.~Craig, T.~Selz, M.~Beylich, and K.~I. Tempest.
\newblock The physics of {AI} weather models.
\newblock arXiv preprint arXiv:2605.23778, 2026.
\newblock URL \url{https://arxiv.org/abs/2605.23778}.

\bibitem[d'Ascoli et~al.(2024)d'Ascoli, Becker, Mathis, Schwaller, and Kilbertus]{dascoli2024}
S.~d'Ascoli, S.~Becker, A.~Mathis, P.~Schwaller, and N.~Kilbertus.
\newblock {ODEFormer}: Symbolic regression of dynamical systems with transformers.
\newblock In \emph{International Conference on Learning Representations}, 2024.

\bibitem[Furuya et~al.(2025)Furuya, de~Hoop, and Peyr{\'e}]{furuya2024universal}
T.~Furuya, M.~V. de~Hoop, and G.~Peyr{\'e}.
\newblock Transformers are universal in-context learners.
\newblock In \emph{International Conference on Learning Representations}, 2025.
\newblock URL \url{https://arxiv.org/abs/2408.01367}.

\bibitem[Garg et~al.(2022)Garg, Tsipras, Liang, and Valiant]{garg2022}
S.~Garg, D.~Tsipras, P.~Liang, and G.~Valiant.
\newblock What can transformers learn in-context? {A} case study of simple function classes.
\newblock In \emph{Advances in Neural Information Processing Systems}, 2022.

\bibitem[Goel and Bartlett(2024)]{goel2024kalman}
G.~Goel and P.~Bartlett.
\newblock Can a transformer represent a {K}alman filter?
\newblock In \emph{Proceedings of the 6th Annual Learning for Dynamics and Control Conference}, volume 242 of \emph{Proceedings of Machine Learning Research}, pages 1502--1512, 2024.
\newblock URL \url{https://proceedings.mlr.press/v242/goel24a.html}.

\bibitem[Grohs et~al.(2019)Grohs, Hornung, Jentzen, and Zimmermann]{grohs2019spacetime}
P.~Grohs, F.~Hornung, A.~Jentzen, and P.~Zimmermann.
\newblock Space-time error estimates for deep neural network approximations for differential equations.
\newblock arXiv preprint arXiv:1908.03833, 2019.
\newblock URL \url{https://arxiv.org/abs/1908.03833}.

\bibitem[Heinonen et~al.(2018)Heinonen, Yildiz, Mannerstr{\"o}m, Intosalmi, and L{\"a}hdesm{\"a}ki]{heinonen2018}
M.~Heinonen, C.~Yildiz, H.~Mannerstr{\"o}m, J.~Intosalmi, and H.~L{\"a}hdesm{\"a}ki.
\newblock Learning unknown {ODE} models with {G}aussian processes.
\newblock In \emph{Proceedings of the 35th International Conference on Machine Learning}, volume~80, pages 1959--1968, 2018.

\bibitem[Herde et~al.(2024)Herde, Raoni{\'c}, Rohner, K{\"a}ppeli, Molinaro, de~B{\'e}zenac, and Mishra]{herde2024poseidon}
M.~Herde, B.~Raoni{\'c}, T.~Rohner, R.~K{\"a}ppeli, R.~Molinaro, E.~de~B{\'e}zenac, and S.~Mishra.
\newblock {Poseidon}: Efficient foundation models for {PDE}s.
\newblock In \emph{Advances in Neural Information Processing Systems}, volume~37, 2024.
\newblock URL \url{https://arxiv.org/abs/2405.19101}.

\bibitem[Hirsch et~al.(2013)Hirsch, Smale, and Devaney]{hirsch2013differential}
M.~W. Hirsch, S.~Smale, and R.~L. Devaney.
\newblock \emph{Differential Equations, Dynamical Systems, and an Introduction to Chaos}.
\newblock Academic Press, 3 edition, 2013.

\bibitem[Laurent and Massart(2000)]{laurent2000}
B.~Laurent and P.~Massart.
\newblock Adaptive estimation of a quadratic functional by model selection.
\newblock \emph{Annals of Statistics}, 28\penalty0 (5):\penalty0 1302--1338, 2000.

\bibitem[Li et~al.(2020)Li, Tang, and Yu]{li2020powernet}
B.~Li, S.~Tang, and H.~Yu.
\newblock {PowerNet}: Efficient representations of polynomials and smooth functions by deep neural networks with rectified power units.
\newblock \emph{Journal of Mathematical Study}, 53\penalty0 (2):\penalty0 159--191, 2020.
\newblock \doi{10.4208/jms.v53n2.20.03}.

\bibitem[Li et~al.(2025)Li, Jiao, Huang, Wei, and Chen]{li2025icl}
G.~Li, Y.~Jiao, Y.~Huang, Y.~Wei, and Y.~Chen.
\newblock Transformers meet in-context learning: A universal approximation theory.
\newblock arXiv preprint arXiv:2506.05200, 2025.

\bibitem[Li et~al.(2021)Li, Kovachki, Azizzadenesheli, Liu, Bhattacharya, Stuart, and Anandkumar]{li2021fno}
Z.~Li, N.~Kovachki, K.~Azizzadenesheli, B.~Liu, K.~Bhattacharya, A.~Stuart, and A.~Anandkumar.
\newblock Fourier neural operator for parametric partial differential equations.
\newblock In \emph{International Conference on Learning Representations}, 2021.
\newblock URL \url{https://arxiv.org/abs/2010.08895}.

\bibitem[Lu et~al.(2021)Lu, Jin, Pang, Zhang, and Karniadakis]{lu2021deeponet}
L.~Lu, P.~Jin, G.~Pang, Z.~Zhang, and G.~E. Karniadakis.
\newblock Learning nonlinear operators via {DeepONet} based on the universal approximation theorem of operators.
\newblock \emph{Nature Machine Intelligence}, 3\penalty0 (3):\penalty0 218--229, 2021.
\newblock \doi{10.1038/s42256-021-00302-5}.

\bibitem[Mauel et~al.(2026)Mauel, H{\"u}bers, Berghaus, Seifner, and Sanchez]{mauel2026fimode}
M.~Mauel, J.~R. H{\"u}bers, D.~Berghaus, P.~Seifner, and R.~J. Sanchez.
\newblock Foundation inference models for ordinary differential equations.
\newblock In \emph{International Conference on Machine Learning}, 2026.
\newblock URL \url{https://arxiv.org/abs/2602.08733}.

\bibitem[Mei et~al.(2018)Mei, Bai, and Montanari]{mei2018}
S.~Mei, Y.~Bai, and A.~Montanari.
\newblock The landscape of empirical risk for nonconvex losses.
\newblock \emph{Annals of Statistics}, 46\penalty0 (6A):\penalty0 2747--2774, 2018.

\bibitem[Ramsay et~al.(2007)Ramsay, Hooker, Campbell, and Cao]{ramsay2007}
J.~O. Ramsay, G.~Hooker, D.~Campbell, and J.~Cao.
\newblock Parameter estimation for differential equations: A generalized smoothing approach.
\newblock \emph{Journal of the Royal Statistical Society: Series B}, 69\penalty0 (5):\penalty0 741--796, 2007.

\bibitem[Sch{\"o}tz(2026)]{schoetz2026}
C.~Sch{\"o}tz.
\newblock Nonparametric estimation of ordinary differential equations: Snake and stubble.
\newblock \emph{Bernoulli}, 32\penalty0 (3):\penalty0 1880--1906, 2026.
\newblock \doi{10.3150/25-BEJ1936}.

\bibitem[Seifner et~al.(2025)Seifner, Cvejoski, K{\"o}rner, and Sanchez]{seifner2025fim}
P.~Seifner, K.~Cvejoski, A.~K{\"o}rner, and R.~J. Sanchez.
\newblock Zero-shot imputation with foundation inference models for dynamical systems.
\newblock In \emph{International Conference on Learning Representations}, 2025.
\newblock URL \url{https://proceedings.iclr.cc/paper_files/paper/2025/file/3d03791377a31cb3e7357014ba58eb80-Paper-Conference.pdf}.

\bibitem[Selz et~al.(2026)Selz, Allen, Bruinsma, Craig, Markou, and Turner]{selz2026resolution}
T.~Selz, A.~Allen, W.~P. Bruinsma, G.~C. Craig, S.~Markou, and R.~E. Turner.
\newblock On the effective resolution of {AI} weather prediction models.
\newblock \emph{Artificial Intelligence for the Earth Systems}, 5\penalty0 (4), 2026.
\newblock \doi{10.1175/AIES-D-25-0121.1}.

\bibitem[Shen et~al.(2026)Shen, Hsu, Lai, and Liao]{shen2026manifolds}
Z.~Shen, A.~Hsu, R.~Lai, and W.~Liao.
\newblock Understanding in-context learning on structured manifolds: Bridging attention to kernel methods.
\newblock In \emph{International Conference on Learning Representations}, 2026.
\newblock URL \url{https://proceedings.iclr.cc/paper_files/paper/2026/hash/45e9827fcd83cfb39285669e13b2a30d-Abstract-Conference.html}.

\bibitem[Shi et~al.(2026)Shi, Lai, Cloninger, and Liao]{shi2026crosstask}
Z.~Shi, R.~Lai, A.~Cloninger, and W.~Liao.
\newblock Transformers as cross-task learners: Shared structure drives sample efficiency in in-context learning.
\newblock arXiv preprint arXiv:2609.29060, 2026.
\newblock URL \url{https://arxiv.org/abs/2609.29060}.

\bibitem[Strogatz(2015)]{strogatz2015nonlinear}
S.~H. Strogatz.
\newblock \emph{Nonlinear Dynamics and Chaos: With Applications to Physics, Biology, Chemistry, and Engineering}.
\newblock Westview Press, 2 edition, 2015.

\bibitem[Tarantola(2005)]{tarantola2005}
A.~Tarantola.
\newblock \emph{Inverse Problem Theory and Methods for Model Parameter Estimation}.
\newblock SIAM, 2005.

\bibitem[Vaswani et~al.(2017)Vaswani, Shazeer, Parmar, Uszkoreit, Jones, Gomez, Kaiser, and Polosukhin]{Vaswani2017}
A.~Vaswani, N.~Shazeer, N.~Parmar, J.~Uszkoreit, L.~Jones, A.~N. Gomez, L.~Kaiser, and I.~Polosukhin.
\newblock Attention is all you need.
\newblock In \emph{Advances in Neural Information Processing Systems}, 2017.

\bibitem[von Oswald et~al.(2023)von Oswald, Niklasson, Randazzo, Sacramento, Mordvintsev, Zhmoginov, and Vladymyrov]{vonoswald2023}
J.~von Oswald, E.~Niklasson, E.~Randazzo, J.~Sacramento, A.~Mordvintsev, A.~Zhmoginov, and M.~Vladymyrov.
\newblock Transformers learn in-context by gradient descent.
\newblock In \emph{International Conference on Machine Learning}, 2023.

\bibitem[Wainwright(2019)]{wainwright2019}
M.~J. Wainwright.
\newblock \emph{High-Dimensional Statistics: A Non-Asymptotic Viewpoint}.
\newblock Cambridge University Press, 2019.

\bibitem[Xiong et~al.(2020)Xiong, Yang, He, Zheng, Zheng, Xing, Zhang, Lan, Wang, and Liu]{xiong2020layernorm}
Ruibin Xiong, Yunchang Yang, Di~He, Kai Zheng, Shuxin Zheng, Chen Xing, Huishuai Zhang, Yanyan Lan, Liwei Wang, and Tieyan Liu.
\newblock On layer normalization in the transformer architecture.
\newblock In \emph{Proceedings of the 37th International Conference on Machine Learning}, volume 119 of \emph{Proceedings of Machine Learning Research}, pages 10524--10533. PMLR, 2020.
\newblock URL \url{https://proceedings.mlr.press/v119/xiong20b.html}.

\bibitem[Yang et~al.(2023)Yang, Liu, Meng, and Osher]{yang2023icon}
L.~Yang, S.~Liu, T.~Meng, and S.~J. Osher.
\newblock In-context operator learning with data prompts for differential equation problems.
\newblock \emph{Proceedings of the National Academy of Sciences}, 120\penalty0 (39):\penalty0 e2310142120, 2023.
\newblock \doi{10.1073/pnas.2310142120}.

\bibitem[Yarotsky(2017)]{yarotsky2017}
D.~Yarotsky.
\newblock Error bounds for approximations with deep {ReLU} networks.
\newblock \emph{Neural Networks}, 94:\penalty0 103--114, 2017.

\bibitem[Zhang and Sennrich(2019)]{zhang2019rmsnorm}
Biao Zhang and Rico Sennrich.
\newblock Root mean square layer normalization.
\newblock In \emph{Advances in Neural Information Processing Systems}, volume~32, 2019.
\newblock URL \url{https://arxiv.org/abs/1910.07467}.

\end{thebibliography}

\clearpage
\appendix

\begin{center}
    {\LARGE\bfseries Appendices for ``In-Context Estimation and Prediction for Parametric ODEs''\par}
\end{center}

\paragraph{Overview} This appendix contains the notation, proofs, and additional experiments supporting the main text. Appendix~\ref{app:notation} collects notation and definitions relevant for the proofs, and Appendix~\ref{app:background} states the fundamental concentration and localization results used for the within-task parameter estimation and pretraining proofs, along with descriptions of the architectural components used in the main existence results. Appendix~\ref{app:statistical} proves the statistical parameter-recovery guarantee, covering local strong convexity and separability, concentration and approximation bounds, and the coarse-initialization plus gradient-descent analysis. Appendix~\ref{app:transformer-approximation} shows that transformers can realize the parameter-recovery, prediction, and joint estimators, and also give an Euler-based construction of trajectory and sensitivity networks. Appendix~\ref{app:pretraining} converts these realization results into the pretraining guarantees for parameter estimation, prediction, and the joint model. Appendix~\ref{app:experiments} gives further experimental details: the setup and normalization, architecture ablations (Appendix~\ref{app:exp-architecture}), results and statistical comparisons for the other ODE families (Appendix~\ref{app:exp-statistics}), query-time dependence of the joint model (Appendix~\ref{app:joint-query-sweep}), initial-state recovery (Appendix~\ref{app:initial-state-query}), and descriptive pretraining-risk scaling fits (Appendix~\ref{app:pretraining-risk-fits}). Readers interested only in the proofs can go directly to Appendices~\ref{app:statistical}--\ref{app:pretraining}; readers interested in the empirical results can go directly to Appendix~\ref{app:experiments}.

\section*{Appendix Contents}
\addcontentsline{toc}{section}{Appendix Contents}

\appendixentry{app:notation}{Notations and definitions}
\appendixentry{app:background}{Background Results}
\appendixentry{app:statistical}{Proofs for Statistical Parameter Recovery}
\appendixentry{app:transformer-approximation}{Proofs for Transformer-based Approximation}
\appendixentry{app:pretraining}{Proofs for the Pretraining Guarantee}
\appendixentry{app:experiments}{Additional Experimental Details}

\bigskip

\section{Notations and definitions}
\label{app:notation}

For \(g:[0,T]\to\R^m\), write
\begin{equation}
    \norm{g}_n^2
    :=\frac1n\sum_{i=1}^n\norm{g(t_i)}_2^2,
    \qquad
    \norm{g}_{L^2}^2
    :=\E_t\norm{g(t)}_2^2
    =\int_0^T\norm{g(t)}_2^2\,\nu(dt).
    \label{eq:l2-norms}
\end{equation}
For a centered scalar random variable \(Z\), its sub-Gaussian norm is
\[
    \norm{Z}_{\mathrm{SG}}
    :=\inf\left\{s>0:
       \E e^{\lambda Z}\le e^{s^2\lambda^2/2}
       \text{ for every }\lambda\in\R\right\}.
\]
For a function class \(\mathcal F\) that approximates the true trajectory, define its localized empirical ball
and empirical covering number by
\[
    B_n(\delta;\mathcal F)
    :=\{f\in\mathcal F:\norm{f}_n\le\delta\},
    \qquad
    N_n(u;A)
    :=\mathcal N(A,u;\norm{\cdot}_n).
\]
For a class \(\mathcal F\) of functions \(g:[0,T]\to\R^m\), its star hull
is
\[
    \operatorname{star}(\mathcal F)
    :=\{\beta g:g\in\mathcal F,\ \beta\in[0,1]\}.
\]
For independent \(w_i\sim N(0,I_m)\), conditionally on the observation
times, define the localized vector Gaussian complexity
\[
    \mathfrak{G}_n(\delta;\mathcal F)
    :=\E_w\sup_{\substack{g\in\mathcal F\\\norm g_n\le\delta}}
      \left|\frac1n\sum_{i=1}^n\langle w_i,g(t_i)\rangle\right|,
\]
and, for independent Rademacher variables
\(\varepsilon_1,\ldots,\varepsilon_n\) independent of the observation times,
the localized vector empirical Rademacher complexity
\[
    \widehat{\mathfrak{R}}_n^{\mathrm{vec}}(\delta;\mathcal F)
    :=\E_{\varepsilon}\sup_{\substack{g\in\mathcal F\\\norm g_n\le\delta}}
      \left|\frac1n\sum_{i=1}^n\varepsilon_i\norm{g(t_i)}_2\right|.
\]
The corresponding population-level Rademacher complexity is
\[
    \mathfrak{R}_n^{\mathrm{vec}}(\delta;\mathcal F)
    :=\E_{\varepsilon,t}\sup_{\substack{g\in\mathcal F\\\norm g_{L^2}\le\delta}}
      \left|\frac1n\sum_{i=1}^n\varepsilon_i\norm{g(t_i)}_2\right|.
\]

\section{Background Results}
\label{app:background}

This section states the external results used in the concentration
proofs, in the precise forms needed below, together with the two
vector-valued localized least-squares lemmas derived from them.

\begin{lemma}[Sub-Gaussian process concentration]
\label{lem:bg-subgaussian-process}
Let \(\Theta\subset\R^p\) have diameter \(D\), and let
\(\{Z_i(\theta):\theta\in\Theta\}_{i=1}^n\) be independent copies of a
centered process satisfying
\[
 \norm{Z_i(\theta)}_{\mathrm{SG}}\le B_0,\qquad
 \norm{Z_i(\theta)-Z_i(\theta')}_{\mathrm{SG}}
 \le B_1\norm{\theta-\theta'}_2.
\]
Then a universal \(C>0\) exists such that, with probability at least
\(1-\delta\),
\[
 \sup_{\theta\in\Theta}\left|\frac1n\sum_{i=1}^nZ_i(\theta)\right|
 \le CB_0\sqrt{\frac{p\log(2+2B_1D/B_0)+\log(1/\delta)}n}.
\]
\end{lemma}

\begin{proof}
This is the standard finite-dimensional sub-Gaussian chaining bound. It
uses the Euclidean covering estimate
\[
 \mathcal N(\Theta,u;\norm{\cdot}_2)\le(1+2D/u)^p;
\]
see Wainwright~\cite[Theorem~5.36]{wainwright2019}.
\end{proof}

\begin{lemma}[Laurent--Massart inequality]
\label{lem:bg-laurent-massart}
If \(Z\sim\chi_k^2\), then for every \(u>0\),
\[
 \Pp(Z-k\ge2\sqrt{ku}+2u)\le e^{-u},\qquad
 \Pp(k-Z\ge2\sqrt{ku})\le e^{-u}.
\]
\end{lemma}

\begin{proof}
This is the chi-square deviation inequality of Laurent and
Massart~\cite{laurent2000}.
\end{proof}

\begin{lemma}[Uniform derivative concentration]
\label{lem:bg-mei-concentration}
Let \(\ell(\theta;Z)\) be twice differentiable on a compact ball
\(\Theta\subset\R^p\) of diameter \(D\) centered at the origin. Suppose, for some \(\tau\ge1\),
the centered gradient is sub-Gaussian with scale \(\tau\), every centered
directional Hessian
\(u^\top(\nabla^2\ell-\E\nabla^2\ell)u\), \(\norm u_2=1\), is
sub-exponential with scale \(\tau^2\), and
\[
 \norm{\nabla^2\ell(\theta;Z)-\nabla^2\ell(\theta';Z)}_{\mathrm{op}}
 \le L(Z)\norm{\theta-\theta'}_2,\qquad \E L(Z)\le\tau^3.
\]
Assume also that \(\|\nabla^2\E \ell(\theta_0;Z)\|_{\mathrm{op}}\leq \tau^2\)
at some \(\theta_0\in\Theta\).
Then there is a universal \(C_0>0\) such that, with
\(C_\delta=C_0\left(1+\log\frac{1+D\tau}{2\delta}\right)\), whenever
\(n\ge \max\{2,C_\delta p\log p\}\), with probability at least \(1-2\delta\),
\[
 \sup_{\theta\in\Theta}\norm{\nabla\frac{1}{n} \sum_{i=1}^n \ell(\theta;Z_i) -\nabla \E \ell(\theta;Z)}_2
 \le\tau\sqrt{\frac{C_\delta p\log n}{n}},
\]
and
\[
 \sup_{\theta\in\Theta}
 \norm{\nabla^2\frac{1}{n}\sum_{i=1}^n \ell(\theta;Z_i) -\nabla^2\E \ell(\theta;Z)}_{\mathrm{op}}
 \le\tau^2\sqrt{\frac{C_\delta p\log n}{n}}.
\]
\end{lemma}
\begin{proof}
This is the \(c_h=0\) specialization of
Mei, Bai, and Montanari~\cite[Theorem~1]{mei2018}. 
\end{proof}

\begin{lemma}[Vector-valued oracle inequality]
\label{lem:vector-oracle}
Let \(\mathcal X\) be a class of functions \([0,T]\to\R^m\), and set
\(\mathcal F:=\operatorname{star}(\mathcal X-\mathcal X)\). Suppose
\(Y_i=f^*(t_i)+\sigma\xi_i\) for some \(f^*:[0,T]\to\R^m\) and
independent \(\xi_i\sim N(0,I_m)\), conditionally on the observation
times, and let
\(\widehat f\in\operatorname*{argmin}_{f\in\mathcal X}
\frac1n\sum_{i=1}^n\norm{Y_i-f(t_i)}_2^2\). Suppose \(\delta_{G,n}>0\)
satisfies
\begin{equation}
 \label{eq:gaussian-critical-inequality}
 \frac{\mathfrak{G}_n(\delta_{G,n};\mathcal F)}{\delta_{G,n}}
 \le\frac{\delta_{G,n}}{2\sigma}.
\end{equation}
There is a universal constant \(c_{G,0}>0\) such that, for every
\(r\ge\delta_{G,n}\) and \(\gamma\in(0,1)\),
\begin{align}
 \norm{\widehat f-f^*}_n^2
 &\le\frac{1+\gamma}{1-\gamma}
 \inf_{f\in\mathcal X}
 \norm{f-f^*}_n^2
 +\frac{c_{G,0}}{\gamma(1-\gamma)}r\delta_{G,n}
 \label{eq:vector-oracle-bound}
\end{align}
with conditional probability at least
\(1-\exp(-nr\delta_{G,n}/8\sigma^2)\) given the observation times.
\end{lemma}

\begin{proof}
The optimality of \(\widehat f\) gives, for every \(f\in\mathcal X\),
\begin{align}
 \frac12\norm{\widehat f-f^*}_n^2
 &\le\frac12\norm{f-f^*}_n^2 \notag\\
 &\quad+\left|\frac{\sigma}{n}\sum_{i=1}^n
 \left\langle\xi_i,
 \widehat f(t_i)-f(t_i)\right\rangle\right|.
 \label{eq:basic-oracle-inequality}
\end{align}
Since \(\widehat f,f\in\mathcal X\), the increment \(\widehat f-f\) lies in
\(\mathcal X-\mathcal X\subset\mathcal F\). Fix \(a\ge\delta_{G,n}\) and put
\[
 Z(a):=\sup_{\substack{g\in\mathcal F\\
 \norm g_n\le a}}
 \left|\frac{\sigma}{n}\sum_{i=1}^n\langle\xi_i,g(t_i)\rangle\right|.
\]
Because \(\mathcal F\) is star-shaped, the map
\(\delta\mapsto\mathfrak{G}_n(\delta;\mathcal F)/\delta\) is non-increasing,
so \eqref{eq:gaussian-critical-inequality} gives
\[
 \E Z(a)=\sigma\mathfrak{G}_n(a;\mathcal F)\le\frac{a\delta_{G,n}}{2}\le\frac{a^2}{2}.
\]
To control it's concentration around its mean, we consider it as a function of the \(nm\) Gaussian coordinates
\(\xi=(\xi_1,\ldots,\xi_n)\in\R^{nm}\) via the map
\(\xi\mapsto\frac{\sigma}{n}\sum_{i=1}^n\langle\xi_i,g(t_i)\rangle\). This
is linear, and the Euclidean norm of gradient w.r.t $\xi$ is equal to
\[
 \frac{\sigma}{n}\left(\sum_{i=1}^n\norm{g(t_i)}_2^2\right)^{1/2}
 =\frac{\sigma\norm g_n}{\sqrt n}\le\frac{\sigma a}{\sqrt n},
\]
which implies that the map is \(\sigma a/\sqrt n\)-Lipschitz. The supremum of \(\sigma a/\sqrt n\)-Lipschitz functions is again
\(\sigma a/\sqrt n\)-Lipschitz. This implies gaussian concentration for \(Z(a)\) and hence,
\[
 \Pp\{Z(a)>a^2\}\le\Pp\{Z(a)\ge\E Z(a)+a^2/2\}
 \le\exp\{-na^2/(8\sigma^2)\}.
\]
{ 
Taking $a^2=r\delta_{G,n}$ gives the stated failure probability;
note that $a\ge\delta_{G,n}$ because $r\ge\delta_{G,n}$.
Work on the event $Z(a)\le a^2$. For any $g\in\mathcal F$,
if $\|g\|_n\le a$, the definition of $Z(a)$ gives
\[
 \left|\frac{\sigma}{n}\sum_{i=1}^n
 \langle\xi_i,g(t_i)\rangle\right|\le a^2.
\]
If $\|g\|_n>a$, star-shaped-ness implies
$ag/\|g\|_n\in\mathcal F$, with empirical norm $a$.
Applying the same bound to this rescaled function gives
\[
 \left|\frac{\sigma}{n}\sum_{i=1}^n
 \langle\xi_i,g(t_i)\rangle\right|
 \le a\|g\|_n.
\]
Thus, in both cases, the multiplier is bounded by
$a^2+a\|g\|_n$.

For any $f\in\mathcal X$, put
$d_1=\|\widehat f-f^*\|_n$ and $d_0=\|f-f^*\|_n$.
Using \eqref{eq:basic-oracle-inequality} and
$\|\widehat f-f\|_n\le d_1+d_0$, we obtain
\[
 d_1^2\le d_0^2+2a^2+2a(d_1+d_0).
\]
For $\gamma\in(0,1)$, Young's inequality gives
$2ad_j\le\gamma d_j^2+a^2/\gamma$, so
\[
 (1-\gamma)d_1^2
 \le (1+\gamma)d_0^2+2a^2+\frac{2a^2}{\gamma}
 \le (1+\gamma)d_0^2+\frac{4a^2}{\gamma}.
\]
Dividing by $1-\gamma$, taking the infimum over
$f\in\mathcal X$, and substituting $a^2=r\delta_{G,n}$
proves \eqref{eq:vector-oracle-bound} with $c_{G,0}=4$.
}
\end{proof}

\begin{lemma}[Vector-valued uniform law]
\label{lem:vector-uniform-law}
Let \(\mathcal H\) be a star-shaped class of functions
\([0,T]\to\R^m\) with envelope \(b\), meaning
\(\sup_{h\in\mathcal H}\sup_{t\in[0,T]}\norm{h(t)}_2\le b\). Suppose
\(\delta_{R,n}>0\) satisfies
\begin{equation}
 \label{eq:rademacher-critical-inequality}
 \mathfrak{R}_n^{\mathrm{vec}}(\delta_{R,n};\mathcal H)
 \le\frac{\delta_{R,n}^2}{b}.
\end{equation}
There are universal constants \(c_3,c_4>0\) such that, for every
\(s\ge\delta_{R,n}\), with probability at least
\[
 1-c_3\exp\left(-\frac{c_4ns^2}{b^2}\right),
\]
every \(h\in\mathcal H\) satisfies
\begin{equation}
 \label{eq:vector-uniform-law-bound}
 \bigl|\norm h_n^2-\norm h_{L^2}^2\bigr|
 \le\frac12\norm h_{L^2}^2+\frac12s^2.
\end{equation}
\end{lemma}

\begin{proof}
Apply Wainwright's Theorem~14.1~\cite{wainwright2019} to the scalar class
\(\mathcal K:=\{t\mapsto\norm{h(t)}_2:h\in\mathcal H\}\). Since
\(\mathcal H\) is star-shaped, so is \(\mathcal K\): if
\(h\in\mathcal H\) and \(\beta\in[0,1]\), then \(\beta h\in\mathcal H\)
and \(\norm{(\beta h)(t)}_2=\beta\norm{h(t)}_2\), so \(\beta k\in\mathcal K\)
for \(k(t)=\norm{h(t)}_2\). The envelope of \(\mathcal K\) is bounded by
\(b\), and its localized Rademacher complexity is exactly
\(\mathfrak{R}_n^{\mathrm{vec}}(\delta;\mathcal H)\). Moreover,
\[
 \norm{t\mapsto\norm{h(t)}_2}_n=\norm h_n,
 \qquad
 \norm{t\mapsto\norm{h(t)}_2}_{L^2}=\norm h_{L^2}.
\]
Thus the scalar result gives \eqref{eq:vector-uniform-law-bound}.
\end{proof}

\begin{lemma}[Selective mean attention module]
    \label{lem:selective-mean-attention}
    Let
\[
    X\in\mathbb R^{(d+k)\times n},
    \qquad
     {P_{\mathrm{sel}}}:=
    \begin{pmatrix}
        I_d&0\\
        0&0_{k\times k}
    \end{pmatrix}
    \in\mathbb R^{(d+k)\times(d+k)},
\]
where the columns of $X$ are the tokens. If $T_n:=\frac1n\mathbf 1_n\mathbf 1_n^\top$, then a module consisting of a token-wise affine input map, a residual attention layer with one global head, and a token-wise affine output map realizes exactly the operation
\begin{equation}
    X\longmapsto  {P_{\mathrm{sel}}}XT_n+(I_{d+k}- {P_{\mathrm{sel}}})X.
    \label{eq:general-selective-mean}
\end{equation}
Thus the coordinates selected by $ {P_{\mathrm{sel}}}$ are replaced, in every token, by
their mean over all $n$ tokens, while the complementary coordinates are
left unchanged. The module has
width \(O(2d+k)\), constant depth, and \(O(d+k)\) nonzero parameters in total.
\end{lemma}

\begin{proof}
Write \(X=(A^\top,B^\top)^\top\), where
\(A\in\mathbb R^{d\times n}\) and \(B\in\mathbb R^{k\times n}\).
An affine map appends \(d\) zero coordinates to each token to be intermediate representation $Z\in\mathbb R^{(2d+k)\times n}$, and is followed by a residual attention layer with one head with weight matrices
\[Q=K=0\quad\text{and}\quad V=\begin{pmatrix}
 0_{d\times d}&0_{d\times k}&0_{d\times d}\\
 0_{k\times d}&0_{k\times k}&0_{k\times d}\\
 I_d&0_{d\times k}&0_{d\times d}
 \end{pmatrix}.\]
The attention matrix equals \(T_n\), so the residual attention layer gives
\[
 Z+VZT_n
 =\begin{pmatrix}A\\B\\AT_n\end{pmatrix}.
\]
The token-wise affine map
then returns \(( (AT_n)^\top,B^\top)^\top\), which is
\eqref{eq:general-selective-mean}, and the stated size bounds are clear.
\end{proof}
\begin{remark}
This can be implemented with an arbitrary number of heads $H$ by replacing $V$ with $V/H$.
\end{remark}
\begin{lemma}[Restriction to fixed coordinates]
    \label{lem:restricted-network}
    Let
    \[
        \Phi:\mathbb R^{d+1}\longrightarrow\mathbb R^s
    \]
    be a feedforward network with squared-ReLU activation, and fix
    $\vartheta\in\mathbb R^d$. There is a network with the same activation
    $\Phi_\vartheta:\mathbb R\to\mathbb R^s$ satisfying
    \[
        \Phi_\vartheta(t)=\Phi( {\vartheta,t})
        \qquad\text{for every }t\in\mathbb R.
    \]
    Its depth and width have the same order as those of $\Phi$, and its
    number of unique nonzero parameters is
    \[
        O\!\left(\#\operatorname{nzparams}(\Phi)+d\right).
    \]
\end{lemma}

\begin{proof}
Introduce the affine input layer
\[
    E_\vartheta(t)
    :=
     {\begin{pmatrix}0_{d\times1}\\1\end{pmatrix}}t
    + {\begin{pmatrix}\vartheta\\0\end{pmatrix}}
    = {\begin{pmatrix}\vartheta\\t\end{pmatrix}}.
\]
Then $\Phi_\vartheta=\Phi\circ E_\vartheta$, proving the first statement
and its size bounds.
\end{proof}

\paragraph{Parallel stacking.}
For fixed $\vartheta_1,\ldots,\vartheta_{ N}\in\mathbb R^d$, the restricted
networks can be stacked vertically using block weight matrices.  Use
\[
    E_{\vartheta_1,\ldots,\vartheta_{ N}}(t)
    :=
    \begin{pmatrix}
        E_{\vartheta_1}(t)\\
        \vdots\\
        E_{\vartheta_{ N}}(t)
    \end{pmatrix}
\]
as the first affine layer.  At every subsequent layer, place ${ N}$ copies
of the corresponding weight matrix of $\Phi$ on the diagonal and stack
the biases.  The copies are tied, so only the ${ N}d$ fixed-coordinate
entries are new unique nonzero parameters.  The resulting network realizes
\[
    t\longmapsto
    \begin{pmatrix}
        \Phi( {\vartheta_1,t})\\
        \vdots\\
        \Phi( {\vartheta_{ N},t})
    \end{pmatrix},
\]
has depth of the same order as $\Phi$, width at most ${ N}$ times the
width of $\Phi$, and
\[
    O\!\left(\#\operatorname{nzparams}(\Phi)+{ N}d\right)
\]
unique nonzero parameters.

\begin{lemma}[Exact identity channels]
\label{lem:network-with-identity-channels}
Let \(\Phi:\mathbb R^k\to\mathbb R^s\) be a squared-ReLU network
with an affine output. For every \(r\ge1\), there is a squared-ReLU
network satisfying
\[
 \widetilde\Phi(z,u)=(\Phi(z),u),
 \qquad (z,u)\in\mathbb R^k\times\mathbb R^r.
\]
It has the same depth, adds at most \(4r\) hidden coordinates per
layer, and has \(O(r{ L_\Phi})\) additional nonzero parameters, where
\({ L_\Phi}=\operatorname{depth}(\Phi)\).
If corresponding identity blocks are tied across hidden layers,
their unique-parameter overhead is \(O(r)\).
\end{lemma}
\begin{proof}
Let \(\sigma_2(z)=(z_+)^2\). For a scalar \(u\), define
\[
 a=(1,-1,1,-1)^\top,\qquad b=(1,-1,-1,1)^\top,\qquad
 c=\tfrac14(1,1,-1,-1)^\top,\qquad E(u)=\sigma_2(au+b).
\]
The affine decoder recovers the scalar exactly:
\[
 c^\top E(u)=\frac{(u+1)^2-(u-1)^2}{4}=u.
\]
Thus each subsequent hidden layer preserves the encoded scalar by
\[
 E(u)\longmapsto \sigma_2(ac^\top E(u)+b)=E(u).
\]
Apply this construction independently to the \(r\) coordinates of
\(u\). Append the encoders to the first affine map of \(\Phi\),
append the blocks \(ac^\top\) and biases \(b\) to each intermediate
map, and decode using \(c^\top\) at the final affine output.
These blocks operate separately from the original computation of
\(\Phi\), giving exactly \((\Phi(z),u)\) on the whole input space.
There are four additional hidden coordinates per scalar and
\(O(r)\) nonzero entries per layer, with no added hidden layers.
Tying the repeated blocks gives the asserted unique-parameter bound.
For an affine-only network, append the identity directly.
\end{proof}

\begin{lemma}[Exact squaring and multiplication]
\label{lem:squared-relu-arithmetic}
Squared-ReLU networks with one hidden layer and affine outputs
represent \(z^2\) and \(ab\) exactly on \(\mathbb R\) and
\(\mathbb R^2\), with hidden widths two and four, respectively.
Computing \(r\) such scalar operations in parallel has constant
depth, \(O(r)\) width, and \(O(r)\) nonzero parameters.
\end{lemma}
\begin{proof}
With \(\sigma_2(z)=(z_+)^2\), set
\[
 Q(z)=\sigma_2(z)+\sigma_2(-z)=z^2.
\]
Polarization gives
\[
 M(a,b)=\frac{Q(a+b)-Q(a-b)}4
 =\frac{\sigma_2(a+b)+\sigma_2(-a-b)
       -\sigma_2(a-b)-\sigma_2(b-a)}4=ab.
\]
All activation inputs and output combinations are affine.
Parallel copies give the stated bounds. Affine summation additionally
gives exact squared Euclidean norms and matrix--vector products.
\end{proof}

\begin{lemma}[Parameter bounds and Lipschitz-ness for the clipped transformer class]
\label{lem:transformer-parameter-bounds}
Let \(\mathcal A\) denote the bounded transformer class in
Section~\ref{subsec:pretraining-comparator} with
a full set of trainable parameters without parameter tying (so { $P=O(HW^2L)$}). Its parameter
domain \(\Phi_{M_A,M_b}\subset\R^P\) is compact and is contained in the Euclidean
ball of radius
\begin{equation}
R_\Phi:=\sqrt{L\bigl[(3H+2)WM_A^2+2M_b^2\bigr]
                   +(d_{\mathrm{in}}+d_{\mathrm{out}})M_A^2}.
 \label{eq:pretraining-parameter-radius}
\end{equation}
Here \( {L}\) denotes transformer depth. Furthermore, define
\begin{align*}
 S&:=\max\{M_A\sqrt{d_{\mathrm{in}}}R_1,R_2\},\quad
 Z_*:=M_AS+M_b,\\
 a_*&:=\max\!\left\{1,\,2M_A^2Z_*
 \bigl[1+H(M_A+4M_A^3R_2^2)\bigr]\right\},\\
 c_*^2&:=\bigl[1+H(M_A+4M_A^3R_2^2)\bigr]^2
 \bigl[4M_A^2Z_*^2(S^2+1)+Z_*^4+1\bigr]\\[-1pt]
&\quad +H\bigl(R_2^2+8M_A^4R_2^6\bigr).
\end{align*}
Then, with
\begin{equation}
 \mathfrak{L}_\Phi:=\left[
 M_A^2a_*^{2L}d_{\mathrm{in}}R_1^2
 +M_A^2c_*^2\sum_{j=0}^{L-1}a_*^{2j}+R_2^2
 \right]^{1/2},
 \label{eq:pretraining-lipschitz}
\end{equation}
every \(\phi,\psi\in\Phi\) satisfy
\[
 \sup_{\mathcal D}\|a_\phi(\mathcal D)-a_\psi(\mathcal D)\|_2
 \le \mathfrak{L}_\Phi\|\phi-\psi\|_2,
\]
where $a_\phi$ and $a_\psi$ are elements of $\mathcal{A}$ with parameters $\phi$ and $\psi$, respectively.
In particular, the $R_\Phi$ and $\mathfrak{L}_\Phi$ bounds do not depend on the context length $n$.
\end{lemma}
\begin{proof}
\emph{Parameter radius.}
The Euclidean norm of the parameter vector is the square root of the
sum of the squared Frobenius norms of its weight matrices and the
squared Euclidean norms of its biases. Fixed zero entries do not
contribute. For a matrix \(A\in\R^{r\times s}\),
\(\|A\|_F^2\le\min(r,s)\|A\|_{\mathrm{op}}^2\).
Each head has three square \(W\times W\) matrices, contributing at most
\(3HWM_A^2\)
per layer. The two feedforward matrices contribute at most
\(2WM_A^2\), and the two biases at most \(2M_b^2\). Summing over
the \(L\) layers and adding the input and output projection contributions,
at most \((d_{\mathrm{in}}+d_{\mathrm{out}})M_A^2\), proves
\eqref{eq:pretraining-parameter-radius}.
The fixed zero constraints and norm constraints are closed; thus the
bounded set \(\Phi\) is compact.

Use the maximum token norm \(\|X\|_{\max}:=\max_i\|X_i\|_2\).
For each component \(F_\eta\) of the transformer with parameters $\eta$, we establish a joint estimate
\begin{align*}
 \|F_\eta(X)-F_{\widetilde\eta}(Y)\|_{\max}
 &\le \|F_\eta(X)-F_{\eta}(Y)\|_{\max}+\|F_\eta(Y)-F_{\widetilde\eta}(Y)\|_{\max}\\
 &\le a_F\|X-Y\|_{\max}+b_F\|\eta-\widetilde\eta\|_2.
\end{align*}
In our case, we deal with a two-layer FFN and an attention layer as our two types of components. Coordinate-wise input clipping followed by \(P_{\mathrm{in}}\) gives a token norm at most
\(M_A\sqrt{d_{\mathrm{in}}}R_1\), and subsequent component inputs have norm at
most \(R_2\). Hence every component input has norm at most \(S\).
\\\emph{Feedforward map.} The feedforward pre-activation has norm at most \(Z_*\). Now consider the activation bounds
\[
 \|\sigma_2(v)\|_2\le\|v\|_2^2,\qquad
 \|\sigma_2(v)-\sigma_2(w)\|_2\le2Z_*\|v-w\|_2
 \quad(\|v\|_2,\|w\|_2\le Z_*),
\]
and write \(F_\eta(x)=W_2\sigma_2(W_1x+b_1)+b_2\). The preceding bounds give the
input Lipschitz factor
\(a_{\mathrm{ffn}}:=2M_A^2Z_*\).
For the parameter dependence, fix \(\|x\|_2\le S\) and compare
\(\eta=(W_1,b_1,W_2,b_2)\) with
\(\widetilde\eta=(\widetilde W_1,\widetilde b_1,\widetilde W_2,\widetilde b_2)\).
Set \(z=W_1x+b_1\) and \(\widetilde z=\widetilde W_1x+\widetilde b_1\) where
both have norm at most \(Z_*\). Adding and subtracting
\(W_2\sigma_2(\widetilde z)\) gives the exact decomposition
\[
 F_\eta(x)-F_{\widetilde\eta}(x)
 =W_2\bigl[\sigma_2(z)-\sigma_2(\widetilde z)\bigr]
 +(W_2-\widetilde W_2)\sigma_2(\widetilde z)+(b_2-\widetilde b_2).
\]
Since operator norms are bounded by Frobenius norms,
\[
 \|z-\widetilde z\|_2
 \le S\|W_1-\widetilde W_1\|_F+\|b_1-\widetilde b_1\|_2.
\]
Using the activation bounds above and \(\|W_2\|_{\mathrm{op}}\le M_A\),
we therefore obtain
\begin{align*}
 \|F_\eta(x)-F_{\widetilde\eta}(x)\|_2
 &\le 2M_AZ_*\|z-\widetilde z\|_2
       +Z_*^2\|W_2-\widetilde W_2\|_F+\|b_2-\widetilde b_2\|_2\\
 &\le 2M_AZ_*S\|W_1-\widetilde W_1\|_F
       +2M_AZ_*\|b_1-\widetilde b_1\|_2\\
 &\quad+Z_*^2\|W_2-\widetilde W_2\|_F+\|b_2-\widetilde b_2\|_2.
\end{align*}
The squared norm of the full parameter difference is
\[
 \|\eta-\widetilde\eta\|_2^2
 =\|W_1-\widetilde W_1\|_F^2+\|b_1-\widetilde b_1\|_2^2
  +\|W_2-\widetilde W_2\|_F^2+\|b_2-\widetilde b_2\|_2^2.
\]
Cauchy--Schwarz applied to the four terms thus yields
\(\|F_\eta(x)-F_{\widetilde\eta}(x)\|_2
\le b_{\mathrm{ffn}}\|\eta-\widetilde\eta\|_2\), where
\[
 b_{\mathrm{ffn}}^2:=4M_A^2Z_*^2(S^2+1)+Z_*^4+1.
\]
Thus we obtain for the feedforward map
\[
 \|F_\eta(X)-F_{\widetilde\eta}(Y)\|_{\max}
 \le a_{\mathrm{ffn}}\|X-Y\|_{\max}
       +b_{\mathrm{ffn}}\|\eta-\widetilde\eta\|_2.
\]
\emph{One attention residual.}
Recall for softmax on \(\mathbb R^n\),
\begin{equation}
 \|\operatorname{softmax}(u)-\operatorname{softmax}(v)\|_1
 \le2\|u-v\|_\infty.
 \label{eq:pretraining-softmax-lipschitz}
\end{equation}
For one global head and input \(X=[x_1\ \cdots\ x_n]\), write
\[
 s_{ji}=(Kx_j)^\top Qx_i,\qquad
 \alpha_{\cdot i}=\operatorname{softmax}(s_{\cdot i}),\qquad
 o_i=\sum_j\alpha_{ji}Vx_j.
\]
For a second input \(Y=[y_1\ \cdots\ y_n]\) and parameter matrices
\((\widetilde Q,\widetilde K,\widetilde V)\) for the same head, define
\[
 \widetilde s_{ji}=(\widetilde K y_j)^\top\widetilde Q y_i,\qquad
 \widetilde\alpha_{\cdot i}
   =\operatorname{softmax}(\widetilde s_{\cdot i}),\qquad
 \widetilde o_i=\sum_{j=1}^n\widetilde\alpha_{ji}\widetilde V y_j.
\]
Adding and subtracting
\(\sum_j\alpha_{ji}\widetilde V y_j\)
gives
\[
 \|o_i-\widetilde o_i\|_2
 \le \max_j\|Vx_j-\widetilde V y_j\|_2
      +M_AR_2\|\alpha_{\cdot i}-\widetilde\alpha_{\cdot i}\|_1.
\]
This uses only that the attention weights are nonnegative and sum to one.

Attention inputs have token norm at most \(R_2\). For equal weights, if the input changes
by \(\Delta\) in maximum token norm, each logit changes by at most
\(2M_A^2R_2\Delta\), which propagates to at most \(4M_A^3R_2^2\Delta\) in the second term above. 
Summing over \(H\) heads and including the residual and input perturbation at the value matrix proves the attention
input factor \(a_{\mathrm{att}}:=1+H(M_A+4M_A^3R_2^2)\).
For fixed input \(X=Y\), changing the query and key matrices changes each
score by at most
\[
 |s_{ji}-\widetilde s_{ji}|
 \le M_AR_2^2\bigl(\|Q-\widetilde Q\|_F+\|K-\widetilde K\|_F\bigr).
\]
Combining this with \eqref{eq:pretraining-softmax-lipschitz} and the
preceding output bound gives
\[
 \|o_i-\widetilde o_i\|_2
 \le R_2\|V-\widetilde V\|_F
 +2M_A^2R_2^3
       \bigl(\|Q-\widetilde Q\|_F+\|K-\widetilde K\|_F\bigr).
\]
Write \(\Delta V_h=V_h-\widetilde V_h\), and similarly for
\(\Delta Q_h,\Delta K_h\). Since the input \(X\) is fixed, the residual
terms cancel when comparing the two attention layers. The triangle
inequality over heads therefore gives
\begin{align*}
 \|\mathcal A_{\eta_{\mathrm{att}}}(X)
       -\mathcal A_{\widetilde\eta_{\mathrm{att}}}(X)\|_{\max}
 &\le \sum_{h=1}^H
 \left[R_2\|\Delta V_h\|_F
 +2M_A^2R_2^3\|\Delta Q_h\|_F
 +2M_A^2R_2^3\|\Delta K_h\|_F\right].
\end{align*}
Apply Cauchy--Schwarz to these \(3H\) terms to obtain
\[
 \|\mathcal A_{\eta_{\mathrm{att}}}(X)
       -\mathcal A_{\widetilde\eta_{\mathrm{att}}}(X)\|_{\max}\le b_{\mathrm{att}}\|\eta_{\mathrm{att}}-\widetilde\eta_{\mathrm{att}}\|_2 
 \] where

\[
 \sum_{h=1}^H
 \left(\|\Delta V_h\|_F^2+\|\Delta Q_h\|_F^2+\|\Delta K_h\|_F^2\right)
 =\|\eta_{\mathrm{att}}-\widetilde\eta_{\mathrm{att}}\|_2^2\quad \text{and}\quad b_{\mathrm{att}}^2:=H(R_2^2+8M_A^4R_2^6)
\]
Thus
\[
 \|\mathcal A_{\eta_{\mathrm{att}}}(X)
       -\mathcal A_{\widetilde\eta_{\mathrm{att}}}(Y)\|_{\max}\le a_{\mathrm{att}}\|X-Y\|_{\mathrm{max}}+b_{\mathrm{att}}\|\eta_{\mathrm{att}}-\widetilde\eta_{\mathrm{att}}\|_2 
 \]

\emph{Composition.}
Set \(\overline F_\eta:=\operatorname{Proj}_{R_2}\circ F_\eta\).
Euclidean projection is non-expansive, so \(\overline F_\eta\) has the same
joint Lipschitz constants \(a_{\mathrm{ffn}},b_{\mathrm{ffn}}\) as \(F_\eta\), and its
output tokens have norm at most \(R_2\). For a block with parameter vector
\(\eta=(\eta_{\mathrm{ffn}},\eta_{\mathrm{att}})\), the two joint estimates give
\begin{align*}
 \|\mathcal A_{\eta_{\mathrm{att}}}(\overline F_{\eta_{\mathrm{ffn}}}(X))
 -\mathcal A_{\widetilde\eta_{\mathrm{att}}}(\overline F_{\widetilde\eta_{\mathrm{ffn}}}(Y))\|_{\max}
 &\le a_{\mathrm{att}}a_{\mathrm{ffn}}\|X-Y\|_{\max}\\
 &\quad+a_{\mathrm{att}}b_{\mathrm{ffn}}
       \|\eta_{\mathrm{ffn}}-\widetilde\eta_{\mathrm{ffn}}\|_2
       +b_{\mathrm{att}}\|\eta_{\mathrm{att}}-\widetilde\eta_{\mathrm{att}}\|_2\\
 &\le a_*\|X-Y\|_{\max}+c_*\|\eta-\widetilde\eta\|_2,
\end{align*}
where \(a_*=\max\{1,a_{\mathrm{att}}a_{\mathrm{ffn}}\}\) and
\(c_*^2=a_{\mathrm{att}}^2b_{\mathrm{ffn}}^2+b_{\mathrm{att}}^2\), exactly as stated
in the lemma. If \(e_\ell\) is the maximum token
difference after projected layer \(\ell\), non-expansiveness of
Euclidean projection gives
\[
 e_0\le\sqrt{d_{\mathrm{in}}}R_1
       \|P_{\mathrm{in}}^\phi-P_{\mathrm{in}}^\psi\|_F,\qquad
 e_\ell\le a_*e_{\ell-1}+c_*\|\phi_\ell-\psi_\ell\|_2.
\]
Iteration gives
\[
 e_L\le a_*^L\sqrt{d_{\mathrm{in}}}R_1
       \|P_{\mathrm{in}}^\phi-P_{\mathrm{in}}^\psi\|_F
       +c_*\sum_{\ell=1}^L a_*^{L-\ell}
                      \|\phi_\ell-\psi_\ell\|_2.
\]
The final hidden tokens have norm at most \(R_2\). The output matrix has
operator norm at most \(M_A\); selecting one token and applying
\(\operatorname{Proj}_{R_3}\) are non-expansive. Hence
\[
 \|a_\phi(\mathcal D)-a_\psi(\mathcal D)\|_2
 \le M_Ae_L+R_2\|P_{\mathrm{out}}^\phi-P_{\mathrm{out}}^\psi\|_F.
\]
Cauchy--Schwarz across parameter blocks proves the claimed
bound with \(\mathfrak{L}_\Phi\) above.
\end{proof}

\begin{lemma}[Two-sided Bernstein inequality]
\label{lem:two-sided-bernstein}
Let \(X_1,\ldots,X_M\) be independent, mean-zero real random variables
such that \(|X_j|\le b\) almost surely and
\(\sum_{j=1}^M\E X_j^2\le V\). Then, for every \(t>0\),
\[
 \Pp\left(\left|\sum_{j=1}^M X_j\right|
       >\sqrt{2Vt}+\frac{2bt}{3}\right)\le2e^{-t}.
\]
In particular, if \(\E X_j^2\le v\) for every \(j\), then with
probability at least \(1-2e^{-t}\),
\[
 \left|\frac1M\sum_{j=1}^M X_j\right|
 \le\sqrt{\frac{2vt}{M}}+\frac{2bt}{3M}.
\]
\end{lemma}
\begin{proof}
The bounded-variable Bernstein inequality gives, for \(x>0\),
\[
 \Pp\left(\sum_{j=1}^M X_j\ge x\right)
 \le\exp\left(-\frac{x^2}{2(V+bx/3)}\right);
\]
see \cite[Theorem~2.10 and Corollary~2.11]{boucheron2013}.
The same bound applies to \(-X_j\). For
\(x=\sqrt{2Vt}+2bt/3\), one has \(x^2\ge2t(V+bx/3)\).
A union bound proves the two-sided statement, and taking \(V=Mv\)
and dividing by \(M\) gives the second assertion. If \(V=0\), all
the centered variables vanish almost surely and the conclusions are immediate.
\end{proof}
\begin{theorem}[Generic pretraining guarantee]
\label{thm:generic-pretraining}
Let \(\mathcal A\) be the class in \eqref{eq:pretraining-class}, with
\(R_\Phi,\mathfrak{L}_\Phi\) as in Lemma~\ref{lem:transformer-parameter-bounds}
in Appendix~\ref{app:background}. For a fixed $n>0$, consider $M$ i.i.d. samples of $(U,Z)\sim Q$, where $U\in \R^{d_{\mathrm{in}}\times n}$ and $Z\in \R^{d_{\mathrm{out}}}$ with $\|Z\|_2\leq B_Z$. For each sample, let $f_0(U)$ be the Bayes predictor for $Z$ given $U$, and define 
\[
 \mathcal{R}_{\mathrm{pre}}(f):=\E\|f(U)-Z\|_2^2 \quad\text{for any }f\in\mathcal A,
\]
along with the empirical risk ${\mathcal{R}}_{\mathrm{pre}}^{(M)}(f):=M^{-1}\sum_{j=1}^M\|f(U_j)-Z_j\|_2^2$.
Let
\(\widehat f_M\in\mathcal A\) be a measurable approximate empirical risk
minimizer satisfying
\[
{\mathcal{R}}_{\mathrm{pre}}^{(M)}(\widehat f_M)
 \le\inf_{f\in\mathcal A}{\mathcal{R}}_{\mathrm{pre}}^{(M)}(f)
       +\varepsilon_{\mathrm{opt}}.
\]
For every \(M\ge1\) and \(\delta\in(0,1)\), with probability at least
\(1-\delta\) over the pretraining tasks,
\begin{align}
 \mathcal{R}_{\mathrm{pre}}(\widehat f_M)-\mathcal{R}_{\mathrm{pre}}(f_0)
 &\lesssim \inf_{f\in\mathcal A}
                      [\mathcal{R}_{\mathrm{pre}}(f)-\mathcal{R}_{\mathrm{pre}}(f_0)]\notag\\
 &\quad+\frac{B_*^2}{M}
 \left[P\log\left(e+\frac{MR_\Phi \mathfrak{L}_\Phi}{B_*}\right)
                    +\log\frac2\delta\right]+\varepsilon_{\mathrm{opt}},
 \label{eq:generic-pretraining-rate}
\end{align}
where the constant is universal, and \(B_*:=\max\{R_3+B_Z,2B_Z\}\).
\end{theorem}
\begin{proof}[Proof of Theorem~\ref{thm:generic-pretraining}]
Lemma~\ref{lem:transformer-parameter-bounds} gives a compact parameter
domain \(\Phi\subseteq\overline B(0,R_\Phi)\) and the uniform
parameter Lipschitz constant \(\mathfrak{L}_\Phi\).
\paragraph{Step 1:  concentration via Bernstein.}

We seek a comparison between empirical loss and population excess risk
that holds for every predictor, including the one selected from the data.
First fix \(f\in\mathcal A\), independently of the pretraining sample, and define
\[
 g_f(Z,U):=\|f(U)-Z\|_2^2
                         -\|f_0(U)-Z\|_2^2,
 \qquad E_f:=\E_{Z,U} g_f=\mathcal{R}_{\mathrm{pre}}(f)-\mathcal{R}_{\mathrm{pre}}(f_0)=\E\|f-f_0\|_2^2,
\]
where $E_f$ is the excess risk of the predictor $f$.
Note that the final projection in the architecture implies \(|g_f|\le B_*^2\). Moreover,
\[
 |g_f|=|\langle f-f_0,f-Z+f_0-Z\rangle|
 \le2B_*\|f-f_0\|_2,
 \qquad
 \operatorname{Var}(g_f)\le\E g_f^2\le4B_*^2E_f.
\]
 Considering now the centered variables
\(X_j=g_f(Z_j,U_j)-E_f\), with
\(|X_j|\le2B_*^2\) and \(\E X_j^2\le4B_*^2E_f\), and defining the empirical excess risk \(\widehat E_f=M^{-1}\sum_j g_f(Z_j,U_j)\), we may apply Lemma~\ref{lem:two-sided-bernstein} with \(b=2B_*^2\) and
\(v=4B_*^2E_f\) to obtain for all \(t>0\) 
\[
 |\widehat E_f-E_f|
 \le\sqrt{\frac{8B_*^2E_f t}{M}}+\frac{4B_*^2t}{3M}
 \le\tfrac12E_f+\frac{16B_*^2t}{M}
\]
except on a set of probability at most \(2e^{-t}\). The last step uses
\(\sqrt{uv}\le(u+v)/2\).\\
To make this statement uniform over all \(f\in\mathcal A\), take a deterministic \(u\)-net of
\(\mathcal A\) with \(J\) representatives. In particular, this can be done by take the preimage of a covering in parameter space $\Phi$ to obtain a covering in $\mathcal A$. This way, for any \(u>0\), the parameter Lipschitz property gives
\begin{equation}
 \mathcal N(u;\mathcal A,\|\cdot\|_\infty)
 \le\left(1+\frac{2R_\Phi \mathfrak{L}_\Phi}u\right)^P.
 \label{eq:pretraining-cover}
\end{equation} 
Now choose \(t=\log(2J/\delta)\).
Each representative fails with probability at most \(\delta/J\), so
a union bound gives the preceding inequality for all representatives
on one event of probability at least \(1-\delta\).
For any \(f\in\mathcal A\), choose a representative \(f'\) satisfying
\(\|f-f'\|_\infty\le u\). This gives
\[
 |g_f-g_{f'}|\le2B_*u\Rightarrow
 |\widehat E_f-\widehat E_{f'}|\le2B_*u\Rightarrow
 |E_f-E_{f'}|\le2B_*u.
\]
Consequently, on the same event,
\[
 |\widehat E_f-E_f|
 \le |\widehat E_{f'}-E_{f'}|+4B_*u
 \le\tfrac12 E_f+\frac{16B_*^2t}{M}+5B_*u.
\]
Setting \(u=B_*/M\) gives, via the covering bound \eqref{eq:pretraining-cover},
\[
 t\le P\log\left(1+\frac{2MR_\Phi \mathfrak{L}_\Phi}{B_*}\right)
       +\log\frac2\delta.
\]
Substituting this bound proves, simultaneously for all
\(f\in\mathcal A\),
\begin{equation}
 \left|\frac1M\sum_{j=1}^M g_f(Z_j,U_j)-E_f\right|
 \le\tfrac12E_f+\frac{B_*^2}{M}
 \left[{ 32P}\log\left(e+\frac{MR_\Phi \mathfrak{L}_\Phi}{B_*}\right)
                  +16\log\frac2\delta+5\right]
 \label{eq:pretraining-relative-deviation}
\end{equation}
with probability at least \(1-\delta\).

\paragraph{Step 2: empirical minimization.}
We now work on this uniform event. Because its bound holds for every predictor in
\(\mathcal A\), it also holds for the data-dependent choice
\(\widehat f_M\). Recall that
\[
 \widehat E_{\widehat f_M}=\widehat{\mathcal{R}}_{\mathrm{pre}}^{(M)}(\widehat f_M)
                    -\widehat{\mathcal{R}}_{\mathrm{pre}}^{(M)}(f_0).
\]
For any comparator \(f\in\mathcal A\), approximate empirical
minimization gives
\(\widehat{\mathcal{R}}_{\mathrm{pre}}^{(M)}(\widehat f_M)
\le\widehat{\mathcal{R}}_{\mathrm{pre}}^{(M)}(f)+\varepsilon_{\mathrm{opt}}\).
Subtracting the same empirical Bayes loss on both sides yields
\(\widehat E_{\widehat f_M}\le\widehat E_f+\varepsilon_{\mathrm{opt}}\).
Using \eqref{eq:pretraining-relative-deviation} we obtain
\[
 \tfrac12 E_{\widehat f_M}-C\operatorname{pen}_M(\delta)
 \le\widehat E_{\widehat f_M}
 \le\widehat E_f+\varepsilon_{\mathrm{opt}}
 \le\tfrac32 E_f+C\operatorname{pen}_M(\delta)
                    +\varepsilon_{\mathrm{opt}},
\]
where $\mathrm{pen}_M(\delta)=\frac{B_*^2}{M}
 \left[{ 32P}\log\left(e+\frac{MR_\Phi \mathfrak{L}_\Phi}{B_*}\right)
                  +16\log\frac2\delta+5\right]$
Rearranging gives
\[
 E_{\widehat f_M}\le3E_f+2\varepsilon_{\mathrm{opt}}
                         +4C\operatorname{pen}_M(\delta).
\]
This holds for every comparator on the same event. Taking the infimum
over \(f\in\mathcal A\), using
\(E_f=\mathcal{R}_{\mathrm{pre}}(f)-\mathcal{R}_{\mathrm{pre}}(f_0)\), and enlarging the universal
constant proves \eqref{eq:generic-pretraining-rate}. 
\end{proof}

\section{Proofs for Statistical Parameter Recovery}
\label{app:statistical}
\label{sec:proof-ideas}
We briefly discuss the main arguments underlying the transformer approximation and pretraining guarantees.

\medskip
\paragraph{Parameter recovery and trajectory prediction}
Under the Gaussian observation model, maximum likelihood estimation is equivalent to minimizing a trajectory-matching least-squares loss. For a given true parameter $\theta^*$, consider the within-task population risk
\begin{equation}
    \mathcal{R}_{\theta^*}(\theta)
    :=\E_{t,\xi}\big[\norm{x(\theta,t)-Y(\theta^*,t)}_2^2\big].
    \label{eq:ideal-risk}
\end{equation}
By global identifiability, this risk is uniquely minimized at $\theta^*$. Given $n$ noisy observations, the corresponding empirical risk is
\begin{equation}
    \mathcal{R}_{\theta^*}^{(n)}(\theta)
    :=\frac1n\sum_{i=1}^n
      \norm{x(\theta,t_i)-Y(\theta^*,t_i)}_2^2,
    \label{eq:ideal-empirical-risk}
\end{equation}
with minimizer $\theta_n\in\argmin_{\theta\in\Theta}\mathcal{R}_{\theta^*}^{(n)}(\theta)$.

However, evaluating this empirical risk requires the ODE solution map, which is generally not available in closed form. We therefore replace $x(\theta,t)$ with its approximation $\mathfrak F(\theta,t)$ and consider the corresponding numerical population and empirical risks
\begin{align}
    \widetilde{\mathcal{R}}_{\theta^*}(\theta)
    &:=\E_{t,\xi}\big[\norm{\mathfrak F(\theta,t)-Y(\theta^*,t)}_2^2\big],
       \label{eq:numerical-population-risk}\\
    \widetilde{\mathcal{R}}_{\theta^*}^{(n)}(\theta)
    &:=\frac1n\sum_{i=1}^n
       \norm{\mathfrak F(\theta,t_i)-Y(\theta^*,t_i)}_2^2,
       \label{eq:numerical-empirical-risk}
\end{align}
where $\tilde\theta_n\in\argmin_{\theta\in\Theta}\widetilde{\mathcal{R}}_{\theta^*}^{(n)}(\theta)$. These objectives allow us to distinguish the error due to finite observations from that introduced by approximating the ODE solution.

The main challenge is that the trajectory-matching loss is generally nonconvex, so a direct application of gradient descent may not recover the true parameter. Our analysis instead relies on the local geometry of the population risk. Under the identifiability and regularity assumptions, the population risk is locally strongly convex around $\theta^*$, while its value is uniformly separated from the minimum outside this neighborhood. The constants $\rho$ and $\gamma_\rho$ quantify these properties. We then establish uniform concentration and approximation bounds to show that the empirical and numerical approximations of the risk retain the required curvature and separation when $n$ is sufficiently large and $\varepsilon_{\mathrm{sol}}^{(\mathrm{obs})},\varepsilon_{\mathrm{grad}}$ are sufficiently small.

This allows us to first locate the strongly convex neighborhood through a coarse search over a finite net of the parameter space, without requiring global convexity. Once the initializer lies in this neighborhood, approximate gradient descent can be used to refine the estimate. The resulting recovery error is controlled by the finite-sample estimation error, the sensitivity approximation error, and the error from finitely many gradient descent iterations.

To implement this procedure by a transformer, the feedforward networks perform the required approximations and arithmetic operations, while attention aggregates the empirical loss and gradient contributions across observations. The trajectory prediction result then follows by evaluating an approximation of the ODE solution map at the recovered parameter and the query time.

\medskip
\paragraph{Pretraining guarantees}
For pretraining, the main step is to translate the accuracy of the constructed transformers into a risk bound for a predictor learned from $M$ pretraining tasks. This requires controlling the complexity of the transformer function class. Since the observations contain unbounded Gaussian noise, we introduce input clipping and intermediate projections, together with bounds on the network parameters and outputs. These constraints provide the uniform boundedness and Lipschitz control needed to estimate the covering numbers of the function class.

We then apply localization and Bernstein-type concentration arguments to bound the pretraining estimation error. Using the explicit recovery and prediction transformers as comparators, their within-task guarantees provide the approximation terms in the oracle inequality \eqref{eq:pretraining-oracle-inequality}. Combining these with the pretraining estimation error and the allowed training error gives the stated pretraining risk bounds.

\subsection{Local strong convexity and separability of \texorpdfstring{$\mathcal{R}_{\theta^*}$}{R}}
\label{app:local-conditions}

\begin{lemma}[Uniform local curvature and separation]
\label{lem:uniform-strong-convexity-radius}
Under Assumption~\ref{ass:ode-regularity} suppose that
\[
 \inf_{\vartheta\in\Theta_0}\lambda_{\min}\mathcal I_\nu(\vartheta)\ge\mu>0,
 \qquad
 \E_{t\sim\nu}\|x(\theta,t)-x(\vartheta,t)\|_2^2>0
 \quad\text{for every }\vartheta\in\Theta_0,\ \theta\in\Theta\setminus\{\vartheta\}.
\]
Then there exists $\rho,\gamma_\rho>0$ such that $(\mu,\rho,\gamma_\rho)$
satisfy Assumption~\ref{ass:quantitative-identifiability}.
\end{lemma}

\begin{proof}[Proof of Lemma~\ref{lem:uniform-strong-convexity-radius}]
Define
\begin{equation}
 \rho_{\mathrm{sc}}
 :=\sup\left\{r\le D:
 \inf_{\substack{\vartheta\in\Theta_0,\ \theta\in\Theta\\
 \norm{\theta-\vartheta}_2\le r}}
 \lambda_{\min}(\nabla_\theta^2\mathcal{R}_\vartheta(\theta))\ge\mu\right\},
 \label{eq:strong-convexity-radii}
\end{equation}
We first show $\rho_{\mathrm{sc}}>0$. Differentiating \(\mathcal{R}_\vartheta\) twice with respect to its argument gives
\begin{align}
 \nabla_\theta^2\mathcal{R}_\vartheta(\theta)
 =2\E_t\Bigg[
 \nabla_\theta x(\theta,t) \nabla_\theta x(\theta,t)^\top
 +\sum_{j=1}^m
 [x_j( {\theta,t})-x_j( {\vartheta,t})]
 \nabla_\theta^2x_j( {\theta,t})\Bigg].
 \label{eq:appendix-reference-risk-hessian}
\end{align}
At \(\theta=\vartheta\), the residual term vanishes, so
\[
 \nabla_\theta^2\mathcal{R}_\vartheta(\vartheta)=2\mathcal I_\nu(\vartheta).
\]
The uniform local-identifiability assumption therefore gives
\begin{equation}
 \inf_{\vartheta\in\Theta_0}
 \lambda_{\min}(\nabla_\theta^2\mathcal{R}_\vartheta(\vartheta))
 \ge2\mu.
 \label{eq:appendix-diagonal-curvature}
\end{equation}

By the \(C^{2,1}\) regularity of the trajectory map,
\((\vartheta,\theta)\mapsto\nabla_\theta^2\mathcal{R}_\vartheta(\theta)\) is
continuous on the compact set \(\Theta_0\times\Theta\), hence uniformly
continuous. Thus there is \(r_0>0\) such that
\[
 \norm{\nabla_\theta^2\mathcal{R}_\vartheta(\theta)
 -\nabla_\theta^2\mathcal{R}_\vartheta(\vartheta)}_{\mathrm{op}}\le\mu
 \quad\text{whenever }\vartheta\in\Theta_0,\ 
 \norm{\theta-\vartheta}_2\le r_0.
\]
Weyl's inequality and \eqref{eq:appendix-diagonal-curvature} then imply
\[
 \lambda_{\min}(\nabla_\theta^2\mathcal{R}_\vartheta(\theta))
 \ge2\mu-\mu=\mu
\]
uniformly on these neighborhoods. Hence the set in the definition of
\(\rho_{\mathrm{sc}}\) is nonempty and
\(\rho_{\mathrm{sc}}\ge\min\{r_0,D\}>0\). Continuity of the minimum
eigenvalue also preserves the inequality at the supremal radius. Clearly, setting \(\rho:=\min\{\rho_{\mathrm{sc}},\rho_\partial\}\) satisfies the first condition in Assumption~\ref{ass:quantitative-identifiability}.

For $\gamma_\rho$, set $K_\rho:=\{(\vartheta,\theta)\in\Theta_0\times\Theta: \|\theta-\vartheta\|_2\ge\rho/4\}$. It is compact
and nonempty because $\rho\le\rho_\partial$ and the
ball around each $\vartheta\in\Theta_0$ lies in $\Theta$.
By continuity, and the global identifiability condition, $\E_{t\sim\nu}\|x(\theta,t)-x(\vartheta,t)\|_2^2$ attains a positive minimum on $K_\rho$. 
Setting $\gamma_\rho=\min_{(\vartheta,\theta)\in K_\rho} \E_{t\sim\nu}\|x(\theta,t)-x(\vartheta,t)\|_2^2$ completes the proof.
\end{proof}

\begin{lemma}[Drift-level condition for uniform local identifiability]
\label{lem:drift-sensitivity-nondegeneracy}
Suppose Assumption~\ref{ass:ode-regularity} holds, $\operatorname{supp}(\nu)=[0,T_{\mathrm{obs}}]$,
and $f_\omega(x,t)$ is continuously differentiable in $(x,\omega)$ on a
neighborhood of the trajectories, with these derivatives jointly continuous
in $(x,\omega,t)$. If, for every $\theta=(x_0,\omega)\in\Theta_0$,
\begin{equation}
 \E_{t\sim\nu}\|\nabla_\omega f_\omega(x(\theta,t),t)^\top v\|_2^2>0
 \quad\text{for every }v\in\mathbb R^d\setminus\{0\},
 \label{eq:drift-nondegeneracy}
\end{equation}
then there exists $\mu>0$ such that
\[
 \inf_{\theta\in\Theta_0}\lambda_{\min}\mathcal I_\nu(\theta)\ge\mu.
\]
\end{lemma}

\begin{proof}
Fix $\theta=(x_0,\omega)\in\Theta_0$ and a direction
$w=(u,v)\in\mathbb R^m\times\mathbb R^d$. Define the directional sensitivity
\[
 z_w(t):=\nabla_\theta x(\theta,t)^\top w
 =\nabla_{x_0}x(\theta,t)^\top u+\nabla_\omega x(\theta,t)^\top v.
\]
The transposes follow our convention that
$\nabla_\theta x(\theta,t)\in\mathbb R^{(m+d)\times m}$.
Differentiating the state equation in the direction $w$ gives
\[
 \dot z_w(t)
 =\nabla_x f_\omega(x(\theta,t),t)^\top z_w(t)
 +\nabla_\omega f_\omega(x(\theta,t),t)^\top v,
 \qquad z_w(0)=u,
\]
where the initial condition follows from
$\nabla_{x_0}x(\theta,0)=I_m$ and $\nabla_\omega x(\theta,0)=0$.
Moreover,
\[
 w^\top\mathcal I_\nu(\theta)w
 =\E_{t\sim\nu}\|z_w(t)\|_2^2.
\]
If this quadratic form is zero, then continuity and full support of $\nu$ imply $z_w(t)=0$ for every
$t\in[0,T_{\mathrm{obs}}]$. Evaluating at $t=0$ gives $u=0$, and the sensitivity
equation gives
\[
 \nabla_\omega f_\omega(x(\theta,t),t)^\top v=0
 \quad\text{for every }t\in[0,T_{\mathrm{obs}}].
\]
Condition~\eqref{eq:drift-nondegeneracy} forces $v=0$, so $w=0$.
Thus $\mathcal I_\nu(\theta)$ is positive definite.

Finally, $\theta\mapsto\mathcal I_\nu(\theta)$ is continuous by the joint
continuity and boundedness of the sensitivity on the compact parameter--time
domain. Hence $\theta\mapsto\lambda_{\min}\mathcal I_\nu(\theta)$ is continuous
and strictly positive on the compact set $\Theta_0$. Its attained minimum
is therefore some $\mu>0$, uniform over $\Theta_0$.
\end{proof}

\subsection{Concentration and approximation results}
\begin{lemma}[Uniform numerical-risk concentration]
\label{lem:uniform-risk-concentration}
Assume \(\norm{\mathfrak F}_{L^\infty(\Theta\times [0,T_{\mathrm{obs}}])}<\infty\) and that
\begin{equation}
 \norm{\mathfrak F( {\theta,t})-\mathfrak F( {\theta',t})}_2
 \le L_{\mathfrak F}\norm{\theta-\theta'}_2\quad\text{for all }t\in[0,T_{\mathrm{obs}}].
 \label{eq:F-uniform-lipschitz}
\end{equation}
for some \(L_{\mathfrak F}>0\). Define
\begin{align}
 B_0^2&:=8\sigma^2(\norm{\mathfrak F}_{L^\infty(\Theta\times [0,T_{\mathrm{obs}}])}^2+\norm{x}_{L^\infty(\Theta\times [0,T_{\mathrm{obs}}])}^2)
 +(\norm{\mathfrak F}_{L^\infty(\Theta\times [0,T_{\mathrm{obs}}])}^2+\norm{x}_{L^\infty(\Theta\times [0,T_{\mathrm{obs}}])}^2)^2,
 \label{eq:B0}\\
 B_1&:=2\sqrt{\sigma^2+(\norm{\mathfrak F}_{L^\infty(\Theta\times [0,T_{\mathrm{obs}}])}
 +\norm{x}_{L^\infty(\Theta\times [0,T_{\mathrm{obs}}])})^2}\,L_{\mathfrak F},
 \qquad \kappa:=1+\frac{B_1D}{B_0},
 \label{eq:B1}
\end{align}
Fix $\vartheta\in\Theta_0$ and let $\delta\in (0,1)$. If \(n\ge4\log(2/\delta)/m\), then, for a universal \(C_R>0\), with
probability at least \(1-2\delta\),
\begin{equation}
\begin{aligned}
  \sup_{\theta\in\Theta}|\widetilde{\mathcal{R}}_\vartheta^{(n)}(\theta)-{\mathcal{R}}_\vartheta(\theta)|
  &\le C_RB_0\sqrt{\frac{p\log(2\kappa)+\log(1/\delta)}{n}}
  +3\sigma^2\sqrt{\frac{m\log(2/\delta)}{n}} \\
  &\quad +\norm{\mathfrak F-x}_{L^\infty(\Theta\times [0,T_{\mathrm{obs}}])}^2
  +4\norm{x}_{L^\infty(\Theta\times [0,T_{\mathrm{obs}}])}\norm{\mathfrak F-x}_{L^\infty(\Theta\times [0,T_{\mathrm{obs}}])}.
\end{aligned}
\label{eq:uniform-risk-concentration}
\end{equation}
All constants are independent of \(\vartheta\).
\end{lemma}

\begin{proof}[Proof of Lemma~\ref{lem:uniform-risk-concentration}]
We consider the following decomposition:
\[
 |\widetilde{\mathcal{R}}_\vartheta^{(n)}(\theta)-{\mathcal{R}}_\vartheta(\theta)|
 \le |\widetilde{\mathcal{R}}_\vartheta^{(n)}(\theta)-\widetilde{\mathcal{R}}_\vartheta(\theta)|
 +|\widetilde{\mathcal{R}}_\vartheta(\theta)-{\mathcal{R}}_\vartheta(\theta)|,
\]
and control each term separately. For the second term,
apply \(\norm a_2^2-\norm b_2^2=\norm{a-b}_2^2
+2\langle b,a-b\rangle\) with
\(a=\mathfrak F( {\theta,t})-Y\), \(b=x( {\theta,t})-Y\).
Because \(x( {\theta,t})-Y=x( {\theta,t})-x( {\theta^*,t})-\sigma\xi\), independence
and \(\E\xi=0\) remove the noise cross-term, and this using
\(\norm{x( {\theta,t})-x( {\theta^*,t})}_2\le2\norm{x}_{L^\infty(\Theta\times [0,T_{\mathrm{obs}}])}\) followed by the
Cauchy-Schwarz inequality to get
\[
|\widetilde{\mathcal{R}}_\vartheta(\theta)-{\mathcal{R}}_\vartheta(\theta)|
 \le\norm{\mathfrak F-x}_{L^\infty(\Theta\times [0,T_{\mathrm{obs}}])}^2
 +4\norm{x}_{L^\infty(\Theta\times [0,T_{\mathrm{obs}}])}\norm{\mathfrak F-x}_{L^\infty(\Theta\times [0,T_{\mathrm{obs}}])}.
\]
Now for the second term, we aim to apply Lemma~\ref{lem:bg-subgaussian-process}. Write \(b_\theta(t)=x( {\theta^*,t})-\mathfrak F( {\theta,t})\). The one-observation loss splits as
\[
 \norm{\mathfrak F( {\theta,t})-Y}_2^2
 =\norm{b_\theta(t)}_2^2+2\sigma\langle b_\theta(t),\xi\rangle
 +\sigma^2\norm{\xi}_2^2.
\]
Let \(\ell_1( {\theta,t},\xi)=\norm{b_\theta(t)}_2^2+
2\sigma\langle b_\theta(t),\xi\rangle\). Conditioning first on \(t\)
gives, for every \(\lambda\in\R\),
\begin{align}
 &\E_{t,\xi}\exp\{\lambda(\ell_1-\E\ell_1)\}\notag\\
 &=\E_t\!\left[
 e^{\lambda(\norm{b_\theta(t)}_2^2-\E_t\norm{b_\theta(t)}_2^2)}
 \E_\xi e^{2\sigma\lambda\langle b_\theta,\xi\rangle}\right]\notag\\
 &=\E_t\!\left[
 e^{\lambda(\norm{b_\theta(t)}_2^2-\E_t\norm{b_\theta(t)}_2^2)}
 e^{2\sigma^2\lambda^2\norm{b_\theta}_2^2}\right].
 \label{eq:appendix-risk-mgf}
\end{align}
Uniform boundedness yields
\[
 0\le \norm{b_\theta(t)}_2^2\le
 2(\norm{\mathfrak F}_{L^\infty(\Theta\times [0,T_{\mathrm{obs}}])}^2+\norm{x}_{L^\infty(\Theta\times [0,T_{\mathrm{obs}}])}^2).
\]
Hoeffding's lemma applied to $\norm{b_\theta(t)}_2^2$ therefore implies
\[
 \E_t e^{\lambda(\norm{b_\theta(t)}_2^2-\E_t\norm{b_\theta(t)}_2^2)}
 \le\exp\!\left\{\frac{\lambda^2}{2}
 (\norm{\mathfrak F}_{L^\infty(\Theta\times [0,T_{\mathrm{obs}}])}^2+\norm{x}_{L^\infty(\Theta\times [0,T_{\mathrm{obs}}])}^2)^2\right\}.
\]
Substitution in \eqref{eq:appendix-risk-mgf} proves the point-wise
sub-Gaussian bound with variance proxy \(B_0^2\).

For \(\theta,\theta'\in\Theta\), the difference of their deterministic
squared-error terms is bounded by
\[
 2(\norm{\mathfrak F}_{L^\infty(\Theta\times [0,T_{\mathrm{obs}}])}+\norm{x}_{L^\infty(\Theta\times [0,T_{\mathrm{obs}}])})
 L_{\mathfrak F}\norm{\theta-\theta'}_2,
\]
To verify the increment condition explicitly, put
\(\Delta\mathfrak F=\mathfrak F( {\theta,t})-\mathfrak F( {\theta',t})\) and
\(q_{\theta,\theta'}(t)=\norm{b_\theta(t)}_2^2-\norm{b_{\theta'}(t)}_2^2\). Then
\[
 \ell_1(\theta,t,\xi)-\ell_1(\theta',t,\xi)
 =q_{\theta,\theta'}(t)-2\sigma\langle\Delta\mathfrak F,\xi\rangle.
\]
The displayed deterministic bound means that \(q_{\theta,\theta'}\)
takes values in an interval of length at most
\[
 4(\norm{\mathfrak F}_{L^\infty(\Theta\times [0,T_{\mathrm{obs}}])}+\norm{x}_{L^\infty(\Theta\times [0,T_{\mathrm{obs}}])})
 L_{\mathfrak F}\norm{\theta-\theta'}_2.
\]
Consequently,
\[
 \E_t e^{\lambda(q_{\theta,\theta'}-\E_tq_{\theta,\theta'})}
 \le\exp\!\left\{2\lambda^2
 (\norm{\mathfrak F}_{L^\infty(\Theta\times [0,T_{\mathrm{obs}}])}+\norm{x}_{L^\infty(\Theta\times [0,T_{\mathrm{obs}}])})^2
 L_{\mathfrak F}^2\norm{\theta-\theta'}_2^2\right\}.
\]
For the Gaussian part, conditionally on \(t\),
\[
 \E_\xi e^{-2\lambda\sigma\langle\Delta\mathfrak F,\xi\rangle}
 =e^{2\lambda^2\sigma^2\norm{\Delta\mathfrak F}_2^2}
 \le e^{2\lambda^2\sigma^2L_{\mathfrak F}^2
 \norm{\theta-\theta'}_2^2}.
\]
Combining the last two displays by conditioning on \(t\) yields the
centered increment bound
\[
 \norm{\bar\ell_1(\theta,t,\xi)-\bar\ell_1(\theta',t,\xi)}_{\mathrm{SG}}
 \le B_1\norm{\theta-\theta'}_2,
\]
with \(B_1\) as in \eqref{eq:B1}. Applying the sub-Gaussian process
concentration result in Lemma~\ref{lem:bg-subgaussian-process} over the
\(p\)-dimensional compact set \(\Theta\), whose diameter is \(D\), gives,
with probability \(1-\delta\),
\[
 \sup_{\theta\in\Theta}\left|\frac1n\sum_{i=1}^n
 [\ell_{1}(\theta,t_i,\xi_i)-\E\ell_1(\theta,t,\xi)]\right|
 \le C_RB_0\sqrt{\frac{p\log(2\kappa)+\log(1/\delta)}n}.
\]
Finally, \(\sum_i\norm{\xi_i}_2^2\sim\chi^2_{nm}\).
Lemma~\ref{lem:bg-laurent-massart} and \(n\ge4\log(2/\delta)/m\) imply, with
probability \(1-\delta\),
\[
 \left|\frac{\sigma^2}{n}\sum_{i=1}^n(\norm{\xi_i}_2^2-m)\right|
 \le3\sigma^2\sqrt{\frac{m\log(2/\delta)}n}.
\]
The triangle inequality and a union bound complete the proof.
\end{proof}

We now prove similar concentration result for the hessian which allows for preservation of local curvature.
For this purpose, define the following quantities
\begin{align}
 \overline H&:=2\norm{\nabla_\theta x}_{L^\infty(\Theta\times [0,T_{\mathrm{obs}}])}^2,\\
 \overline J&:=(4+2m)\norm{\nabla_\theta x}_{L^\infty(\Theta\times [0,T_{\mathrm{obs}}])}
 \norm{\nabla_\theta^2x}_{L^\infty(\Theta\times [0,T_{\mathrm{obs}}])}
 +2m(2\norm{x}_{L^\infty(\Theta\times [0,T_{\mathrm{obs}}])}+\sigma)L_H,
 \\\tau^2&:=\max\Bigg\{1,
 4\norm{\nabla_\theta x}_{L^\infty(\Theta\times [0,T_{\mathrm{obs}}])}^2
 [\sigma^2+(2\norm{x}_{L^\infty(\Theta\times [0,T_{\mathrm{obs}}])})^2],
 4\sigma\sqrt m\norm{\nabla_\theta^2x}_{L^\infty(\Theta\times [0,T_{\mathrm{obs}}])},\notag\\
 &\frac4{\log(7/6)}\left[\norm{\nabla_\theta x}_{L^\infty(\Theta\times [0,T_{\mathrm{obs}}])}^2
 +2m\norm{x}_{L^\infty(\Theta\times [0,T_{\mathrm{obs}}])}\norm{\nabla_\theta^2x}_{L^\infty(\Theta\times [0,T_{\mathrm{obs}}])}\right],
 \overline H,\overline J^{2/3}\Bigg\}.
 \label{eq:tau-scale}
\end{align}

\begin{lemma}[Uniform concentration of the derivatives]
\label{lem:uniform-derivative-concentration}
Let $\delta\in(0,1)$. Then there is a universal \(C_0>0\) such that for
\(C_\delta:=C_0\left(1+\log\frac{1+D\tau}{2\delta}\right)\) and
\(n\ge \max\{2,C_\delta p\log p\}\), with probability at least
\(1-2\delta\),
\begin{align}
 \sup_{\theta\in\Theta}\norm{\nabla_\theta \mathcal R_{\theta^*}^{(n)}(\theta)
 -\nabla_\theta \mathcal{R}_{\theta^*}(\theta)}_2
 &\le\tau\sqrt{\frac{C_\delta p\log n}{n}},
 \label{eq:uniform-gradient-concentration}\\
 \sup_{\theta\in\Theta}\norm{\nabla_\theta^2\mathcal R_{\theta^*}^{(n)}(\theta)
 -\nabla_\theta^2\mathcal{R}_{\theta^*}(\theta)}_{\mathrm{op}}
 &\le\tau^2\sqrt{\frac{C_\delta p\log n}{n}}.
 \label{eq:uniform-hessian-concentration}
\end{align}
\end{lemma}

\begin{proof}[Proof of Lemma~\ref{lem:uniform-derivative-concentration}]
We aim to apply Lemma~\ref{lem:bg-mei-concentration}. For the ideal one-observation loss
\(\ell( {\theta,t},\xi)=\norm{x( {\theta,t})-x( {\theta^*,t})-\sigma\xi}_2^2\),
direct differentiation gives
\begin{align*}
 \nabla_\theta\ell
 &=2\nabla_\theta x(x-x^*-\sigma\xi),\\
 \nabla_\theta^2\ell
 &=2\nabla_\theta x\nabla_\theta x^\top
 +2\sum_{j=1}^m(x_j-x_j^*-\sigma\xi_j)\nabla_\theta^2x_j.
\end{align*}
For the gradient, define
\[
 A( {\theta,t})=2\nabla_\theta x( {\theta,t})
 [x( {\theta,t})-x( {\theta^*,t})],
 \qquad
 B( {\theta,t})=2\sigma\nabla_\theta x( {\theta,t}).
\]
Thus \(\nabla_\theta\ell=A-B\xi\). For every \(u\in\R^p\), conditioning on
\(t\) gives
\begin{align}
 &\E\exp\{\langle u,\nabla_\theta\ell-\E\nabla_\theta\ell\rangle\}\notag\\
 &=\E_t\!\left[
 e^{\langle u,A-\E_tA\rangle}
 \E_\xi e^{-\langle u,B\xi\rangle}\right]
 =\E_t\!\left[
 e^{\langle u,A-\E_tA\rangle}e^{\norm{B^\top u}_2^2/2}\right].
 \label{eq:appendix-gradient-mgf}
\end{align}
The uniform bounds imply
\[
 \norm{B^\top u}_2^2
 \le4\sigma^2\norm{\nabla_\theta x}_{L^\infty(\Theta\times [0,T_{\mathrm{obs}}])}^2\norm u_2^2.
\]
Moreover, \(\langle u,A\rangle\) lies in an interval of length at most
\(4\norm u_2\norm{\nabla_\theta x}_{L^\infty(\Theta\times [0,T_{\mathrm{obs}}])}
(2\norm{x}_{L^\infty(\Theta\times [0,T_{\mathrm{obs}}])})\). Hoeffding's lemma and
\eqref{eq:appendix-gradient-mgf} therefore give
\[
 \E e^{\langle u,\nabla_\theta\ell-\E\nabla_\theta\ell\rangle}
 \le\exp\!\left\{
 2\norm u_2^2\norm{\nabla_\theta x}_{L^\infty(\Theta\times [0,T_{\mathrm{obs}}])}^2
 [\sigma^2+(2\norm{x}_{L^\infty(\Theta\times [0,T_{\mathrm{obs}}])})^2]\right\},
\]
which is the required sub-Gaussian gradient condition at scale \(\tau\).

For the hessian condition, consider a unit \(u\in\R^p\) and set
\[
 c_j( {\theta,t})=-2\sigma u^\top\nabla_\theta^2x_j( {\theta,t})u
\]
and write
\[
 u^\top\nabla_\theta^2\ell u
 =D( {\theta,t})+\langle c( {\theta,t}),\xi\rangle,
\]
where
\[
 D({\theta,t})=2\norm{\nabla_\theta x( {\theta,t})^\top u}_2^2
 -\sigma^{-1}\sum_{j=1}^m[x_j( {\theta,t})-x_j( {\theta^*,t})]c_j,
\]
The displayed formulas give
\begin{align*}
 \norm{c(\theta,t)}_2&\le2\sigma\sqrt m\norm{\nabla_\theta^2x}_{L^\infty(\Theta\times [0,T_{\mathrm{obs}}])},
 \\
 |D(\theta,t)|&\le2\norm{\nabla_\theta x}_{L^\infty(\Theta\times [0,T_{\mathrm{obs}}])}^2
 +4m\norm{x}_{L^\infty(\Theta\times [0,T_{\mathrm{obs}}])}\norm{\nabla_\theta^2x}_{L^\infty(\Theta\times [0,T_{\mathrm{obs}}])}.
\end{align*}
For \(s>0\), the folded-normal MGF is
\[
 \E_\xi e^{|\langle c,\xi\rangle|/s}
 =2e^{\norm c_2^2/(2s^2)}\Phi(\norm c_2/s),
\]
which is at most \(12/7\) whenever
\(s\ge4\sigma\sqrt m\norm{\nabla_\theta^2x}_{L^\infty(\Theta\times [0,T_{\mathrm{obs}}])}\). Also,
\[
 \E_t e^{|D-\E_tD|/s}
 \le\exp\!\left\{\frac4s\left[
 \norm{\nabla_\theta x}_{L^\infty(\Theta\times [0,T_{\mathrm{obs}}])}^2+
 2m\norm{x}_{L^\infty(\Theta\times [0,T_{\mathrm{obs}}])}\norm{\nabla_\theta^2x}_{L^\infty(\Theta\times [0,T_{\mathrm{obs}}])}\right]\right\},
\]
which is at most \(7/6\) under the fourth lower bound on \(s\) in
\eqref{eq:tau-scale}. Since
\(|u^\top\nabla_\theta^2\ell\,u-\E u^\top\nabla_\theta^2\ell\,u|\le|D-\E_tD|+|\langle c,\xi\rangle|\), conditioning first on
\(t\) and taking \(s=\tau^2\) imply
\[
 \E\exp\!\left(
 \frac{|u^\top(\nabla_\theta^2\ell-\E\nabla_\theta^2\ell)u|}{\tau^2}
 \right)\le2,
\]
because \((7/6)(12/7)=2\). Thus every centered directional Hessian is
sub-exponential with scale \(\tau^2\).

It remains to control increments. Fix \(\theta_1,\theta_2\in\Theta\) and
write \(J_k=\nabla_\theta x( {\theta_k,t})\),
\(H_{j,k}=\nabla_\theta^2x_j( {\theta_k,t})\), and
\(r_{j,k}=x_j( {\theta_k,t})-x_j( {\theta^*,t})-\sigma\xi_j\). First,
\begin{align*}
 \norm{J_1J_1^\top-J_2J_2^\top}_{\mathrm{op}}
 &\le(\norm{J_1}_{\mathrm{op}}+\norm{J_2}_{\mathrm{op}})
 \norm{J_1-J_2}_{\mathrm{op}}\\
 &\le2\norm{\nabla_\theta x}_{L^\infty(\Theta\times [0,T_{\mathrm{obs}}])}
 \norm{\nabla_\theta^2x}_{L^\infty(\Theta\times [0,T_{\mathrm{obs}}])}\norm{\theta_1-\theta_2}_2.
\end{align*}
For each coordinate,
\[
 r_{j,1}H_{j,1}-r_{j,2}H_{j,2}
 =r_{j,1}(H_{j,1}-H_{j,2})
 +[x_j( {\theta_1,t})-x_j( {\theta_2,t})]H_{j,2}.
\]
The mean-value theorem yields
\begin{align*}
 \norm{H_{j,1}-H_{j,2}}_{\mathrm{op}}
 &\le{ L_H}\norm{\theta_1-\theta_2}_2,\\
 |x_j( {\theta_1,t})-x_j( {\theta_2,t})|
 &\le\norm{\nabla_\theta x}_{L^\infty(\Theta\times [0,T_{\mathrm{obs}}])}\norm{\theta_1-\theta_2}_2,\\
 |r_{j,1}|&\le2\norm{x}_{L^\infty(\Theta\times [0,T_{\mathrm{obs}}])}+\sigma|\xi_j|.
\end{align*}
Substitution into the Hessian formula and
\(\sum_j|\xi_j|\le\sqrt m\norm{\xi}_2\) gives
\[
 \norm{\nabla_\theta^2\ell( {\theta,t},\xi)
 -\nabla_\theta^2\ell(\theta',t,\xi)}_{\mathrm{op}}
 \le L(t,\xi)\norm{\theta-\theta'}_2,
\]
where
\begin{align*}
 L(t,\xi):={}&(4+2m)\norm{\nabla_\theta x}_{L^\infty(\Theta\times [0,T_{\mathrm{obs}}])}
 \norm{\nabla_\theta^2x}_{L^\infty(\Theta\times [0,T_{\mathrm{obs}}])}\\
 &+2[2m\norm{x}_{L^\infty(\Theta\times [0,T_{\mathrm{obs}}])}+\sigma\sqrt m\norm{\xi}_2]
 { L_H}.
\end{align*}
Since \(\E\norm{\xi}_2\le\sqrt m\), we have \(\E L\le\overline J\), while
\(\norm{\nabla_\theta^2\mathcal{R}_{\theta^*}(\theta^*)}_{\mathrm{op}}\le\overline H\).
The last two entries of \eqref{eq:tau-scale} therefore put these quantities
on the same scale: \(\overline H\le\tau^2\) and \(\overline J\le\tau^3\).
Now we apply Lemma~\ref{lem:bg-mei-concentration}, which gives
\eqref{eq:uniform-hessian-concentration}.
\end{proof}

The two previous results alone are enough to ensure convergence of our empirical minimization algorithm, but we may prove a tighter excess risk bound for the empirical minimizer \(\theta_n\in\operatorname*{argmin}_{\theta\in\Theta}\mathcal R_{\theta^*}^{(n)}(\theta)\) using standard localization arguments, improving the final error bound of our algorithm.

\begin{lemma}[Tighter population-risk bound]
\label{lem:tighter-population-risk}
Let $n>1$. Fix \(\alpha\in(0,1)\). With probability at
least \(1-2\alpha\),
\begin{equation}
 \mathcal{R}_{\theta^*}(\theta_n)-\mathcal{R}_{\theta^*}(\theta^*)
 \lesssim C(\cdot)\bigg[\frac{(2p+1)\log n}{n}
 +\frac{(2p+1)\log\sqrt m+\log(1/\alpha)}{n}\bigg],
 \label{eq:tighter-population-risk}
\end{equation}
where the implicit constant is universal. Moreover, $C(\cdot)$ is quadratic order in $\sigma$ and $\|x\|_{L^\infty(\Theta\times [0,T_{\mathrm{obs}}])}$, and logarithmic order in $\|\nabla_\theta x\|_{L^\infty(\Theta\times [0,T_{\mathrm{obs}}])}$ and $D$.
\end{lemma}

\begin{proof}[Proof of Lemma~\ref{lem:tighter-population-risk}]
We apply Lemma~\ref{lem:vector-oracle} to the prediction class
\(\mathcal X:=\{x( {\theta,\cdot}):\theta\in\Theta\}\), so that
\[
 \partial x
 :=\mathcal X-\mathcal X
 =\{x( {\theta,\cdot})-x( {\theta',\cdot}):\theta,\theta'\in\Theta\},
 \qquad
 \mathcal F:=\operatorname{star}(\partial x).
\]
Since \(Y_i=x( {\theta^*,t_i})+\sigma\xi_i\) and \(\theta_n\) minimizes
\(\mathcal R_{\theta^*}^{(n)}\) over \(\Theta\), the pair \((f^*,\widehat f)=(x( {\theta^*,\cdot}),
x( {\theta_n,\cdot}))\) satisfies the conditions of
Lemma~\ref{lem:vector-oracle} with \(\mathcal X\) and \(\mathcal F\) as
above. For this we require an explicit sufficient Gaussian critical
radius \(\delta_{G,n}\) for \(\mathcal F\), which we calculate as
follows.

The map
\[
(\beta,\theta,\theta')\mapsto
\beta\bigl(x( {\theta,\cdot})-x( {\theta',\cdot})\bigr)
\]
from $[0,1]\times\Theta^2$ onto
$\mathcal{F}$ satisfies
\begin{align}
    &\|\beta(x( {\theta,\cdot})-x( {\theta',\cdot}))
      -\bar\beta(x( {\bar\theta,\cdot})-
                         x( {\bar\theta',\cdot}))\|_n \notag\\
    &\qquad\le
      \sqrt{4\|x\|_{L^\infty(\Theta\times [0,T_{\mathrm{obs}}])}^2
            +2\|\nabla_\theta x\|_{L^\infty(\Theta\times [0,T_{\mathrm{obs}}])}^2}
      \|(\beta,\theta,\theta')-
         (\bar\beta,\bar\theta,\bar\theta')\|_2.
      \label{eq:gaussian-parametrization-lipschitz}
\end{align}
The domain has dimension $2p+1$ and diameter $\sqrt{1+2D^2}$.
Constructing a Euclidean net directly in the preimage of the localized
ball and applying the Lipschitz bound gives
\begin{align}
    N_n\bigl(u;B_n(\delta;
      \mathcal F)\bigr)
    \le\left(1+\frac{1}{u}
      \sqrt{4\|x\|_{L^\infty(\Theta\times [0,T_{\mathrm{obs}}])}^2
            +2\|\nabla_\theta x\|_{L^\infty(\Theta\times [0,T_{\mathrm{obs}}])}^2}
      \sqrt{1+2D^2}\right)^{2p+1}.
      \label{eq:gaussian-covering-number}
\end{align}
For $0<\delta\le4\sigma\sqrt m$, choose
$\varepsilon=\delta^2/(4\sigma\sqrt m)$. The vector-analog of the truncated Dudley bound from
Wainwright~\cite[Corollary~13.7]{wainwright2019} shows that
\eqref{eq:gaussian-critical-inequality} holds whenever
\begin{equation}
    \label{eq:gaussian-dudley-condition}
    \frac{16}{\sqrt n}\int_{\delta^2/(4\sigma\sqrt m)}^\delta
      \sqrt{\log N_n\bigl(u;B_n(\delta;
      \mathcal F)\bigr)}\,du
    \le\frac{\delta^2}{4\sigma}.
\end{equation}

Bounding the integrand by its value at the lower endpoint reduces this to
\begin{align}
    \frac{16\delta}{\sqrt n}\Bigg[(2p+1)\log\Bigg(&1+
      \frac{4\sigma\sqrt m}{\delta^2}
      \sqrt{4\|x\|_{L^\infty(\Theta\times [0,T_{\mathrm{obs}}])}^2
            +2\|\nabla_\theta x\|_{L^\infty(\Theta\times [0,T_{\mathrm{obs}}])}^2}
      \sqrt{1+2D^2}\Bigg)\Bigg]^{1/2}
    \le\frac{\delta^2}{4\sigma}.
    \label{eq:gaussian-radius-inequality}
\end{align}
Put $\Delta=\delta^2/\sigma^2$. After squaring, it is enough that
\begin{align}
    \log\Bigg(1+\frac{4\sqrt m}{\sigma\Delta}
      \sqrt{4\|x\|_{L^\infty(\Theta\times [0,T_{\mathrm{obs}}])}^2
            +2\|\nabla_\theta x\|_{L^\infty(\Theta\times [0,T_{\mathrm{obs}}])}^2}
      \sqrt{1+2D^2}\Bigg)
    \le\frac{n\Delta}{4096(2p+1)}.
    \label{eq:gaussian-delta-inequality}
\end{align}
Since $\log(1+a/\Delta)\le\log(1+an)\le\log(e+an)$ when
$\Delta\ge1/n$, an explicit sufficient choice is
\begin{align}
    \delta_{G,n}^2
    :=\frac{4096\sigma^2(2p+1)}{n}
      \log\Bigg(&e+\frac{4n\sqrt m}{\sigma}
      \sqrt{4\|x\|_{L^\infty(\Theta\times [0,T_{\mathrm{obs}}])}^2
            +2\|\nabla_\theta x\|_{L^\infty(\Theta\times [0,T_{\mathrm{obs}}])}^2}
      \sqrt{1+2D^2}\Bigg),
      \label{eq:explicit-gaussian-radius}
\end{align}

\paragraph{Note on large Gaussian radii.}
The truncation above is admissible whenever
\(\delta_{G,n}\le4\sigma\sqrt m\), since then
\(\delta_{G,n}^2/(4\sigma\sqrt m)\le\delta_{G,n}\).
If \(\delta_{G,n}>4\sigma\sqrt m\), Cauchy--Schwarz and Jensen give
\[
 \mathfrak{G}_n(\delta_{G,n};\mathcal F)
 \le\delta_{G,n}\,\E_w
 \left(\frac1n\sum_i\norm{w_i}_2^2\right)^{1/2}
 \le\sqrt m\,\delta_{G,n}
 \le\frac{\delta_{G,n}^2}{2\sigma}.
\]
Thus \eqref{eq:explicit-gaussian-radius} satisfies
\eqref{eq:gaussian-critical-inequality} for every sample size.
\paragraph{Note on vector-analog of Dudley truncation}
The slight vector modification in Wainwright~\cite[Corollary~13.7]{wainwright2019}
is that if \(\norm{g-g_j}_n\le\varepsilon\), then
\begin{align}
 \E_w\left|\frac1n\sum_{i=1}^n
 \langle w_i,g(t_i)-g_j(t_i)\rangle\right|
 &\le\E_w\left(\frac1n\sum_{i=1}^n\norm{w_i}_2^2\right)^{1/2}
 \norm{g-g_j}_n
 \le\sqrt m\,\varepsilon.
 \label{eq:vector-dudley-factor}
\end{align}

Now apply \ref{lem:vector-oracle} with \(\gamma=1/2\) and
\[
 r:=\max\left\{\delta_{G,n},
 \frac{8\sigma^2}{n\delta_{G,n}}\log\frac1\alpha\right\}.
\]
Its failure probability is at most \(\alpha\), and
\[
 r\delta_{G,n}
 \le\delta_{G,n}^2+\frac{8\sigma^2}n\log\frac1\alpha.
\]
Choosing \(\theta=\theta^*\) in the oracle infimum makes its
approximation term exactly zero, because
\(x( {\theta^*,\cdot})=x( {\theta^*,\cdot})\). Hence
\begin{align}
 \norm{x( {\theta_n,\cdot})-x( {\theta^*,\cdot})}_n^2
 \le{}&4c_{G,0}\left(\delta_{G,n}^2
 +\frac{8\sigma^2}n\log\frac1\alpha\right).
 \label{eq:oracle-at-truth}
\end{align}
The bound holds for every $t_i$, so a conditional
failure bound of \(\alpha\) also gives unconditional failure probability
at most \(\alpha\), by averaging over the observation times.
\\For the empirical-to-population transfer, define
\begin{equation}
 \mathcal H^*
 :=\{t\mapsto\beta[x( {\theta,t})-x( {\theta^*,t})]:
       \beta\in[0,1],\ \theta\in\Theta\},
 \label{eq:error-star-class}
\end{equation}
and set
\[
 b:=2\norm{x}_{L^\infty(\Theta\times [0,T_{\mathrm{obs}}])},\qquad
 \mathcal K:=\{t\mapsto\norm{h(t)}_2:h\in\mathcal H^*\}.
\]
The class \(\mathcal H^*\) has envelope \(b\). We calculate a radius
satisfying the critical inequality in Lemma~\ref{lem:vector-uniform-law},
namely \(\mathfrak{R}_n^{\mathrm{vec}}(\delta;\mathcal H^*)\le\delta^2/b\).
For this we first consider the empirical counterpart: \(\widehat{\mathfrak{R}}_n^{\mathrm{vec}}(\delta;\mathcal H^*)\le\delta^2/b\) The reverse triangle inequality and the mean-value theorem give
\[
 \left|\norm{x( {\theta,t})-x( {\theta^*,t})}_2
       -\norm{x( {\theta',t})-x( {\theta^*,t})}_2\right|
 \le\norm{\nabla_\theta x}_{L^\infty(\Theta\times [0,T_{\mathrm{obs}}])}\norm{\theta-\theta'}_2.
\]
Consequently, the parametrization
\((\beta,\theta)\mapsto\beta\norm{x( {\theta,\cdot})-x( {\theta^*,\cdot})}_2\)
of \(\mathcal K\) is Lipschitz with constant
\[
 \sqrt{b^2+\norm{\nabla_\theta x}_{L^\infty(\Theta\times [0,T_{\mathrm{obs}}])}^2},
\]
has dimension \(p+1\), and has domain diameter \(\sqrt{1+D^2}\).
The same parameter-space covering argument gives
\begin{equation}
 N_n(u;B_n(\delta;\mathcal K))
 \le\left(1+\frac1u
 \sqrt{b^2+\norm{\nabla_\theta x}_{L^\infty(\Theta\times [0,T_{\mathrm{obs}}])}^2}
 \sqrt{1+D^2}\right)^{p+1}.
 \label{eq:rademacher-covering-number}
\end{equation}
For \(0<\delta\le4b\), take \(u_0=\delta^2/(4b)\le\delta\).
Conditionally on the observation times, the Rademacher process has
sub-Gaussian increments in the empirical metric, since
\[
 \E_\varepsilon\exp\left\{\frac\lambda n\sum_i\varepsilon_i
 [k(t_i)-k'(t_i)]\right\}
 \le\exp\left\{\frac{\lambda^2}{2n}\norm{k-k'}_n^2\right\}.
\]
Thus the same truncated Dudley bound from
the proof of Wainwright~\cite[Corollary~13.7]{wainwright2019} applies. In this case, it yields
\[
 \E_\varepsilon\sup_{k\in B_n(\delta;\mathcal K)}
 \left|\frac1n\sum_i\varepsilon_i k(t_i)\right|
 \le\frac{\delta^2}{4b}
 +\frac{16}{\sqrt n}\int_{\delta^2/(4b)}^\delta
       \sqrt{\log N_n(u;B_n(\delta;\mathcal K))}\,du.
\]
The covering bound holds for every $t_i$. In particular, the empirical counterpart of
\eqref{eq:rademacher-critical-inequality} holds if
\begin{equation}
 \frac{16\delta}{\sqrt n}
 \left[(p+1)\log\left(1+\frac{4b}{\delta^2}
 \sqrt{b^2+\norm{\nabla_\theta x}_{L^\infty(\Theta\times [0,T_{\mathrm{obs}}])}^2}
 \sqrt{1+D^2}\right)\right]^{1/2}
 \le\frac{\delta^2}{4b}.
 \label{eq:rademacher-radius-inequality}
\end{equation}
Writing \(\Delta=\delta^2/b^2\), squaring reduces this condition to
\[
 \log\left(1+\frac4{b\Delta}
 \sqrt{b^2+\norm{\nabla_\theta x}_{L^\infty(\Theta\times [0,T_{\mathrm{obs}}])}^2}\sqrt{1+D^2}\right)
 \le\frac{n\Delta}{4096(p+1)}.
\]
An explicit sufficient choice is therefore
\begin{equation}
 d_{R,n}^2
 :=\frac{4096b^2(p+1)}n
 \log\left(e+\frac{4n}b
 \sqrt{b^2+\norm{\nabla_\theta x}_{L^\infty(\Theta\times [0,T_{\mathrm{obs}}])}^2}\sqrt{1+D^2}\right).
 \label{eq:explicit-rademacher-radius}
\end{equation}
Indeed, this choice has \(\Delta\ge1/n\), and hence
\(\log(1+a/\Delta)\le\log(1+an)\le\log(e+an)\).
The remainder and integral each contribute at most \(d_{R,n}^2/(4b)\),
so their sum is at most \(d_{R,n}^2/(2b)\le d_{R,n}^2/b\).

\paragraph{Note on large Rademacher radii.}
If \(d_{R,n}>4b\), the envelope bound gives directly
\[
 \E_\varepsilon\sup_{k\in B_n(d_{R,n};\mathcal K)}
 \left|\frac1n\sum_i\varepsilon_i k(t_i)\right|
 \le b\le\frac{d_{R,n}^2}{b}.
\]
Thus the conditional empirical critical inequality holds at \(d_{R,n}\)
for every $t_i$ and $n$.
\\Now in order to solve the population-level critical inequality, we proceed as follows.
Let \(r_n\) and \(\widehat r_n\) denote the population and conditional
empirical critical radii of \(\mathcal K\), with target \(r^2/b\).
The calculation above gives
\(\widehat r_n\le d_{R,n}\) for every $t_i$. Proposition~14.25 of
Wainwright~\cite{wainwright2019}, rescaled by \(b\), gives
\(r_n\le4\widehat r_n\le 4d_{R,n}\) with probability at least
\(1-c_1e^{-c_2nr_n^2/b^2}\). Note that if
\(r_n>C d_{R,n}\) for a sufficiently large universal \(C>4\), it makes this
probability positive, contradicting $r_n\leq 4d_{R,n}$.
Thus \(r_n\le C d_{R,n}\) and \(\delta_{R,n}:=C d_{R,n}\) satisfies the population critical
inequality \eqref{eq:rademacher-critical-inequality}.

Now we apply Lemma~\ref{lem:vector-uniform-law} to
\(h=x( {\theta_n,\cdot})-x( {\theta^*,\cdot})\in\mathcal H^*\) with
\[
 s^2:=\max\left\{\delta_{R,n}^2,
 \frac{(2\norm{x}_{L^\infty(\Theta\times [0,T_{\mathrm{obs}}])})^2}{c_4n}\log\frac{c_3}\alpha\right\}.
\]
Its failure probability is at most \(\alpha\), and
\begin{align}
 \norm{x( {\theta_n,\cdot})-x( {\theta^*,\cdot})}_{L^2}^2
 \le{}&2\norm{x( {\theta_n,\cdot})-x( {\theta^*,\cdot})}_n^2
 +\delta_{R,n}^2\notag\\
 &+\frac{(2\norm{x}_{L^\infty(\Theta\times [0,T_{\mathrm{obs}}])})^2}{c_4n}\log\frac{c_3}\alpha.
 \label{eq:population-transfer-at-estimator}
\end{align}

Recall that
 $\mathcal{R}_{\theta^*}(\theta_n)-\mathcal{R}_{\theta^*}(\theta^*)
 =\norm{x( {\theta_n,\cdot})-x( {\theta^*,\cdot})}_{L^2}^2.
$
Combining \eqref{eq:oracle-at-truth},
\eqref{eq:population-transfer-at-estimator} with this, and taking a union bound
gives the critical-radius bound. Substituting
\eqref{eq:explicit-gaussian-radius} and
\eqref{eq:explicit-rademacher-radius} yields
\eqref{eq:tighter-population-risk}.
\end{proof}

Now we may combine the previous results by using the local strong convexity and excess risk bound to localize the parameter. For this, define
\begin{align*}
 \eta_n
 &:=\sup_{\theta\in\Theta}
 \norm{\nabla_\theta^2\mathcal R_{\theta^*}^{(n)}(\theta)-\nabla_\theta^2\mathcal{R}_{\theta^*}(\theta)}_{\mathrm{op}},
 \\
 \varepsilon_n^{(SS)}(\alpha)
 &:=\frac{(2p+1)\log n+(2p+1)\log\sqrt m+\log(1/\alpha)}n,
\end{align*}
and let \(C_{SS}>0\) denote the implicit constant in
Lemma~\ref{lem:tighter-population-risk} multiplied by $C(\cdot)$. Also define the \emph{isolating radius}
\begin{equation}
 \rho':=\min\left\{\frac\rho4,
 \sqrt{\frac{2\gamma_\rho}{\mu}}\right\}>0.
 \label{eq:isolating-radius}
\end{equation}

\begin{corollary}[Parameter localization and empirical curvature]
\label{cor:empirical-localization-curvature}
Fix \(\alpha,\delta\in(0,1)\) and suppose the sample-size conditions in
Lemma
\ref{lem:uniform-derivative-concentration} holds. On the intersection of the events in Lemma
\ref{lem:uniform-derivative-concentration} and \ref{lem:tighter-population-risk}, suppose
\begin{equation}
 C_{SS}\varepsilon_n^{(SS)}(\alpha)
 <\frac\mu2(\rho')^2,
 \qquad
 \tau^2\sqrt{\frac{C_\delta p\log n}{n}}<\mu.
 \label{eq:empirical-preservation-conditions}
\end{equation}
Then every global minimizer \(\theta_n\) of \(\mathcal R_{\theta^*}^{(n)}\) satisfies
\begin{equation}
 \norm{\theta_n-\theta^*}_2
 \le\sqrt{\frac{2C_{SS}}{\mu}\varepsilon_n^{(SS)}(\alpha)}
 <\rho'\le\rho/4,
 \label{eq:empirical-minimizer-localization}
\end{equation}
and
\begin{equation}
 \nabla_\theta^2\mathcal R_{\theta^*}^{(n)}(\theta)
 \succeq\left(\mu-\tau^2
 \sqrt{\frac{C_\delta p\log n}{n}}\right)I_p\succ0
 \quad\text{for every }\theta\in B(\theta^*,\rho).
 \label{eq:empirical-strong-convexity-preserved}
\end{equation}
Consequently, the global empirical minimizer is unique.
\end{corollary}

\begin{proof}[Proof of Corollary~\ref{cor:empirical-localization-curvature}]
We first establish the localization implication needed below. Let
\(0<r<\rho'\) and suppose that \(\bar\theta\in\Theta\) satisfies
\begin{equation}
 \mathcal{R}_{\theta^*}(\bar\theta)-\mathcal{R}_{\theta^*}(\theta^*)\le\frac\mu2r^2.
 \label{eq:appendix-small-excess-risk}
\end{equation}
If \(\bar\theta\notin B(\theta^*,\rho/4)\), the definition of
\(\gamma_\rho\) and \eqref{eq:isolating-radius} give
\[
 \mathcal{R}_{\theta^*}(\bar\theta)-\mathcal{R}_{\theta^*}(\theta^*)\ge\gamma_\rho
 \ge\frac\mu2(\rho')^2.
\]
When \(r<\rho'\), this contradicts
\eqref{eq:appendix-small-excess-risk}; hence
\(\bar\theta\in B(\theta^*,\rho/4)\).

The segment joining \(\theta^*\) and \(\bar\theta\) lies in \(\Theta\)
because \(\Theta\) is convex, and it remains inside
\(B(\theta^*,\rho)\). Since \(\theta^*\) minimizes \(\mathcal{R}_{\theta^*}\) in the interior,
\(\nabla_\theta \mathcal{R}_{\theta^*}(\theta^*)=0\). Taylor's formula with integral remainder
and assumption \ref{ass:quantitative-identifiability} therefore yield
\begin{align*}
 \mathcal{R}_{\theta^*}(\bar\theta)-\mathcal{R}_{\theta^*}(\theta^*)
 &=\int_0^1(1-s)
 (\bar\theta-\theta^*)^\top
 \nabla_\theta^2\mathcal{R}_{\theta^*}(\theta^*+s(\bar\theta-\theta^*))
 (\bar\theta-\theta^*)\,ds\\
 &\ge\frac\mu2\norm{\bar\theta-\theta^*}_2^2.
\end{align*}
Together with \eqref{eq:appendix-small-excess-risk}, this proves
\(\norm{\bar\theta-\theta^*}_2\le r\).

We now apply this implication to the empirical minimizer. On the event of
Lemma~\ref{lem:tighter-population-risk},
\[
 \mathcal{R}_{\theta^*}(\theta_n)-\mathcal{R}_{\theta^*}(\theta^*)
 \le C_{SS}\varepsilon_n^{(SS)}(\alpha).
\]
Set
\[
 r_n:=\sqrt{\frac{2C_{SS}}{\mu}
 \varepsilon_n^{(SS)}(\alpha)}.
\]
The first inequality in \eqref{eq:empirical-preservation-conditions}
gives \(r_n<\rho'\), and the preceding excess-risk bound is precisely
\(\mathcal{R}_{\theta^*}(\theta_n)-\mathcal{R}_{\theta^*}(\theta^*)\le\mu r_n^2/2\). The localization implication
therefore proves \eqref{eq:empirical-minimizer-localization}. The argument
applies to every global empirical minimizer.

On the event of Lemma~\ref{lem:uniform-derivative-concentration},
\[
 \eta_n
 \le\tau^2\sqrt{\frac{C_\delta p\log n}{n}}.
\]
For every \(\theta\in B(\theta^*,\rho)\), Weyl's inequality and
assumption \ref{ass:quantitative-identifiability} give
\begin{align*}
 \lambda_{\min}(\nabla_\theta^2\mathcal R_{\theta^*}^{(n)}(\theta))
 &\ge\lambda_{\min}(\nabla_\theta^2\mathcal{R}_{\theta^*}(\theta))-\eta_n\\
 &\ge\mu-\tau^2\sqrt{\frac{C_\delta p\log n}{n}}>0.
\end{align*}
This proves \eqref{eq:empirical-strong-convexity-preserved}. Thus \(\mathcal R_{\theta^*}^{(n)}\)
is strictly convex on the convex ball \(B(\theta^*,\rho)\). Every global
minimizer lies in the smaller ball \(B(\theta^*,\rho')\), so two distinct
global minimizers cannot exist.

\end{proof}

\subsection{Coarse-initialization followed by GD algorithm guarantees}
\label{subsec:empirical-minimization}

Given the uniform concentration of $\widetilde{\mathcal{R}}_{\theta^*}^{(n)}$ to $\mathcal{R}_{\theta^*}$ we prove it is possible to initialize into the strongly convex neighborhood using the numerical empirical risk, and using $\nabla^2_\theta \mathcal R_{\theta^*}^{(n)}$ concentrating to $\nabla^2_\theta \mathcal{R}_{\theta^*}$, we prove an error bound on the subsequent approximate gradient descent due to preservation of local curvature.

\begin{figure}[htbp]
\centering
\begin{minipage}[c]{0.48\linewidth}
\centering
\begin{tikzpicture}[x=1.45cm,y=1.65cm]
  \draw[->] (-1.6,0) -- (1.65,0) node[right] {\(x\)};
  \draw[->] (-1.55,-0.12) -- (-1.55,1.2);
  \node[above] at (-1.55,1.2) {\(w\)};
  \draw[densely dashed,gray] (-1,0) -- (-1,1);
  \draw[densely dashed,gray] (1,0) -- (1,1);
  \draw[blue,thick] (-1.5,0) -- (-1,0);
  \draw[blue,thick,domain=-1:0,samples=40] plot (\x,{0.5*(\x+1)^2});
  \draw[blue,thick,domain=0:1,samples=40] plot (\x,{1-0.5*(1-\x)^2});
  \draw[blue,thick] (1,1) -- (1.5,1);
  \node[below] at (-1,0) {\(y-\tau_{\mathrm{sel}}\)};
  \node[below] at (0,0) {\(y\)};
  \node[below] at (1,0) {\(y+\tau_{\mathrm{sel}}\)};
  \node[left] at (-1.55,1) {1};
  \node[left] at (-1.55,0) {0};
\end{tikzpicture}
\end{minipage}\hfill
\begin{minipage}[c]{0.51\linewidth}
\centering
\begin{tikzpicture}[x=1cm,y=0.72cm, every node/.style={font=\scriptsize}]
  \draw[->] (0.72,1.5) -- (1.3,1.08);
  \draw[->] (0.72,0.5) -- (1.3,0.82);
  \draw[->] (0.72,-0.5) -- (1.3,-0.82);
  \draw[->] (0.72,-1.5) -- (1.3,-1.08);
  \draw[->] (2.72,0.95) -- (3.3,0.16);
  \draw[->] (2.72,-0.95) -- (3.3,-0.16);
  \node at (0,1.5) {\( (s_1^{(0)},v_1^{(0)}) \)};
  \node at (0,0.5) {\( (s_2^{(0)},v_2^{(0)}) \)};
  \node at (0,-0.5) {\( (s_3^{(0)},v_3^{(0)}) \)};
  \node at (0,-1.5) {\( (s_4^{(0)},v_4^{(0)}) \)};
  \node at (2,0.95) {\( (s_1^{(1)},v_1^{(1)}) \)};
  \node at (2,-0.95) {\( (s_2^{(1)},v_2^{(1)}) \)};
  \node at (4.4,0) {\( (s_1^{(2)},v_1^{(2)}=:\widehat\theta_n) \)};
\end{tikzpicture}
\end{minipage}%
\caption{Left: comparison weight \(w(x)=G((x-y+\tau_{\mathrm{sel}})/(2\tau_{\mathrm{sel}}))\) for arbitrary \(y\) and \(\tau_{\mathrm{sel}}>0\). If the loss score differs by more than tau, only the smaller parameter is retained. Otherwise, a convex combination is taken. Right: a four-"leaf" example of the pairwise reduction, i.e., $d_{\mathrm{sel}}=2$. Each "node" carries the blended score \(s\) and candidate parameter \(v\), and the root returns our initializer.}
\label{fig:selection-weight}
\end{figure}
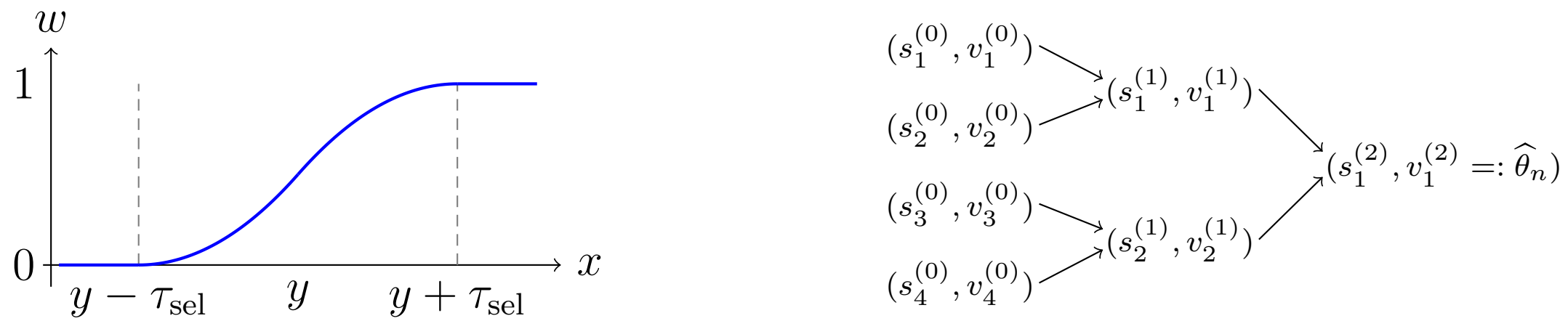

Set
\begin{equation}
 K:=2\|\nabla_\theta x\|_{L^\infty(\Theta\times [0,T_{\mathrm{obs}}])}^{2}
+4\|x\|_{L^\infty(\Theta\times [0,T_{\mathrm{obs}}])}\|\nabla_\theta^2x\|_{L^\infty(\Theta\times [0,T_{\mathrm{obs}}])}\geq\sup_{\vartheta\in\Theta_0,\theta\in\Theta}
 \norm{\nabla_\theta^2\mathcal{R}_\vartheta(\theta)}_{\mathrm{op}}.
 \label{eq:global-curvature-and-inner-radius}
\end{equation}
Choose a fixed \(h\)-net \(\mathcal{N}_h=\{\theta_1,\ldots,\theta_{ {N}}\}\subset\Theta\), where
\begin{equation}
 h:=\min\{\rho/4,\sqrt{\gamma_\rho/(2K)}\}.
 \label{eq:selection-gap}
\end{equation}
Given computed approximate losses \(\widetilde{\mathcal{R}}_{\theta^*}^{(n)}(\theta_1),\ldots,\widetilde{\mathcal{R}}_{\theta^*}^{(n)}(\theta_N)\), define the initializer
as follows. For \( {N}=1\), set \(\widehat\theta_n=\theta_1\).
For \( {N}\ge2\), let
\[
 d_{\mathrm{sel}}:=\lceil\log_2 {N}\rceil,\qquad
 \tau_{\mathrm{sel}}:=\frac{\gamma_\rho}{16d_{\mathrm{sel}}},\qquad
 \sigma_2(z):=(z_+)^2,
\]
and set
\[
 G(z):=2\sigma_2(z)-4\sigma_2(z-1/2)+2\sigma_2(z-1).
\]
Initialize \((s_j^{(0)},v_j^{(0)})=(\widetilde{\mathcal{R}}_{\theta^*}^{(n)}(\theta_j),\theta_j)\) for
\(j=1,\ldots, {N}\), repeating the last pair until there are \(2^{d_{\mathrm{sel}}}\) pairs.
For \(k=1,\ldots,d_{\mathrm{sel}}\), replace adjacent pairs in parallel by computing,
for \(j=1,\ldots,2^{d_{\mathrm{sel}}-k}\),
\begin{align}
 w_j^{(k)}
 &=G\!\left(\frac{s_{2j}^{(k-1)}-s_{2j-1}^{(k-1)}
                  +\tau_{\mathrm{sel}}}{2\tau_{\mathrm{sel}}}\right),\notag\\
 s_j^{(k)}
 &=w_j^{(k)}s_{2j-1}^{(k-1)}
   +(1-w_j^{(k)})s_{2j}^{(k-1)},\notag\\
 v_j^{(k)}
 &=w_j^{(k)}v_{2j-1}^{(k-1)}
   +(1-w_j^{(k)})v_{2j}^{(k-1)}.
 \label{eq:selection-ffn-stages}
\end{align}
Thus after $d_{\mathrm{sel}}$ stages of reduction to a final pair, return
\begin{equation}
 \widehat\theta_n:=v_1^{(d_{\mathrm{sel}})}.
 \label{eq:selection-initializer}
\end{equation}
Figure~\ref{fig:selection-weight} illustrates the comparison weight and the resulting reduction tree.

\begin{lemma}[Squared-ReLU initialization and localization]
\label{lem:selection-localization}
The map from $(\widetilde{\mathcal{R}}_{\theta^*}^{(n)}(\theta_1),\ldots,\allowbreak\widetilde{\mathcal{R}}_{\theta^*}^{(n)}(\theta_N))$ to $\widehat\theta_n$ given by \eqref{eq:selection-initializer} is represented exactly by a
feedforward network with squared-ReLU hidden activations and an affine
output layer, of depth \(O(\log(2 {N}))\) and width \(O( {N}(p+1))\).
Additionally if
\begin{equation}
 \max_j|\widetilde{\mathcal{R}}_{\theta^*}^{(n)}(\theta_j)-\mathcal{R}_{\theta^*}(\theta_j)|\leq\gamma_\rho/4,
 \label{eq:selection-risk-event}
\end{equation}
then \(\widehat\theta_n\in B(\theta^*,\rho/4)\). In particular, on the event of Lemma \ref{lem:uniform-risk-concentration}, 
\[C_RB_0\sqrt{\frac{p\log(2\kappa)+\log(1/\delta)}{n}}
 +3\sigma^2\sqrt{\frac{m\log(2/\delta)}{n}}
+ \|\mathfrak F-x\|_\infty^2
+4\|x\|_\infty\|\mathfrak F-x\|_\infty \le\gamma_\rho/4\]
 suffices for initialization into the strongly convex neighbourhood.
\end{lemma}

\begin{proof}[Proof of Lemma~\ref{lem:selection-localization}]
We first study the procedure defined in \eqref{eq:selection-ffn-stages}. For any input pairs \((a,u)\) and \((b,v)\) under comparison, the comparison returns
\[
 w=G\!\left(\frac{b-a+\tau}{2\tau}\right),\qquad
 s=wa+(1-w)b,\qquad z=wu+(1-w)v.
\]
Recall
\[
G(t) = \begin{cases}
0 & \text{if } t < 0 \\
2t^2 & \text{if } t \in [0, 1/2], \\
1 - 2(1-t)^2 & \text{if } t \in [1/2, 1],
\\ 1 &\text{if } t > 1.
\end{cases}
\] 
Hence \(0\le w\le1\). If \(b-a\ge\tau\), then
\(w=1\), and if \(a-b\ge\tau\), then \(w=0\). Otherwise
\(|a-b|<\tau\) and \(s\) lies between \(a\) and \(b\).
These cases imply
\[
 s\le\min(a,b)+\tau,\qquad
 w>0\ \Longrightarrow\ a\le s+\tau,\qquad
 1-w>0\ \Longrightarrow\ b\le s+\tau.
\]
\paragraph{Error bound.} For any node $(s^{(k)},v^{(k)})$ at level $k$, define the set of \emph{descendant leaves}, i.e., the set of nodes at level $0$ that descend into $(s,v)$, as $I(s^{(k)},v^{(k)})$. Then we prove the following inductive hypothesis: for any $0\leq k\leq d_{\mathrm{sel}}$, every node $(s^{(k)},v^{(k)})$ at level $k$ satisfies
\[
 s^{(k)}\le\min_{j\in I(s^{(k)},v^{(k)})}\widetilde{\mathcal{R}}_{\theta^*}^{(n)}(\theta_j)+k\tau,
\]
and there are coefficients \(c_j\ge0\), \(\sum_{j\in I}c_j=1\), with
\[
 v=\sum_{j\in I}c_j\theta_j,\qquad
 s^{(k)}=\sum_{j\in I}c_j\widetilde{\mathcal{R}}_{\theta^*}^{(n)}(\theta_j),\qquad
 c_j>0\ \Longrightarrow\ \widetilde{\mathcal{R}}_{\theta^*}^{(n)}(\theta_j)\le s^{(k)}+k\tau.
\]
All assertions hold for nodes at \(k=0\) trivially. Suppose they hold for nodes at level \(k-1\). Take a level \(k\) node $(s^{(k)},v^{(k)})$ and consider its two descendents at level \(k-1\).  Denote them as $(a,x)$ and $(b,y)$. The one-comparison
bounds give
\begin{equation}
\begin{split}
 s^{(k)}&\le\min(a,b)+\tau
 \\
 &\le\min\{\min_{j\in I(a,x)}\widetilde{\mathcal{R}}_{\theta^*}^{(n)}(\theta_j)+(k-1)\tau,\min_{j\in I(b,y)}\widetilde{\mathcal{R}}_{\theta^*}^{(n)}(\theta_j)+(k-1)\tau\}+\tau
 \\
 &=\min_{j\in I(s^{(k)},v^{(k)})}\widetilde{\mathcal{R}}_{\theta^*}^{(n)}(\theta_j)+k\tau,
\end{split}
\label{eq:ind-one}
\end{equation}
since \(I(s^{(k)},v^{(k)})=I(a,x)\cup I(b,y)\).
The subsequent convex combination update step from level \(k-1\) to level \(k\) also preserves the convex combination property assumed for nodes at level \(k-1\). 
Any resulting positive coefficient $c_j$ is necessarily $w'p_j$ for some $w'>0$ from the update step and positive coefficient $p_j$ of one of the descendants. WLOG, assume that the first descendant has the positive coefficient $p_j$. Thus, using the inductive assumption at level $k-1$, we have
\begin{equation}
     \widetilde{\mathcal{R}}_{\theta^*}^{(n)}(\theta_j)\le a+(k-1)\tau\le s^{(k)}+k\tau,
     \label{eq:ind-two}
\end{equation}

completing the induction.\\
At level $d_{\mathrm{sel}}$, combining \eqref{eq:ind-one} and \eqref{eq:ind-two}, the two bounds yields for its coefficients \(\{c_j\}_{j=1}^{2^{d_{\mathrm{sel}}}}\)
\begin{equation}
 c_j>0\ \Longrightarrow\
 \widetilde{\mathcal{R}}_{\theta^*}^{(n)}(\theta_j)\le\min_i\widetilde{\mathcal{R}}_{\theta^*}^{(n)}(\theta_i)+2d_{\mathrm{sel}}\tau\leq \min_i \widetilde{\mathcal{R}}_{\theta^*}^{(n)}(\theta_i) +\gamma_\rho/8.
 \label{eq:ind-three}
\end{equation}
After adding the coefficients of repeated leaves, we therefore have
\[
 \widehat\theta_n=\sum_{j=1}^N c_j\theta_j,\qquad
 c_j\ge0,\qquad \sum_{j=1}^N c_j=1.
\]
For every \(j\) with \(c_j>0\), it follows then
\begin{align*}
 \mathcal{R}_{\theta^*}(\theta_j)
 &\le\widetilde{\mathcal{R}}_{\theta^*}^{(n)}(\theta_j)+\gamma_\rho/4\quad\text{by \eqref{eq:selection-risk-event}}\\
 &\le\min_i\widetilde{\mathcal{R}}_{\theta^*}^{(n)}(\theta_i)+\gamma_\rho/8+\gamma_\rho/4\quad\text{by \eqref{eq:ind-three}}\\
 &\le\widetilde{\mathcal{R}}_{\theta^*}^{(n)}(\theta_{j^*})+\gamma_\rho/8+\gamma_\rho/4\quad\text{for some }\theta_{j^*}\in B(\theta^*,h)\\
 &\le \mathcal{R}_{\theta^*}(\theta_{j^*})+\gamma_\rho/4+\gamma_\rho/8+\gamma_\rho/4\quad\text{by \eqref{eq:selection-risk-event}}\\
 &\le \mathcal{R}_{\theta^*}(\theta^*)+Kh^2/2+\gamma_\rho/4+\gamma_\rho/8+\gamma_\rho/4\quad\text{by Taylor's formula}\\
 &\le \mathcal{R}_{\theta^*}(\theta^*)+7\gamma_\rho/8,
\end{align*}
By definition of $\gamma_\rho$, this implies
\(\|\theta_j-\theta^*\|_2<\rho/4\). Consequently,
\[
 \|\widehat\theta_n-\theta^*\|_2
 \le\sum_{j=1}^N c_j\|\theta_j-\theta^*\|_2<\rho/4.
\]
The output also lies in \(\Theta\) by convexity. 

\paragraph{Exact network representation.}
The input loss score vector
\((\widetilde{\mathcal{R}}_{\theta^*}^{(n)}(\theta_1),\ldots,
\widetilde{\mathcal{R}}_{\theta^*}^{(n)}(\theta_N))\in\mathbb{R}^N\) is mapped via 
an affine map containing the fixed representatives $\{\theta_1,\ldots,\theta_N\}$ as weights to
\[
 \big(s_1^{(0)},v_1^{(0)},\ldots,
 s_{2^{d_{\mathrm{sel}}}}^{(0)},v_{2^{d_{\mathrm{sel}}}}^{(0)}\big)
 \in\mathbb{R}^{2^{d_{\mathrm{sel}}}(p+1)}\]
At reduction stage \(k\),
the network processes adjacent pairs of entries in parallel: for each
\(j=1,\ldots,2^{d_{\mathrm{sel}}-k}\), it maps
\(\big(s_{2j-1}^{(k-1)},v_{2j-1}^{(k-1)},
s_{2j}^{(k-1)},v_{2j}^{(k-1)}\big)\) to
\(\big(s_j^{(k)},v_j^{(k)}\big)\) by the formulas in
\eqref{eq:selection-ffn-stages}. Explicitly, for one such comparison,
write the input as \((a,u,b,v)\), where \(a,b\in\mathbb R\) are scores
and \(u,v\in\mathbb R^p\) are candidate vectors. An affine preactivation
forms \(q=(b-a+\tau_{\mathrm{sel}})/(2\tau_{\mathrm{sel}})\), and three hidden
units compute \(\sigma_2(q)\), \(\sigma_2(q-1/2)\), and
\(\sigma_2(q-1)\). Their affine combination
\(2\sigma_2(q)-4\sigma_2(q-1/2)+2\sigma_2(q-1)\) is exactly the gate
\(G(q)\). Identity channels from Lemma \ref{lem:network-with-identity-channels} preserve the entries of
\((a,u,b,v)\) to get $(a,u,b,v,G(q))$. Appropriate scalar multiplications between the weights and the coordinates via \[ab=\sigma_2(a+b)/4+\sigma_2(-a-b)/4-\sigma_2(b-a)/4-\sigma_2(a-b)/4\] give the final update exactly. Thus one comparison is a fixed-depth
squared-ReLU network, and running these sub-networks in parallel implements
one level update. Consequently, after \(k\) updates, the
intermediate vector is
\[
 \big(s_1^{(k)},v_1^{(k)},\ldots,
 s_{2^{d_{\mathrm{sel}}-k}}^{(k)},v_{2^{d_{\mathrm{sel}}-k}}^{(k)}\big)
 \in\mathbb{R}^{2^{d_{\mathrm{sel}}-k}(p+1)}.
\]
At the final stage, \(z^{(d_{\mathrm{sel}})}=(s_1^{(d_{\mathrm{sel}})},
v_1^{(d_{\mathrm{sel}})})\in\mathbb{R}^{p+1}\), and an affine output map
projects onto its last \(p\) coordinates to return
\(\widehat\theta_n=v_1^{(d_{\mathrm{sel}})}\in\mathbb{R}^p\).
The displayed square
and multiplication identities show that each comparison requires
constant depth and \(O(p+1)\) width and nonzero parameters.
Comparisons within each level run in parallel. Since
\(N\le2^{d_{\mathrm{sel}}}<2N\), the \(d_{\mathrm{sel}}\) levels give depth \(O(\log(2N))\)
and width \(O(N(p+1))\). 
\end{proof}

Now we turn to the analysis of the approximate gradient descent algorithm.
Let $\eta_\star:=\frac{2}{K+\mu}$, and
\begin{equation}
 q_n:=\frac{K-\mu+2\eta_n}{K+\mu}.
 \label{eq:gradient-contraction-factor}
\end{equation}
Let \(T_n(\theta)=\theta-\eta_\star\nabla_\theta \mathcal R_{\theta^*}^{(n)}(\theta)\).
Suppose the implemented update
\(\widetilde T_{n}\) satisfies
\begin{equation}
 \sup_{\theta\in B(\theta^*,\rho)}
 \norm{\widetilde T_{n}(\theta)-T_n(\theta)}_2
 \le\varepsilon_{\mathrm{step}},
 \label{eq:approximate-gradient-step}
\end{equation}
for some $\varepsilon_{\mathrm{step}} \ge 0$, and set
\(\widehat\theta_n^0=\widehat\theta_n\),
\(\widehat\theta_n^{k+1}=
\widetilde T_{n}(\widehat\theta_n^k)\).

\begin{lemma}[Approximate gradient-descent error]
\label{lem:approximate-gradient-descent}
Assume the same event and conditions as Corollary~\ref{cor:empirical-localization-curvature}
and Lemma~\ref{lem:selection-localization}, suppose
\[
 \varepsilon_{\mathrm{step}}
 \leq(1-q_n)\frac{3\rho}{4}.
\]
Then every iterate remains in \(B(\theta^*,\rho)\) and, for every
\(k\ge0\),
\begin{equation}
 \norm{\widehat\theta_n^k-\theta_n}_2
 \le q_n^k\norm{\widehat\theta_n-\theta_n}_2
 +\varepsilon_{\mathrm{step}}\frac{1-q_n^k}{1-q_n}.
 \label{eq:approximate-step-geometric-rate}
\end{equation}
In particular, the transient term decays geometrically and the bound
converges to \(\varepsilon_{\mathrm{step}}/(1-q_n)\).
\end{lemma}

\begin{proof}[Proof of Lemma~\ref{lem:approximate-gradient-descent}]
First note that $q_n<1$ given the conditions and event of Corollary~\ref{cor:empirical-localization-curvature}.
On \(B(\theta^*,\rho)\), the ideal-risk Hessian has eigenvalues in
\([\mu,K]\). By the definition of \(\eta_n\) and Weyl's inequality,
\[
 (\mu-\eta_n)I_p\preceq\nabla_\theta^2\mathcal R_{\theta^*}^{(n)}(\theta)
 \preceq(K+\eta_n)I_p.
\]
For \(\theta,\theta'\) in this ball
\[
 T_n(\theta)-T_n(\theta')
 =\left[I_p-\eta_\star
 \int_0^1\nabla_\theta^2\mathcal R_{\theta^*}^{(n)}(\theta'+s(\theta-\theta'))\,ds\right]
 (\theta-\theta').
\]
\[
 \Rightarrow \norm{T_n(\theta)-T_n(\theta')}_2\le q_n\norm{\theta-\theta'}_2.
\]
Hence \(T_n\) is \(q_n\)-Lipschitz on the ball. Since
\(\nabla_\theta \mathcal R_{\theta^*}^{(n)}(\theta_n)=0\), \(T_n(\theta_n)=\theta_n\), and
\eqref{eq:approximate-gradient-step} yields, whenever
\(\widehat\theta_n^k\in B(\theta^*,\rho)\),
\[
 \norm{\widehat\theta_n^{k+1}-\theta_n}_2
 \le q_n\norm{\widehat\theta_n^k-\theta_n}_2
 +\varepsilon_{\mathrm{step}}.
\]
Write
\[
 e_k:=\norm{\widehat\theta_n^k-\theta_n}_2,
 \qquad
 U_k:=q_n^ke_0+\varepsilon_{\mathrm{step}}
              \frac{1-q_n^k}{1-q_n}.
\]
where $e_0=\norm{\widehat\theta_n-\theta_n}_2$ and $U_k$ is the right hand side of \eqref{eq:approximate-step-geometric-rate}. On the event in Lemma~\ref{lem:selection-localization} and
Corollary~\ref{cor:empirical-localization-curvature}, $e_0\le \rho/2$ and hence \(U_k\le 3\rho/4\) for every \(k\ge0\) due to the additional $\varepsilon_{\mathrm{step}}$ bound.

We now prove simultaneously, by induction, that
\(\widehat\theta_n^k\in B(\theta^*,\rho)\) and \(e_k\le U_k\).
For \(k=0\), Lemma~\ref{lem:selection-localization} gives
\[
 \widehat\theta_n^0=\widehat\theta_n
 \in B(\theta^*,\rho/4)\subset B(\theta^*,\rho),
\]
and \(e_0=U_0\). Now suppose that
\(\widehat\theta_n^k\in B(\theta^*,\rho)\) and \(e_k\le U_k\).
Because the current iterate lies in the strongly convex neighborhood,
the one-step estimate from above applies and yields
\[
 e_{k+1}\le q_ne_k+\varepsilon_{\mathrm{step}}
 \le q_nU_k+\varepsilon_{\mathrm{step}}=U_{k+1}\leq\frac{3\rho}{4}.
\]
The triangle inequality and the statistical localization of
\(\theta_n\) therefore give
\[
 \begin{aligned}
 \norm{\widehat\theta_n^{k+1}-\theta^*}_2
 &\le
 \norm{\widehat\theta_n^{k+1}-\theta_n}_2
 +\norm{\theta_n-\theta^*}_2\\
 &=e_{k+1}+\norm{\theta_n-\theta^*}_2\\
 &<\frac{3\rho}{4}+\frac{\rho}{4}=\rho.
 \end{aligned}
\]
Thus \(\widehat\theta_n^{k+1}\in B(\theta^*,\rho)\), and we have already
shown \(e_{k+1}\le U_{k+1}\). Both induction claims therefore hold at
iteration \(k+1\). This proves
\eqref{eq:approximate-step-geometric-rate} for every \(k\), while also
justifying at each step the local estimate used to obtain it. Setting
\(\varepsilon_{\mathrm{step}}=0\) gives the exact-gradient bound.
\end{proof}

Now we simply instantiate a specific instance of the minimization procedure described above.
Define \[
 A( {\theta,t}):=2\nabla_\theta x( {\theta,t}),\qquad
 b( {\theta,t}):=A( {\theta,t})x( {\theta,t}),\qquad
 B_Y:=\norm{x}_{L^\infty(\Theta\times [0,T_{\mathrm{obs}}])}+\sigma(\sqrt m+1).
\]
Recall the \(h\)-net $\mathcal{N}_h$ of \(\Theta\) with cardinality \( {N}\) satisfying
\[
  {N}\le\left(1+\frac{D}{h}\right)^p
 =\left(1+\max\left\{\frac{4D}{\rho},
                  D\sqrt{\frac{2K}{\gamma_\rho}}\right\}\right)^p,
\]
in the worst-case where $\Theta$ is a $p$-dimensional ball of radius $D/2$. Note that if $\Theta$ is degenerate in some way, the bound on $  {N}$ could be much smaller. 
\\Now set
\[
 C_\alpha:=C_0\left(1+\log\frac{1+D\tau}{2\alpha}\right),\qquad
\]
and for any fixed numerical approximation \(\mathfrak F\), evaluate \(B_0,B_1,\kappa\) from
\eqref{eq:B0}--\eqref{eq:B1}, and set
\[
 a_n(\alpha):=\frac{C_RB_0\sqrt{p\log(2\kappa)+\log(1/\alpha)}
 +3\sigma^2\sqrt{m\log(2/\alpha)}}{\sqrt n},\qquad
\]
and let
\[
 J(\alpha,\mathfrak F):=
 \left(\frac{8C_RB_0\sqrt{p\log(2\kappa)+\log(1/\alpha)}
 +24\sigma^2\sqrt{m\log(2/\alpha)}}{\gamma_\rho}\right)^2
\]
for an arbitrary fixed $\mathfrak F$.
Define
\begin{equation}
\begin{aligned}
    N(\alpha)
    :=\max\Bigg\{\,&
      \frac{4\log(2/\alpha)}{m}, C_\alpha p\log p, 2\log\bigg(\frac{1}{\alpha}\bigg),\\
      &\frac{8\tau^4C_\alpha p}{\mu^2}
       \log\!\left(e+\frac{8\tau^4C_\alpha p}{\mu^2}\right),\\
      &\frac{8C_{SS}(2p+1)}{\mu(\rho')^2}
       \log\!\left(e+\frac{8C_{SS}(2p+1)}{\mu(\rho')^2}\right),\\
      &\frac{4C_{SS}}{\mu(\rho')^2}
       \left[(2p+1)\log\sqrt m+\log\frac1\alpha\right]
      \Bigg\},
\end{aligned}
 \label{eq:minimization-statistical-conditions}
\end{equation}
so that these thresholds on sample sizes ensure the relevant events and inequalities derived throughout the appendix.
\begin{theorem}[Parameter recovery guarantee]
\label{thm:statistical-recovery-theorem}
Assume that \(\mathfrak F\) is a numerical approximation of \(x\) satisfying
\begin{equation}
\begin{split}
 \varepsilon_{\mathrm{sol}}^{(\mathrm{obs})}&:=
 \|\mathfrak{F}(\theta,t)-x(\theta,t)\|_{L^\infty(\Theta\times [0,T_{\mathrm{obs}}])}\\
 &\le 2\left(\sqrt{\|x\|^2_{L^\infty(\Theta\times [0,T_{\mathrm{obs}}])}+\gamma_\rho/32}-\|x\|_{L^\infty(\Theta\times [0,T_{\mathrm{obs}}])}\right) \text{ and } L_\mathfrak{F}< \infty,
\end{split}
\end{equation}
and that numerical approximations $\widehat A$ and $\widehat b$ of $A$ and $b$ respectively satisfies
\begin{equation}
 \varepsilon_{\mathrm{grad}}:= {\sup_{(\theta,t)\in\Theta\times[0,T_{\mathrm{obs}}]}}
 \sqrt{\norm{\widehat A(\theta,t)-A(\theta,t)}_F^2
       +\norm{\widehat b(\theta,t)-b(\theta,t)}_2^2}\leq\frac{3\rho\mu}{8(1+B_Y)}.
 \label{eq:approximation-budgets}
\end{equation}
Then for the approximate gradient descent update
\begin{equation}
  \tilde{T}_{n}(\theta)=\theta-\frac{\eta_\star}{n}\sum_i[\widehat b( {\theta,t_i})-\widehat A( {\theta,t_i})Y_i] 
\end{equation}
and the finite net $\mathcal{N}_h$ defined above, we have the following error guarantee:\\
For all $\theta^*\in\Theta_0$, \(\alpha\in(0,1/7)\), \(n>\max\{N(\alpha),J(\alpha,\mathfrak F)\}\), and every $k>0$, the iterate \(\widehat\theta^k\) of approximate GD using $\tilde{T}_n$ after coarse initialization on $\mathcal{N}_h$ satisfies
\begin{equation} 
\norm{\widehat\theta^k-\theta^*}_2
 \le\sqrt{\frac{2C_{SS}}\mu\varepsilon_n^{(SS)}(\alpha)}
 +\bigg(\frac{K}{K+\mu}\bigg)^k\frac{\rho}{2}
 +\frac{2(1+B_Y)}\mu\varepsilon_{\mathrm{grad}}
\label{eq:parameter-recovery-error}
\end{equation}
with probability at least \(1-7\alpha\). 
\end{theorem}

\begin{proof}[Proof of Theorem~\ref{thm:statistical-recovery-theorem}]
Fix \(\theta^*\in\Theta_0\), \(\alpha\in(0,1/7)\), and
\(n>\max\{N(\alpha),J(\alpha,\mathfrak F)\}\).

\paragraph{Consequences of the sample-size conditions.}
First, \(n>2\log(1/\alpha)>e\). For \(a>0\), the elementary
implication
\[
 n>e,\quad n>2a\log(e+2a)\quad\Longrightarrow\quad n>a\log n
\]
inverts the logarithmic conditions in \( \mathfrak N(\alpha)\).
Applying this with \(a=4\tau^4C_\alpha p/\mu^2\) gives
\[
 H_n:=\tau^2\sqrt{\frac{C_\alpha p\log n}{n}}<\frac\mu2.
\]
The same implication with \(a=4C_{SS}(2p+1)/[\mu(\rho')^2]\), and simply inverting the last term in \(\mathfrak N(\alpha)\), gives respectively
\[
 \frac{C_{SS}(2p+1)\log n}{n}<\frac{\mu(\rho')^2}{4},
 \qquad
 \frac{C_{SS}[(2p+1)\log\sqrt m+\log(1/\alpha)]}{n}<\frac{\mu(\rho')^2}{4}.
\]
Consequently,
\[
 C_{SS}\varepsilon_n^{(SS)}(\alpha)<\frac\mu2(\rho')^2 
 \Rightarrow \sqrt{\frac{2C_{SS}}\mu\varepsilon_n^{(SS)}(\alpha)}
 <\rho'\le\rho/4.
\]
Finally, \(n>J(\alpha,\mathfrak F)\) gives
\(a_n(\alpha)=\frac{C_RB_0\sqrt{p\log(2\kappa)+\log(1/\alpha)}
 +3\sigma^2\sqrt{m\log(2/\alpha)}}{\sqrt n}\le\gamma_\rho/8\). Also note that $\mathfrak{F}$ is fixed by choosing $\varepsilon_{\mathrm{sol}}^{(\mathrm{obs})}$ that satisfies the condition in the theorem (could be chosen to be the boundary value itself), and hence $J(\alpha,\mathfrak F)$ is a fixed quantity for the fixed ODE family.

\paragraph{The common probability event.}
Let \(\theta_n\in \argmin \mathcal R_{\theta^*}^{(n)}\), and define
\begin{align*}
 \mathcal E_R&:=\left\{
 \sup_{\theta\in\Theta}|\widetilde{\mathcal{R}}_{\theta^*}^{(n)}(\theta)-\widetilde{\mathcal{R}}_{\theta^*}(\theta)|
 \le a_n(\alpha)\right\},\\
 \mathcal E_H&:=\left\{
 \eta_n:=\sup_{\theta\in\Theta}
 \norm{\nabla_\theta^2\mathcal R_{\theta^*}^{(n)}(\theta)-\nabla_\theta^2\mathcal{R}_{\theta^*}(\theta)}_{\mathrm{op}}
 \le H_n\right\},\\
 \mathcal E_{SS}&:=\left\{
 \mathcal{R}_{\theta^*}(\theta_n)-\mathcal{R}_{\theta^*}(\theta^*)\le C_{SS}\varepsilon_n^{(SS)}(\alpha)
 \right\},\\
 \mathcal E_M&:=\left\{\frac1n\sum_{i=1}^n\norm{Y_i}_2\le B_Y\right\}.
\end{align*}
The conditions \(n>4\log(2/\alpha)/m\) and
\(n>C_\alpha p\log p\) allow us to apply
Lemmas~\ref{lem:uniform-risk-concentration} and
\ref{lem:uniform-derivative-concentration}. Together with
Lemma~\ref{lem:tighter-population-risk}, they bound each of
\(\mathbb P(\mathcal E_R^c),\mathbb P(\mathcal E_H^c),
\mathbb P(\mathcal E_{SS}^c)\) by \(2\alpha\).

To control \(\mathcal E_M\), the function
\((\xi_1,\ldots,\xi_n)\mapsto n^{-1}\sum_i\norm{\xi_i}_2\)
is \(n^{-1/2}\)-Lipschitz, and
\(\E\norm{\xi_i}_2\le\sqrt m\). Gaussian concentration yields
\[
 \mathbb P\left\{\frac1n\sum_i\norm{\xi_i}_2>
 \sqrt m+\sqrt{\frac{2\log(1/\alpha)}n}\right\}\le\alpha.
\]
Since \(n>2\log(1/\alpha)\) and
\(\norm{Y_i}_2\le \|x\|_{L^\infty(\Theta\times [0,T_{\mathrm{obs}}])}+\sigma\norm{\xi_i}_2\), this gives
\(\mathbb P(\mathcal E_M^c)\le\alpha\). Thus the single event
\[
 \mathcal E:=\mathcal E_R\cap\mathcal E_H\cap\mathcal E_{SS}
             \cap\mathcal E_M
\]
has probability at least \(1-7\alpha\). We work on this event below.
The localization argument of
Corollary~\ref{cor:empirical-localization-curvature} gives
\(\norm{\theta_n-\theta^*}_2\le r_n<\rho/4\) and positive
empirical curvature on \(B(\theta^*,\rho)\).

\paragraph{Initialization and update errors.}
The trajectory-error assumption is equivalent to
\((\varepsilon_{\mathrm{sol}}^{(\mathrm{obs})})^2+4\|x\|_{L^\infty(\Theta\times [0,T_{\mathrm{obs}}])}\varepsilon_{\mathrm{sol}}^{(\mathrm{obs})}\le\gamma_\rho/8\).
Lemma~\ref{lem:uniform-risk-concentration}, \(\mathcal E_R\), and the relevant sample-size condition therefore gives
\[
  \sup_\theta|\widetilde{\mathcal{R}}_{\theta^*}^{(n)}(\theta)-\mathcal{R}_{\theta^*}(\theta)|
 \le\gamma_\rho/4.
\]
Lemma~\ref{lem:selection-localization} applies using $\ell_j=\widetilde{\mathcal{R}}_{\theta^*}^{(n)}(\theta_j)$ on the net $\mathcal{N}_h$ so the initializer \(\widehat\theta^0\) satisfies
\[
 \norm{\widehat\theta^0-\theta^*}_2<\rho/4,
 \qquad e_0:=\norm{\widehat\theta^0-\theta_n}_2
 <\rho/4+r_n<\rho/2.
\]
By the definitions of \(A,b\), the exact gradient map is
\[
 T_n(\theta)=\theta-\frac{\eta_\star}{n}
 \sum_i[b( {\theta,t_i})-A( {\theta,t_i})Y_i].
\]
The coefficient-error assumption \eqref{eq:approximation-budgets} and \(\mathcal E_M\) imply
\begin{align*}
 \sup_{\theta\in\Theta}\norm{\widetilde T_n(\theta)-T_n(\theta)}_2
 &\le\eta_\star\varepsilon_{\mathrm{grad}}
       \left(1+\frac1n\sum_i\norm{Y_i}_2\right)\\
 &\le\eta_\star(1+B_Y)\varepsilon_{\mathrm{grad}}
 =:\varepsilon_{\mathrm{step}}.
\end{align*}
Thus 
\[
\varepsilon_{\mathrm{step}}\le\bigg(1-\frac{K}{K+\mu}\bigg)\frac{3\rho}{4}
\]
Combining this with
\[
 q_n=\frac{K-\mu+2\eta_n}{K+\mu}
 <\frac K{K+\mu}<1 \quad \text{on }\mathcal E_H,
\]
gives $\varepsilon_{\mathrm{step}}\le(1-q_n)\frac{3\rho}{4}$ on the same event, and so Lemma~\ref{lem:approximate-gradient-descent} ensures that all
iterates remain in \(B(\theta^*,\rho)\), and simultaneously for
every integer \(k\ge0\),
\begin{align*}
 \norm{\widehat\theta^k-\theta^*}_2
 &\le \norm{\theta_n-\theta^*}_2+\norm{\widehat\theta^k-\theta_n}_2\leq \norm{\theta_n-\theta^*}_2+q_n^k e_0+
             \varepsilon_{\mathrm{step}}\frac{1-q_n^k}{1-q_n}\\
 &\le \norm{\theta_n-\theta^*}_2+\bigg(\frac{K}{K+\mu}\bigg)^k\frac\rho2+
             \frac{\eta_\star(1+B_Y)\varepsilon_{\mathrm{grad}}}{1-\bigg(\frac{K}{K+\mu}\bigg)}\\
 &=\sqrt{\frac{2C_{SS}}\mu\varepsilon_n^{(SS)}(\alpha)}
   +\left(\frac K{K+\mu}\right)^k\frac\rho2
   +\frac{2(1+B_Y)}\mu\varepsilon_{\mathrm{grad}}.
\end{align*}
This proves \eqref{eq:parameter-recovery-error} on \(\mathcal E\).
\end{proof}

\section{Transformer class approximation results}
\label{app:transformer-approximation}
\subsection{Proof of Theorem~\ref{thm:inverse-transformer-realization}}
\label{subsec:proof-realization-parameter}
\begin{proof}[Proof of Theorem~\ref{thm:inverse-transformer-realization}]

We describe the assembly underlying
Theorem~\ref{thm:inverse-transformer-realization}. First note that all feedforward calculations act on each column
with the same weights, and the selective mean module is performed only by
Lemma~\ref{lem:selective-mean-attention}. All scalar arithmetic below
is exact by Lemma~\ref{lem:squared-relu-arithmetic}, and all coordinates
that must be retained are carried by
Lemma~\ref{lem:network-with-identity-channels}. In feedforward-only layers (i.e., FFNs with more than two layers), the attention
value maps are implicitly take to be zero by setting the query, key, and value matrices to be zero. 
\paragraph{Existence of $\mathrm{NN}^{\varepsilon_{\mathrm{sol}}}_{\mathrm{sol},T}$ and $\mathrm{NN}^{\varepsilon_{\mathrm{grad}}}_{\mathrm{grad}}$.} Firstly, note that by standard Universal approximation results \cite{yarotsky2017,li2020powernet}, the existence of $\mathrm{NN}^{\varepsilon_{\mathrm{sol}}}_{\mathrm{sol},T}$ and $\mathrm{NN}^{\varepsilon_{\mathrm{grad}}}_{\mathrm{grad}}$ and their respective size-scalings $L_{(\cdot)}(\varepsilon_{(\cdot)}),W_{(\cdot)}(\varepsilon_{(\cdot)}),P_{(\cdot)}(\varepsilon_{(\cdot)})$ is guaranteed. The construction below is then a direct implementation of the approximate gradient descent procedure in Theorem~\ref{thm:statistical-recovery-theorem} using these networks as sub-networks.
\paragraph{Sample-size threshold.} Set $\mathfrak{N}(\alpha) := \max\{N(\alpha),J(\alpha,\mathrm{NN}^{\varepsilon_{\mathrm{sol}}^{(\mathrm{obs})}}_{\mathrm{sol},T_{\mathrm{obs}}})\}$, where this now only depends on $\alpha$ and the ODE family once $\mathrm{NN}^{\varepsilon_{\mathrm{sol}}^{(\mathrm{obs})}}_{\mathrm{sol},T_{\mathrm{obs}}}$ is fixed to have error bounded by $\varepsilon_{\mathrm{sol}}^{(\mathrm{obs})}$ as in Theorem~\ref{thm:statistical-recovery-theorem}. 
\paragraph{Trajectory evaluation and candidate losses.}
Fix $\mathfrak{F}(\theta,t):=\mathrm{NN}^{\varepsilon_{\mathrm{sol}}^{(\mathrm{obs})}}_{\mathrm{sol},T_{\mathrm{obs}}}(\theta,t)$. Given finite depth and width, $L_\mathfrak{F}<\infty$ so that Theorem~\ref{thm:statistical-recovery-theorem} eventually applies.
Restrict \(\mathrm{NN}^{\varepsilon_{\mathrm{sol}}^{(\mathrm{obs})}}_{\mathrm{sol},T_{\mathrm{obs}}}\) to each fixed \(\theta_j\) using
Lemma~\ref{lem:restricted-network}, and stack the resulting maps in
parallel, tying their corresponding weights. Preserve one copy of
\((t_i,Y_i)\) to obtain
\[
 (t_i,Y_i)\longmapsto
 (\mathfrak F( {\theta_1,t_i}),\ldots,
  \mathfrak F( {\theta_{ N},t_i}),t_i,Y_i).
\]
An affine map forms all coordinate-wise residuals $\mathfrak{F}_k(\theta_j,t_i)-Y_{ik}$, and coordinate-wise squares followed by affine
sums then give
\[
 (r_{i1},\ldots,r_{i {N}},t_i,Y_i),\qquad
 r_{ij}:=\norm{\mathfrak F( {\theta_j,t_i})-Y_i}_2^2.
\]
A selective mean attention module is applied to the first \( {N}\) coordinates.
Every token now contains
\[
 (\ell_1,\ldots,\ell_{ {N}},t_i,Y_i),\qquad
 \ell_j=\frac1n\sum_i r_{ij}=\widetilde{\mathcal{R}}_{\theta^*}^{(n)}(\theta_j).
\]

\paragraph{Common initialization.}
Apply the map of Lemma~\ref{lem:selection-localization} to the first
\({ N}\) coordinates, preserving \((t_i,Y_i)\) throughout its reductions.
Since the loss vector is the same in every token, the output is
\[
 (\widehat\theta^0,t_i,Y_i)
\]
with the same initializer in every column.

\paragraph{One update followed by averaging.}
Define $\widehat A$ and $\widehat b$ as the outputs of $\mathrm{NN}^{\varepsilon_{\mathrm{grad}}}_{\mathrm{grad}}$ approximating \(2\nabla_\theta x(\theta,t)\) and \(2\nabla_\theta x(\theta,t)x(\theta,t)\), respectively.
Suppose every column contains \((\widehat\theta^k,t_i,Y_i)\).
\(\mathrm{NN}^{\varepsilon_{\mathrm{grad}}}_{\mathrm{grad}}\) applies to \(( {\widehat\theta^k,t_i})\), while identity channels preserve
\((\widehat\theta^k,t_i,Y_i)\). 
Its outputs are
\(\widehat A( {\widehat\theta^k,t_i})\) and
\(\widehat b( {\widehat\theta^k,t_i})\).
Compute the \(pm\) products
\(\widehat A_{ab}( {\widehat\theta^k,t_i})(Y_i)_b\) in parallel,
sum over \(b\), and take the affine combination
\[
 u_i^k:=\widehat\theta^k-\eta_\star
 [\widehat b( {\widehat\theta^k,t_i})
  -\widehat A( {\widehat\theta^k,t_i})Y_i].
\]
Discard the intermediate coordinates, retaining \((u_i^k,t_i,Y_i)\).
A selective mean attention module on the first \(p\) coordinates gives
\[
 (u_i^k,t_i,Y_i)\longmapsto
 (\widehat\theta^{k+1},t_i,Y_i),\qquad
 \widehat\theta^{k+1}=\frac1n\sum_i u_i^k
 =\widetilde T_n(\widehat\theta^k).
\]
Thus this block realizes exactly one iteration of the approximate
procedure in Theorem~\ref{thm:statistical-recovery-theorem}.
Repeat it with tied weights for \(I\) iterations, where
\[
 I:=\max\left\{1,
 \left\lceil\frac{\log(\rho/(2\varepsilon_{\mathrm{gd}}))}
                       {\log((K+\mu)/K)}\right\rceil\right\}.
\]
so that\(\big(\frac{K}{K+\mu}\big)^I\rho/2\le\varepsilon_{\mathrm{gd}}\).
\paragraph{Final output.} A final affine map projects onto the top $p$ parameter coordinates, returning
\(\widehat\theta^I\mathbf1_n^\top\).
The error bound of Theorem~\ref{thm:statistical-recovery-theorem} applies under the same conditions on $\varepsilon_{\mathrm{sol}}^{(\mathrm{obs})}$ and $\varepsilon_{\mathrm{grad}}$ to give
\eqref{eq:transformer-component-recovery} with the stated sample-size
condition and probability. All calculations use the same weights for
every \(n\), independently of \(\alpha\) and \(\theta^*\).  
\paragraph{Size accounting.}
It is clear that the construction above has depth
\[
 O({ L_{\mathrm{sol},T_{\mathrm{obs}}}(\varepsilon_{\mathrm{sol}}^{(\mathrm{obs})})}+\log(2 {N})+I{ L_{\mathrm{grad}}(\varepsilon_{\mathrm{grad}})})
\]
and width
\[
 O\bigl( {N}W_{\mathrm{sol},T_{\mathrm{obs}}}(\varepsilon_{\mathrm{sol}}^{(\mathrm{obs})})+W_{\mathrm{grad}}(\varepsilon_{\mathrm{grad}})\bigr).
\]
when counting the width and depth of the initial coarse initialization over $N$ candidates via $\mathrm{NN}^{\varepsilon_{\mathrm{sol}}^{(\mathrm{obs})}}_{\mathrm{sol},T_{\mathrm{obs}}}$, followed by $I$
iterations of the GD update block using $\mathrm{NN}^{\varepsilon_{\mathrm{grad}}}_{\mathrm{grad}}$ (we use here the fact that the width of each sub-network is at least its input and output dimension to simplify the bounds).
Using $\log(2 {N})=O(p)$ in the worst case, the total parameter count is
\[
 O(N^2P_{\mathrm{sol},T_{\mathrm{obs}}}(\varepsilon_{\mathrm{sol}}^{(\mathrm{obs})})+\log(e/\varepsilon_{\mathrm{gd}})P_{\mathrm{grad}}(\varepsilon_{\mathrm{grad}})).
\]
\end{proof}

\subsection{Proof of Corollary~\ref{cor:prediction-realization}}
\label{subsec:proof-realization-prediction}
\begin{proof}[Proof of Corollary~\ref{cor:prediction-realization}]
Ignore the query register in all recovery computations and attention
queries, keys, and values, preserving it through the residual and
feedforward layers by Lemma~\ref{lem:network-with-identity-channels}.
The final tokens are $(\widehat\theta_n^{\mathrm{out}},\tau)$; apply one
shared position-wise copy of $\mathrm{NN}^{\varepsilon_{\mathrm{sol}}^{(\mathrm{pred})}}_{\mathrm{sol},T_{\mathrm{pred}}}$ to each. Using the derivative bound along 
with the error bound from Theorem~\ref{thm:inverse-transformer-realization} gives
\[
 \|x(\widehat\theta_n^{\mathrm{out}},\tau)-x(\theta^*,\tau)\|_2
 \le \|\nabla_\theta x(\theta,t)\|_{L^\infty(\Theta\times [0,T_{\mathrm{pred}}])}\|\widehat\theta_n^{\mathrm{out}}-\theta^*\|_2.
\]
Adding the uniform solver error proves \eqref{eq:prediction-error}, and adding the width and depth of $\mathrm{NN}^{\varepsilon_{\mathrm{sol}}^{(\mathrm{pred})}}_{\mathrm{sol},T_{\mathrm{pred}}}$ to the previous construction gives the size bounds.
\end{proof}

\subsection{Proof of Corollary~\ref{cor:joint-realization}}
\label{subsec:proof-realization-joint}
\begin{proof}
Use the construction of Corollary~\ref{cor:prediction-realization}, retaining
$\widehat\theta_n^{\mathrm{out}}$ through the final solution network by the identity
channels of Lemma~\ref{lem:network-with-identity-channels}. The final affine
map returns $(\widehat\theta_n^{\mathrm{out}},\widehat x_n(\tau))$ in each token.
The parameter output is independent of $\tau$, and both error bounds hold
on the same recovery event, so no additional failure probability is needed.
The identity channels add $O(p)$ width and $O(pL_{\mathrm{sol},T_{\mathrm{pred}}})$ nonzero parameters,
which are absorbed by the stated construction bounds, without changing
their order of depth.
\end{proof}

\subsection{Euler-based trajectory and sensitivity networks}
\label{subsec:euler-nn1}
\begin{theorem}[Joint Euler realization of trajectories and sensitivities]
\label{thm:euler-trajectory-sensitivity}
Under the trajectory regularity and existence assumptions, suppose additionally
that the joint map
\[
 (u,\omega,s)\longmapsto
 \bigl(f_\omega(u,s),\nabla_u f_\omega(u,s),\nabla_\omega f_\omega(u,s)\bigr)
\]
is represented exactly by a squared-ReLU FFN on a neighborhood of the
compact trajectory set, with depth $L_f$, width $W_f$, and $P_f$ nonzero
parameters (hence is at least $C^1$ in all inputs).
Then there are constants $C,K_0$ such that, for every integer $K\ge K_0$,
a squared-ReLU FFN maps $(\theta,t)$ to $(u_{K},S_{K})$ satisfying
\[
 \sup_{\theta\in\Theta,\,t\in[0,T_{\mathrm{obs}}]}
 \bigl(\|u_{K}-x(\theta,t)\|_2+
 \|S_{K}-\nabla_\theta x(\theta,t)\|_F\bigr)\le C/K.
\]
Setting $\mathrm{NN}^{\varepsilon_{\mathrm{sol}}^{(\mathrm{obs})}}_{\mathrm{sol},T_{\mathrm{obs}}}=u_{K}$ and appending exact arithmetic to form
\[
 \mathrm{NN}^{\varepsilon_{\mathrm{grad}}}_{\mathrm{grad}}=(\operatorname{vect}(2S_{K}),2S_{K}u_{K})
\]
gives the required trajectory and coefficient errors of order $K^{-1}$.
Both networks have size bounds
\[
 L=O\bigl(K L_f\bigr),\qquad
 W=O(W_f+m^2p),\qquad
 P=O\bigl(K[P_f+m^2pL_f]\bigr),
\]
where $P$ counts nonzero parameters. Thus there is no curse of parameter dimensionality in the error exponent.\end{theorem}
\begin{proof}
We implement Euler-unrolling through a squared ReLU FFN exactly. This idea is not new\cite{grohs2019spacetime}.
We adapt it to jointly evolve the trajectory and its parameter sensitivity,
using exact squared-ReLU multiplication.
With $S=\nabla_\theta x\in\R^{p\times m}$, the augmented ODE is
\[
 \dot x=f_\omega(x,s),\qquad
 \dot S=S\nabla_u f_\omega(x,s)+\begin{bmatrix}0_{m\times m}\\ \nabla_\omega f_\omega(x,s)\end{bmatrix},
 \qquad (x,S)|_{s=0}=\left(x_0,\begin{bmatrix}I_m\\0_{d\times m}\end{bmatrix}\right).
\]
For input $(x_0,\omega,t)$, set $h=t/K$ and perform $K$ updates
\[
 \begin{aligned}
 u_{k+1}&=u_k+h f_\omega(u_k,kh),\\
 S_{k+1}&=S_k+h\left(S_k\nabla_u f_\omega(u_k,kh)+
                  \begin{bmatrix}0_{m\times m}\\\nabla_\omega f_\omega(u_k,kh)\end{bmatrix}\right).
 \end{aligned}
\]
The register $(u_k,\operatorname{vect}S_k,\omega,t)$ is preserved through
the drift FFN by identity channels, affine maps form $kh=kt/K$,
and exact multiplication and addition implement the updates.
The last step reaches $Nh=t$, for every query time using the same network.

The augmented drift is $C^1$ on a neighborhood of its compact exact
trajectory set. Thus usual arguments give Euler unrolling error of order $O(1/K)$.
\end{proof}

\section{Pretraining Guarantees}
\label{app:pretraining}
\subsection{Proof of Theorem~\ref{thm:joint-pretraining}}
\label{subsec:proof-pretraining-joint}
\begin{proof}[Proof of Theorem~\ref{thm:joint-pretraining}]
\emph{Input boundedness event.}
For a new task with true parameter \(\vartheta\in\Theta_0\), define
\[
 \mathcal E_\eta:=\left\{\max_{1\le i\le n}\|Y_i\|_\infty
                                      \le R_1^{\mathrm{joint}}(\eta)\right\}.
\]
Clearly
\[
 \Pp(\mathcal E_\eta^c\mid\vartheta)
 \le2mn\exp\left(-\frac{[R_1^{\mathrm{joint}}(\eta)-\|x\|_{L^\infty(\Theta\times [0,T_{\mathrm{obs}}])}]^2}{2\sigma^2}\right)
 \le\eta.
\]
This bound is uniform in \(\vartheta\) and in the observation times.
Since \(t_i\in[0,T_{\mathrm{obs}}]\), \(\tau\in[0,T_{\mathrm{pred}}]\) and \(R_1^{\mathrm{joint}}\ge T_{\mathrm{pred}}\), clipping never changes the
time coordinates or the query time. On \(\mathcal E_\eta\), it also leaves all observation
coordinates unchanged, so the entire dataset is unchanged.

\emph{Comparator risk.}
For each fixed \(\vartheta\in\Theta_0\), let \(\mathcal G_{\vartheta,\alpha}\)
be the recovery event from Theorem~\ref{thm:inverse-transformer-realization}, so that
\(\Pp(\mathcal G_{\vartheta,\alpha}\cap\mathcal E_\eta\mid\vartheta)\ge1-7\alpha-\eta\).
Let $g_{\mathrm{joint}}^\circ$ be the joint construction with the prescribed
clipping, projections and readout.
On \(\mathcal G_{\vartheta,\alpha}\cap\mathcal E_\eta\),
initial clipping and all intermediate projections are inactive.
The target $Z=(\vartheta,x(\vartheta,\tau))$ has norm at most
$D_{\mathrm{joint}}/2$, so the final projection cannot increase its squared error.
On the common recovery and input-boundedness event, the squared error is
therefore at most $e_n^{\mathrm{est}}(\alpha)^2+e_n^{\mathrm{pred}}(\alpha)^2$ upto a universal constant factor, by Corollary~\ref{cor:joint-realization}.
On the complement, the squared error is at most $D_{\mathrm{joint}}^2$.
Averaging over the task and query time therefore gives
\[
 \mathcal R_{\mathrm{joint}}(g_{\mathrm{joint}}^\circ)
 \lesssim e_n^{\mathrm{est}}(\alpha)^2+e_n^{\mathrm{pred}}(\alpha)^2
       +D_{\mathrm{joint}}^2(7\alpha+\eta).
\]
The same bound controls its excess risk because the Bayes risk is
nonnegative. Apply Theorem~\ref{thm:generic-pretraining} with
$U=\mathcal D_\tau^{(n)}$, $Z=(\vartheta,x(\vartheta,\tau))$ and
$R_3=B_Z=D_{\mathrm{joint}}/2$, so that $B_*=D_{\mathrm{joint}}$.
Substitution yields \eqref{eq:joint-pretraining}.
\end{proof}

\subsection{Pretraining guarantee for parameter estimation}
\label{subsec:proof-pretraining-parameter}
We consider the estimation transformer from Theorem~\ref{thm:inverse-transformer-realization}. Fix tolerances $\varepsilon_{\mathrm{sol}}^{(\mathrm{obs})},\varepsilon_{\mathrm{grad}},\varepsilon_{\mathrm{gd}}$ satisfying the conditions of that theorem, and let $\mathcal T_{\mathrm{est}}^{\varepsilon_{\mathrm{sol}}^{(\mathrm{obs})},\varepsilon_{\mathrm{grad}},\varepsilon_{\mathrm{gd}}}$ denote the resulting construction. Since this transformer has finitely many parameters, we can choose $M_A^{\mathrm{est}}$ and $M_b^{\mathrm{est}}$ sufficiently large to accommodate all its weights and biases.

We choose the depth and width of the bounded transformer class to match those of the constructed estimator, so that
\begin{align*}
L_{\mathrm{est}}
&=O\left(
L_{\mathrm{sol},T_{\mathrm{obs}}}(\varepsilon_{\mathrm{sol}}^{(\mathrm{obs})})+\log(2N)
+\log(e/\varepsilon_{\mathrm{gd}})L_{\mathrm{grad}}(\varepsilon_{\mathrm{grad}})
\right),\\
W_{\mathrm{est}}
&=O\left(
NW_{\mathrm{sol},T_{\mathrm{obs}}}(\varepsilon_{\mathrm{sol}}^{(\mathrm{obs})})+W_{\mathrm{grad}}(\varepsilon_{\mathrm{grad}})
\right),
\end{align*}
where the dependence of $L_{\mathrm{est}}$ and $W_{\mathrm{est}}$ on the approximation tolerances is suppressed for brevity.
To incorporate the clipping and projections without affecting the comparator error bound, we choose suitable radii. For $\eta\in(0,1)$, set
\[
R_1^{\mathrm{est}}(\eta)
:=
\max\left\{
T_{\mathrm{obs}},\,
\|x\|_{L^\infty(\Theta\times[0,T_{\mathrm{obs}}])}
+\sigma\sqrt{2\log(2mn/\eta)}
\right\},
\]
and $R_3^{\mathrm{est}}:=D/2$. The input clipping then leaves the observations unchanged with probability at least $1-\eta$. Since the clipped inputs are bounded and the transformer has finitely many layers, we can choose $R_2^{\mathrm{est}}$ sufficiently large so that the intermediate projections do not alter the constructed computation. For our fixed transformer, a worst-case bound on $R_2^{\mathrm{est}}$ can be obtained by propagating the input magnitude through its layers, yielding polynomial dependence on $(R_1^{\mathrm{est}})^L$, with degree independent of $n$.

We therefore define the estimation class
\[
\mathcal A_{\mathrm{est}}
:=
\mathfrak T^{R_1^{\mathrm{est}}(\eta),R_2^{\mathrm{est}},R_3^{\mathrm{est}},
M_A^{\mathrm{est}},M_b^{\mathrm{est}}}_{m+1,p,W_{\mathrm{est}},L_{\mathrm{est}},H,j},
\]
where $H\ge1$ and $j\in\{1,\ldots,n\}$. This class contains the estimation construction with the corresponding clipping, projections, and readout. Let $P_{\mathrm{est}}$, $R_\Phi^{\mathrm{est}}$, and $\mathfrak L_\Phi^{\mathrm{est}}$ denote its number of trainable parameters, parameter-domain radius, and parameter-Lipschitz constant, respectively, as characterized in Lemma~\ref{lem:transformer-parameter-bounds}.

The following corollary combines the parameter recovery guarantee with the pretraining oracle inequality.

\begin{corollary}[Pretraining guarantee for parameter estimation]
\label{cor:recovery-pretraining}
Under the conditions of Theorem~\ref{thm:inverse-transformer-realization}, fix $\eta\in(0,1)$, $\alpha\in(0,1/7)$, and $n>\mathfrak N(\alpha)$. For any $\delta\in(0,1)$ and any $\varepsilon_{\mathrm{opt}}$-approximate empirical pretraining risk minimizer $\widehat g_{\mathrm{est}}$ over $\mathcal A_{\mathrm{est}}$, we have
\begin{equation}
\begin{aligned}
\mathcal R_{\mathrm{est}}(\widehat g_{\mathrm{est}})
-\mathcal R_{\mathrm{est}}(g_{\mathrm{est}}^{\mathrm{bayes}})
\lesssim\;&
e_n^{\mathrm{est}}(\alpha)^2+D^2(7\alpha+\eta)\\
&+\frac{D^2}{M}
\left[
P_{\mathrm{est}}\log\left(
e+\frac{
M\mathfrak L_\Phi^{\mathrm{est}}(n/\eta)
R_\Phi^{\mathrm{est}}}{D}
\right)
+\log\frac{2}{\delta}
\right]
+\varepsilon_{\mathrm{opt}},
\end{aligned}
\label{eq:recovery-pretraining-rate}
\end{equation}
with probability at least $1-\delta$ over the pretraining tasks, where the implied constant is universal and $\mathfrak L_\Phi^{\mathrm{est}}(n/\eta)$ is polylogarithmic in $n/\eta$.
\end{corollary}

\begin{proof}[Proof of Corollary~\ref{cor:recovery-pretraining}]
Let $g_{\mathrm{est}}^\circ$ correspond to the element in the function class for $\mathcal T_{\mathrm{est}}^{\varepsilon_{\mathrm{sol}}^{(\mathrm{obs})},\varepsilon_{\mathrm{grad}},\varepsilon_{\mathrm{gd}}}$ composed with the relevant clipping, projections and readout (from any position $j$).
Since \(t_i\in[0,T_{\mathrm{obs}}]\) and \(R_1^{\mathrm{est}}\ge T_{\mathrm{obs}}\), clipping never changes the time coordinates.
On the intersection of the recovery event and the input boundedness
event from the proof of Theorem~\ref{thm:joint-pretraining}, which has probability at least $1-7\alpha-\eta$, initial clipping and all intermediate projections are inactive, and the
final projection is onto \(\Theta=\overline B(0,D/2)\), so that
\(\|g_{\mathrm{est}}^\circ(\mathcal D)-\vartheta\|_2\lesssim e_n^{\mathrm{est}}(\alpha)\).
For every dataset, both \(g_{\mathrm{est}}^\circ(\mathcal D)\) and \(\vartheta\)
belong to \(\Theta\), so their distance is at most its diameter \(D\).
The same conditional union bound and averaging over \(\vartheta\sim\Pi\) therefore give
\[
 \mathcal{R}_{\mathrm{est}}(g_{\mathrm{est}}^\circ)
 \lesssim e_n^{\mathrm{est}}(\alpha)^2+D^2(7\alpha+\eta),
\]
which also bounds the approximation term since \(\mathcal{R}_{\mathrm{est}}(g_{\mathrm{est}}^{\mathrm{bayes}})\ge0\).
Finally apply Theorem~\ref{thm:generic-pretraining} with $U=\mathcal{D}^{(n)},Z=\vartheta,R_3=B_Z=D/2$, and substitute this approximation bound. This proves \eqref{eq:recovery-pretraining-rate}.
\end{proof}

\subsection{Pretraining guarantee for trajectory prediction}
\label{subsec:proof-pretraining-prediction}
We next apply the same argument to the prediction transformer from Corollary~\ref{cor:prediction-realization}. Define
\[
D_{\mathrm{pred}}:=2\max\{1,\|x\|_{L^\infty(\Theta\times[0,T_{\mathrm{pred}}])}\},
\]
which provides a uniform bound on the prediction target.

Fix the approximation tolerances $\varepsilon_{\mathrm{sol}}^{(\mathrm{obs})},\varepsilon_{\mathrm{grad}},\varepsilon_{\mathrm{gd}},\varepsilon_{\mathrm{sol}}^{(\mathrm{pred})}$ as in Corollary~\ref{cor:prediction-realization}, and let $\mathcal T_{\mathrm{pred}}^{\varepsilon_{\mathrm{sol}}^{(\mathrm{obs})},\varepsilon_{\mathrm{grad}},\varepsilon_{\mathrm{gd}},\varepsilon_{\mathrm{sol}}^{(\mathrm{pred})}}$ denote the corresponding construction. As before, we choose the depth and width of the bounded transformer class to match this construction, with
\begin{align*}
L_{\mathrm{pred}}
&=O\left(
L_{\mathrm{sol},T_{\mathrm{obs}}}(\varepsilon_{\mathrm{sol}}^{(\mathrm{obs})})+\log(2N)
+\log(e/\varepsilon_{\mathrm{gd}})L_{\mathrm{grad}}(\varepsilon_{\mathrm{grad}})
+L_{\mathrm{sol},T_{\mathrm{pred}}}(\varepsilon_{\mathrm{sol}}^{(\mathrm{pred})})
\right),\\
W_{\mathrm{pred}}
&=O\left(
NW_{\mathrm{sol},T_{\mathrm{obs}}}(\varepsilon_{\mathrm{sol}}^{(\mathrm{obs})})+W_{\mathrm{grad}}(\varepsilon_{\mathrm{grad}})
+W_{\mathrm{sol},T_{\mathrm{pred}}}(\varepsilon_{\mathrm{sol}}^{(\mathrm{pred})})
\right).
\end{align*}
For $\eta\in(0,1)$, set
\[
R_1^{\mathrm{pred}}(\eta)
:=
\max\left\{
T_{\mathrm{pred}},\,
\|x\|_{L^\infty(\Theta\times[0,T_{\mathrm{obs}}])}
+\sigma\sqrt{2\log(2mn/\eta)}
\right\},
\]
and $R_3^{\mathrm{pred}}:=D_{\mathrm{pred}}/2$. We choose $R_2^{\mathrm{pred}}$, $M_A^{\mathrm{pred}}$, and $M_b^{\mathrm{pred}}$ sufficiently large to accommodate the prediction construction without altering its computation on the event that input clipping is inactive.

The resulting prediction class is
\[
\mathcal A_{\mathrm{pred}}
:=
\mathfrak T^{R_1^{\mathrm{pred}}(\eta),R_2^{\mathrm{pred}},R_3^{\mathrm{pred}},
M_A^{\mathrm{pred}},M_b^{\mathrm{pred}}}_{m+2,m,W_{\mathrm{pred}},L_{\mathrm{pred}},H,j},
\]
where $H\ge1$ and $j\in\{1,\ldots,n\}$. As in the estimation case, let $P_{\mathrm{pred}}$, $R_\Phi^{\mathrm{pred}}$, and $\mathfrak L_\Phi^{\mathrm{pred}}$ denote its number of trainable parameters, parameter-domain radius, and parameter-Lipschitz constant, respectively.

\begin{corollary}[Pretraining guarantee for trajectory prediction]
\label{cor:prediction-pretraining}
Under the conditions of Corollary~\ref{cor:prediction-realization}, fix $\eta\in(0,1)$, $\alpha\in(0,1/7)$, and $n>\mathfrak N(\alpha)$. For any $\delta\in(0,1)$ and any $\varepsilon_{\mathrm{opt}}$-approximate empirical pretraining risk minimizer $\widehat g_{\mathrm{pred}}$ over $\mathcal A_{\mathrm{pred}}$, we have
\begin{equation}
\begin{aligned}
\mathcal R_{\mathrm{pred}}(\widehat g_{\mathrm{pred}})
-\mathcal R_{\mathrm{pred}}(g_{\mathrm{pred}}^{\mathrm{bayes}})
\lesssim\;&
\left(
\varepsilon_{\mathrm{sol}}^{(\mathrm{pred})}+
\|\nabla_\theta x\|_{L^\infty(\Theta\times[0,T_{\mathrm{pred}}])}
e_n^{\mathrm{est}}(\alpha)
\right)^2
+D_{\mathrm{pred}}^2(7\alpha+\eta)\\
&+\frac{D_{\mathrm{pred}}^2}{M}
\left[
P_{\mathrm{pred}}\log\left(
e+\frac{
M\mathfrak L_\Phi^{\mathrm{pred}}(n/\eta)
R_\Phi^{\mathrm{pred}}}{D_{\mathrm{pred}}}
\right)
+\log\frac{2}{\delta}
\right]
+\varepsilon_{\mathrm{opt}},
\end{aligned}
\label{eq:prediction-pretraining}
\end{equation}
with probability at least $1-\delta$ over the pretraining tasks, where the implied constant is universal and $\mathfrak L_\Phi^{\mathrm{pred}}(n/\eta)$ is polylogarithmic in $n/\eta$.
\end{corollary}

\begin{proof}[Proof of Corollary~\ref{cor:prediction-pretraining}]
Use the constructed prediction transformer $\mathcal T_{\mathrm{pred}}^{\varepsilon_{\mathrm{sol}}^{(\mathrm{obs})},\varepsilon_{\mathrm{grad}},\varepsilon_{\mathrm{gd}},\varepsilon_{\mathrm{sol}}^{(\mathrm{pred})}}$ composed with the relevant clipping, projections and readout to define the comparator $g_{\mathrm{pred}}^\circ$.
On the intersection of the recovery event and the input boundedness
event from the proof of Theorem~\ref{thm:joint-pretraining}, clipping
and intermediate projections are inactive. Input clipping preserves
$\tau$ since its threshold is at least $T_{\mathrm{pred}}$. Output projection preserves the comparator error, and hence it is
at most $\varepsilon_{\mathrm{sol}}^{(\mathrm{pred})}+\|\nabla_\theta x(\theta,t)\|_{L^\infty(\Theta\times [0,T_{\mathrm{pred}}])} e_n^{\mathrm{est}}(\alpha)$ on this event upto some universal constant factor.
The same conditional union bound and averaging over tasks therefore give
\[
 \mathcal R_{\mathrm{pred}}(g_{\mathrm{pred}}^\circ)
 \lesssim(\varepsilon_{\mathrm{sol}}^{(\mathrm{pred})}+\|\nabla_\theta x(\theta,t)\|_{L^\infty(\Theta\times [0,T_{\mathrm{pred}}])}e_n^{\mathrm{est}}(\alpha))^2
       +{ D_{\mathrm{pred}}^2}(7\alpha+\eta).
\]
Just as in the proof of Theorem~\ref{thm:joint-pretraining}, we can now apply Theorem~\ref{thm:generic-pretraining} with $U=(\mathcal{D}^{(n)}_\tau),Z=x(\vartheta,\tau),R_3=B_Z=D_{\mathrm{pred}}/2$, and substitute this approximation bound. This proves \eqref{eq:prediction-pretraining}.
\end{proof}

\section{Additional Experimental Details}
\label{app:experiments}

\medskip

\paragraph{Data generation and normalization}
For each task, we draw a parameter $\theta$ from the corresponding uniform prior and sample $n_{\mathrm{train}}$ observation times independently from $\mathcal U[0,3.2]$. For state coordinate $k$, we generate $y_{i,k}=x_{\theta,k}(t_i)+qS_{X,k}\xi_{i,k}$, where $\xi_{i,k}\overset{\mathrm{iid}}{\sim}\mathcal N(0,1)$ and $q=0.02$. The coordinate scales are determined by Monte Carlo estimates of $S_{X,k}^2=\E_{\theta,t}[x_{\theta,k}(t)^2]$, with the expectation taken over the parameter prior and observation-time distribution. The reported state scale values for normalization are $0.653$ for forced linear dynamics, $0.716$ for logistic growth, $1.007$ for birth--death dynamics, and $(0.656,0.680)$ for Stuart--Landau. Dividing each observed coordinate by $S_{X,k}$ gives a normalized noise standard deviation of $0.02$ across families.
For parameter estimation, observation $i$ is encoded as $(t_i/4,y_{i,1}/S_{X,1},\ldots,y_{i,m}/S_{X,m})$. A parameter coordinate with prior interval $[\ell_j,u_j]$ is normalized to $z_j=1.6(\theta_j-(\ell_j+u_j)/2)/(u_j-\ell_j)\in[-0.8,0.8]$. For prediction, we draw $\tau\sim\mathcal U[0,4]$, append $\tau/4$ to every input token, and use the normalized clean state $x_\theta(\tau)/S_X$ as the target at each output token, with division understood coordinate-wise. Because prediction times can exceed the observation interval, the prediction task includes extrapolation as well as interpolation.

The validation and main test sets contain 4,096 and 8,192 independently generated tasks, respectively. Comparisons with statistical procedures use 256 paired tasks per seed because of computational cost. Parameter estimation and trajectory prediction use the same noisy observations, with the query time additionally included for prediction. We do not use positional embeddings or causal masks.
Moroever, within each seed, shared parameter draws and prefixes of sampled observation sequences allow paired comparisons across values of $M$ and $n$.

Training and inference were performed on a single NVIDIA A100 PCIe 80\,GB GPU using PyTorch 2.8.0, CUDA 12.8, and float32 arithmetic.

\medskip
\paragraph{Architecture size}
Our default transformer has $16$ blocks with token width $64$. Each block consists of a squared-ReLU feedforward network
of dimensions $64 \rightarrow 256 \rightarrow 64$, followed by two-head attention. We use pre-RMSNorm, residual
connections, and dropout of $0.1$ on residual updates. Each attention head has separate $64 \times 64$
query, key, and value matrices, with the head outputs summed and attention scores scaled by
$1/\sqrt{64}$. 

\medskip
\paragraph{Training and evaluation}
We train using AdamW with initial learning rate $3\times10^{-4}$, weight decay $10^{-4}$, batch size 256, and gradient-norm clipping at 1. A 5\% linear warm-up is followed by cosine learning-rate decay over 30 epochs, and we select the model with the lowest validation loss. Dropout is disabled during evaluation.

For the forced-linear family, we vary $n_{\mathrm{train}}\in\{2,8,32,128\}$ and $M\in\{2^{11},2^{13},2^{15},2^{17},2^{19}\}$ for both tasks. For the other families, we fix $M=2^{19}$ and use $n_{\mathrm{train}}\in\{8,128\}$. Each model is trained with two random seeds, 20260906 and 20260907.

All plots report results averaged over the two seeds, except for the Gaussian prior experiment, which uses only seed 20260906. Shaded regions represent pointwise 95\% percentile-bootstrap confidence intervals obtained from 5,000 resamples of test tasks, computed separately for each trained model.

\FloatBarrier
\subsection{Depth, width, and head-count scaling}
\label{app:exp-architecture}
The architecture screen uses the forced-linear parameter task, $M=2^{19}$, training $n=128$, 30 epochs, and the same optimizer and validation-selection protocol as the main experiments. Each variant is trained with both seeds and evaluated over inference contexts $2$ through $512$. The default configuration is reused from the main grid. Nine additional variants per seed change one factor at a time: mean-output supervision, ReLU activation, one/four/eight heads, eight/thirty-two blocks, and half/double width. This gives 18 additional trained models.

\begin{figure}[!htbp]
 \centering
 \includegraphics[width=.49\linewidth,trim=0bp 12.5bp 0bp 0bp,clip]{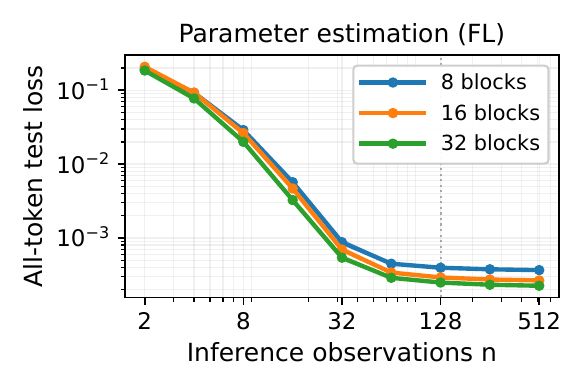}\hfill
 \includegraphics[width=.49\linewidth,trim=0bp 12.5bp 0bp 0bp,clip]{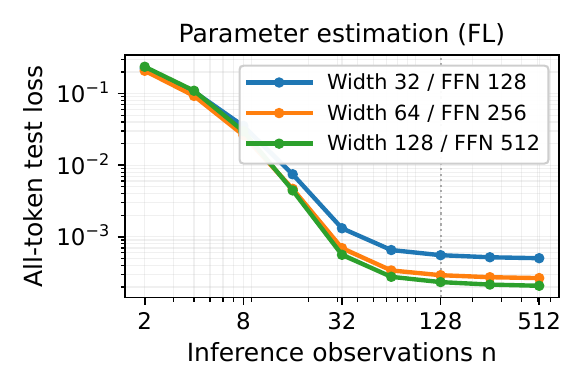}%
 \caption{Depth (left: 8, 16, 32 blocks) and width (right: 32, 64, 128) at fixed $M=2^{19}$ and training $n=128$. Models use two global heads.}
 \label{fig:exp-architecture-grid}
\end{figure}

Figure~\ref{fig:exp-architecture-grid} varies depth over $8,16,32$ with width 64, and width over $32,64,128$ with depth 16. The feed-forward width scales as four times the hidden width. Larger models can improve performance, but the exact effect varies with inference context and seed. Thus we stick to depth 16, width 64 models for the main experiments.

\begin{figure}[!htbp]
 \centering
 \includegraphics[width=.7\linewidth,trim=0bp 12.4bp 0bp 0bp,clip]{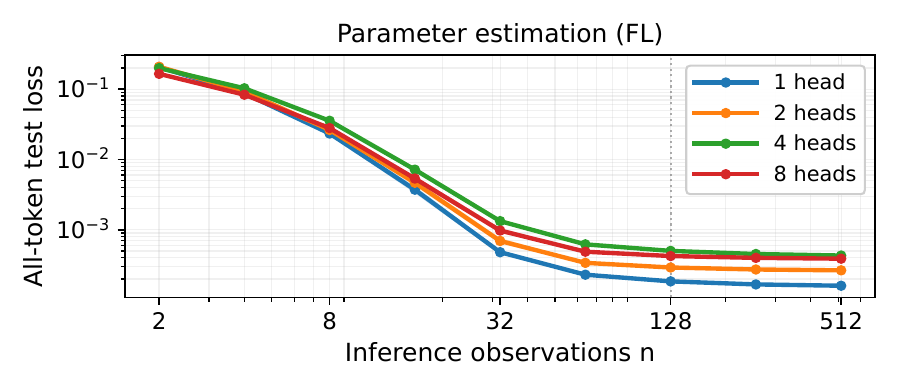}%
 \caption{Varying full-width global heads at fixed depth 16, width 64, $M=2^{19}$, and training $n=128$. Each head has square query, key, and value matrices and increasing head count increases parameter count.}
 \label{fig:exp-depth-heads}
\end{figure}

Figure~\ref{fig:exp-depth-heads} changes only the number of global heads, over $1,2,4,8$. Every head retains the full hidden width, and its output is added to the others. Interestingly, the single and double-headed transformers perform better over varying inference context-lengths, including the training context-length, as compared to four or eight heads, even if with substantially lower number of parameters. Thus, as higher number of heads are strictly more expressive than one head, the gap in performance can only be due to optimization difficulty. Our construction in Theorem~\ref{thm:inverse-transformer-realization} also says that the number of heads does not correlate to better performance in that construction, and so it could be interesting to see if this can be justified more rigorously. For minimal expressivity, we retain two heads for our main experiments.
\FloatBarrier
\subsection{Activation and readout screens}
\label{subsec:experiments-architecture}
\begin{figure}[!htbp]
 \centering
 \includegraphics[width=.49\linewidth,trim=0bp 12.5bp 0bp 0bp,clip]{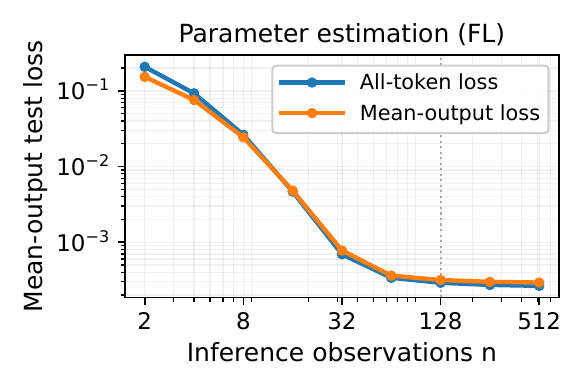}\hfill
 \includegraphics[width=.49\linewidth,trim=0bp 12.5bp 0bp 0bp,clip]{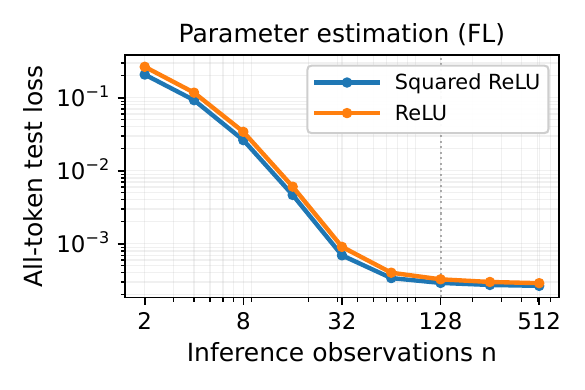}%
 \caption{Readout loss (left) and activation (right). The readout panel evaluates mean-output MSE for both the $L_{\mathrm{all}}$ and $L_{\mathrm{mean}}$ models. The activation panel evaluates all-token MSE for squared ReLU and ReLU.}
 \label{fig:architecture-grid}
\end{figure}

The activation comparison changes squared ReLU to ReLU while retaining the default architecture, data, and all-token loss. The readout comparison instead changes the training objective to
\[
 L_{\mathrm{mean}}=\frac{1}{d_z}\left\|\frac{1}{n}\sum_{i=1}^n\widehat z_i-z\right\|_2^2.
\]
For this panel only, both networks are evaluated using mean-output parameter MSE, including the network trained with $L_{\mathrm{all}}$. This compares the quality of the aggregate estimator without adding a token-disagreement penalty to only one curve. The identity
\[
 L_{\mathrm{all}}=L_{\mathrm{mean}}
 +\frac{1}{nd_z}\sum_{i=1}^n\|\widehat z_i-\overline z\|_2^2,\qquad
 \overline z=\frac1n\sum_{i=1}^n\widehat z_i,
\]
explains the extra agreement constraint imposed by all-token supervision. Figure~\ref{fig:architecture-grid} reports these two screens. ReLU seems to perform only slightly worse than squared ReLU, and the mean-output loss similarly slightly worse than the all-token loss. 

\begin{remark}
 Other token-specific readouts do not make sense for inference context-length scaling due to lack of positional encoding since there is no distinguished token position to keep track of for larger inference contexts.One could sort the trajectory data and this could impose some sort of probabilistic positional encoding w.r.t the time coordinate, but it is not clear how inference-context scaling would make sense for such a model.
\end{remark}
\begin{figure}[!htbp]
 \centering
 \includegraphics[width=0.9\linewidth,trim=0bp 20.1bp 0bp 0bp,clip]{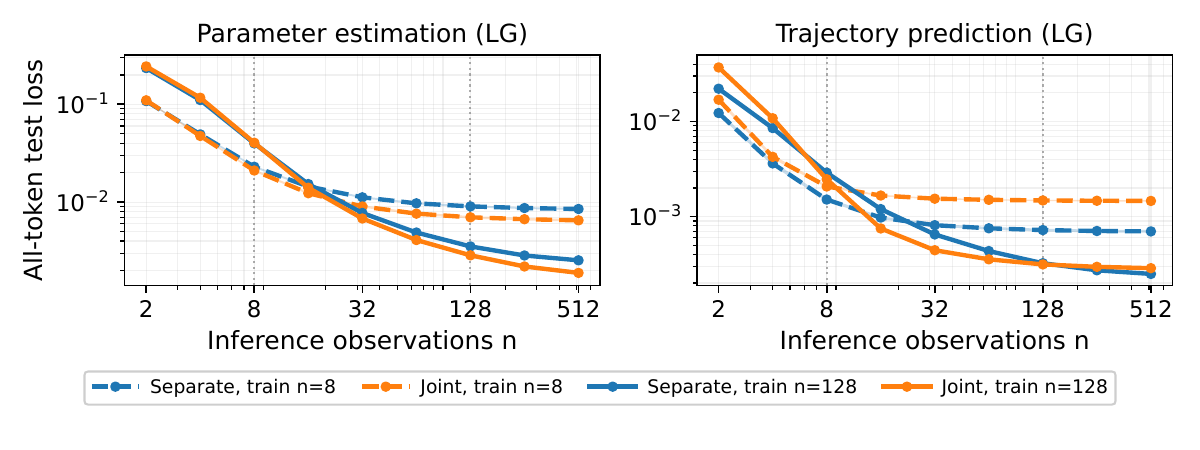}%
 \caption{Logistic growth: joint versus separate models at $M=2^{19}$ and training $n=8,128$.}
 \label{fig:exp-families}
\end{figure}
\begin{figure}[!htbp]
 \centering
 \includegraphics[width=0.9\linewidth,trim=0bp 20.1bp 0bp 0bp,clip]{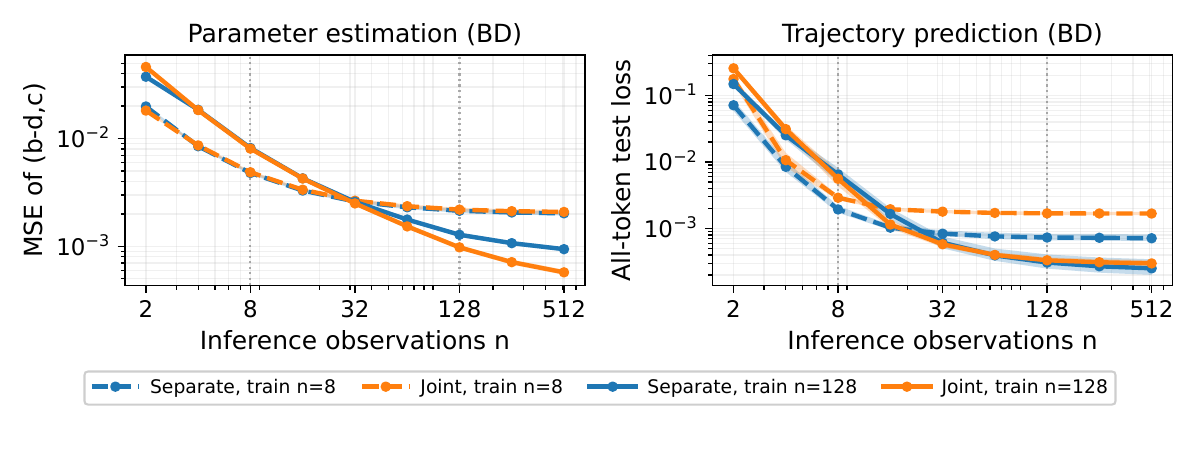}%
 \caption{Birth--death: joint versus separate models at $M=2^{19}$ and training $n=8,128$. }
 \label{fig:exp-birth-scaling}
\end{figure}

\begin{figure}[!htbp]
 \centering
 \includegraphics[width=0.9\linewidth,trim=0bp 20.1bp 0bp 0bp,clip]{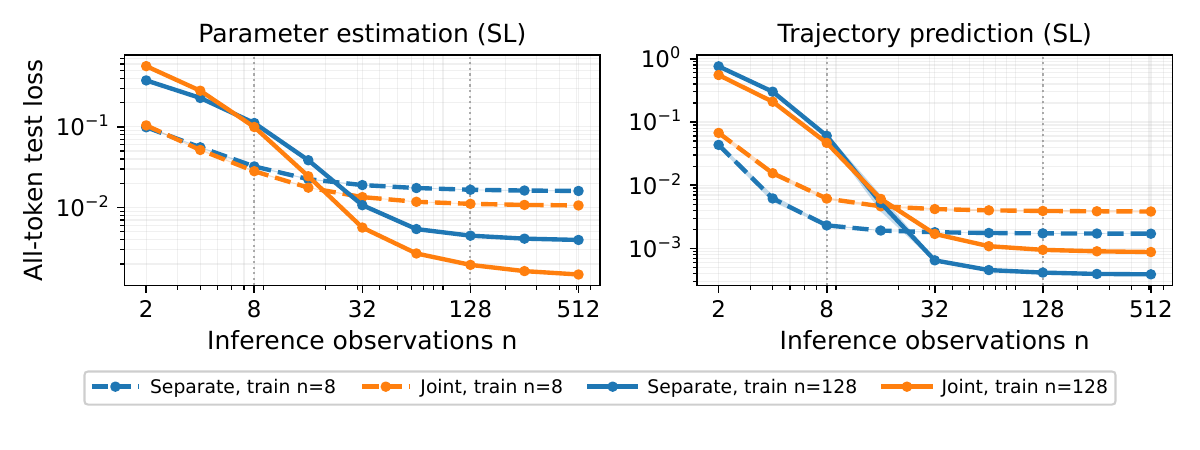}%
 \caption{Stuart--Landau: joint versus separate models at $M=2^{19}$ and training $n=8,128$.}
 \label{fig:exp-stuart-scaling}
\end{figure}
\subsection{Other-family scaling}
\label{app:exp-statistics}
\label{app:exp-family-results}
For logistic growth, birth--death, and Stuart--Landau, each task is trained at $M=2^{19}$ and $n_{\mathrm{train}}=8$ or $128$ with both seeds. Figures~\ref{fig:exp-families}--\ref{fig:exp-stuart-scaling} retain these fixed networks and vary inference context over $2,4,\ldots,512$. The scalar families use three parameter coordinates; Stuart--Landau has four parameters and two observed state coordinates. Inputs, targets, and noise use the same normalization as in the main section.

For logistic growth, results are similar qualitatively to the linear problem. Birth--death is evaluated through the identifiable pair $(b-d,c)$ on the parameter side. Its trajectory can be predicted accurately despite ambiguity in the separate labeled rates. Stuart--Landau also benefits from more observations, but the network trained at $n=128$ can perform poorly when evaluated on substantially shorter contexts. This is likely due to oscillatory dynamics, and the existence of limit cycles which make it difficult to extrapolate or estimate parameters from a small number of samples. 

\subsection{Gaussian prior statistical comparison}
\FloatBarrier
For the additional forced-linear comparison in the main paper, we replace the uniform parameter prior by a truncated Gaussian while retaining the parameter box $[-0.8,0.8]^3$. The coordinates are independent with $\theta_j\sim\mathcal N(0.2,S^2)$ conditioned on $[-0.8,0.8]$. Here $S\in\{0.2,0.1\}$ is the standard deviation before truncation; smaller $S$ means greater concentration around the mode. Samples are generated by inverse-CDF sampling from the truncated distributions. We cross these priors with normalized noise levels $q\in\{0.1,0.2\}$, giving four parameter-estimation models at $n_{\mathrm{train}}=8$ and $M=2^{19}$. 

For the statistical comparison, all methods receive the same first 256 test trajectories at each $n\in\{2,4,8,16,32,64,128,256,512\}$. We evaluate the transformer's mean output using squared error averaged over normalized parameter coordinates. Bounded MLE uses \texttt{scipy.optimize.lsq\_linear} on the original box, with tolerance $10^{-11}$ and up to 1,000 iterations. At $n=2$, the least-squares minimizer can be nonunique, and so the reported estimate is the solver's selected minimizer. Projected SGD starts at the box midpoint and uses 3,000 updates, observation minibatches of size $\min\{8,n\}$ sampled with replacement, and step size $0.05/(1+k/200)^{0.6}$, projecting onto the box after each update.

The posterior mean uses the corresponding truncated prior and Gaussian likelihood with known noise scale. We first compute the untruncated Gaussian posterior by completing the square, then integrate over the box using sequential truncated-normal importance sampling with scrambled Sobol points. We use two independent scrambles and successively $2^{10},2^{12},2^{14},2^{16}$ points per scramble. Stopping requires successive posterior means and scramble deviations to agree within $10^{-4}+0.005\|\widehat\theta\|_\infty$ and effective sample size at least 128. All reported estimates passed these numerical checks.

\subsection{Other statistical comparisons}
\begin{figure}[!htbp]
 \centering
 \includegraphics[width=\linewidth,trim=0bp 0.2bp 0bp 0bp,clip]{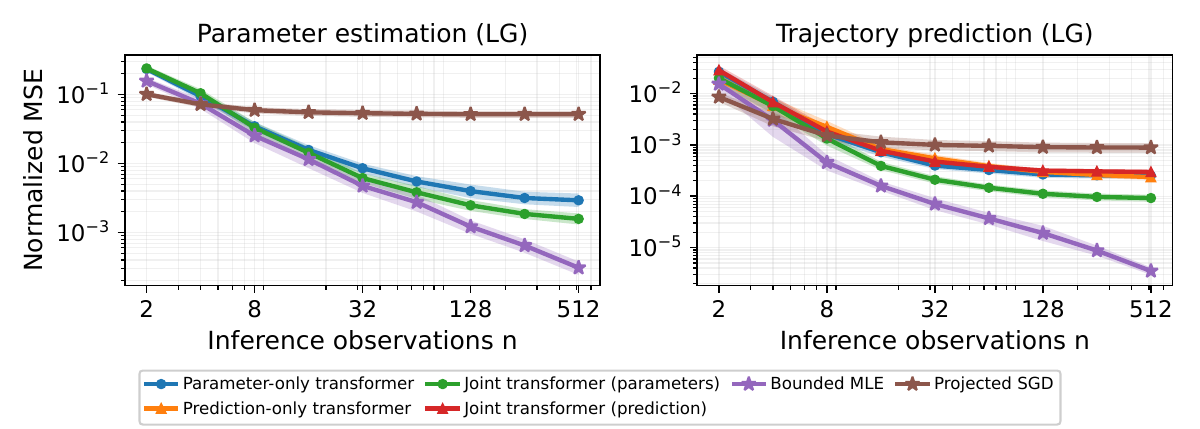}%
 \caption{Logistic statistical comparisons with neural models use $M=2^{19}$ and training $n=128$. Prediction baselines compose parameter estimates with the exact solution.}
 \label{fig:exp-logistic-statistics}
\end{figure}

\begin{figure}[!htbp]
 \centering
 \includegraphics[width=\linewidth,trim=0bp 0.2bp 0bp 0bp,clip]{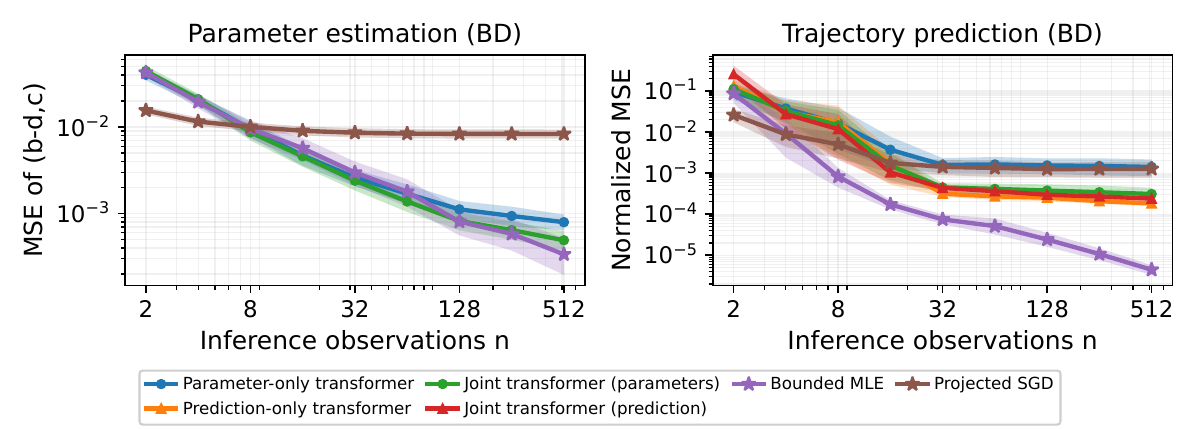}%
 \caption{Birth--death statistical comparisons with neural models use $M=2^{19}$ and training $n=128$. Prediction baselines compose parameter estimates with the exact solution.}
 \label{fig:exp-identifiable}
\end{figure}

\begin{figure}[!htbp]
 \centering
 \includegraphics[width=\linewidth,trim=0bp 0.2bp 0bp 0bp,clip]{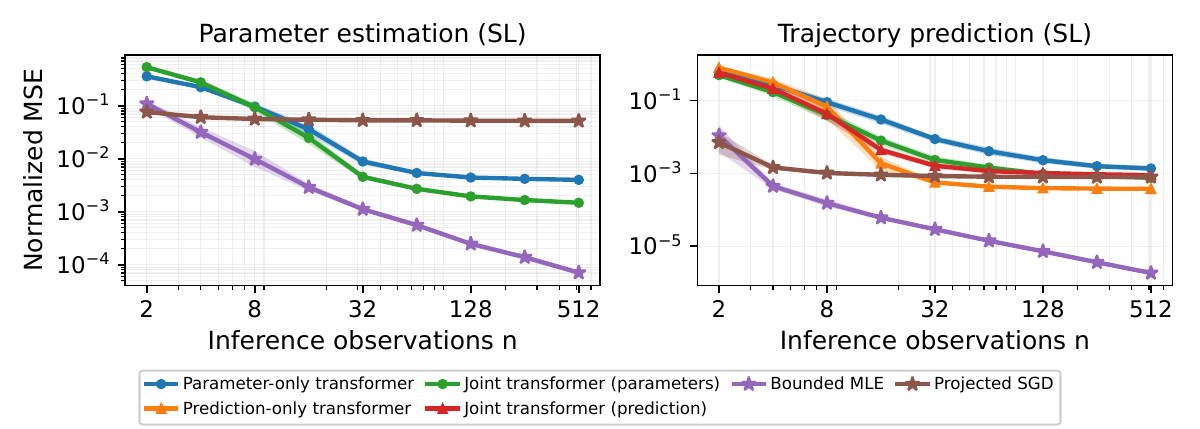}%
 \caption{Stuart--Landau statistical comparisons with neural models use $M=2^{19}$ and training $n=128$. Prediction baselines compose parameter estimates with the exact solution.}
 \label{fig:exp-stuart-statistics}
\end{figure}

We next compare the pretrained transformers with classical statistical estimators under a uniform parameter distribution. These include the maximum likelihood estimator (MLE) and projected stochastic gradient descent (SGD) using the exact trajectory-matching loss for parameter estimation, and the posterior mean for the FL family. 
While for the FL family, the MLE reduces to box-constrained linear least squares, nonlinear least squares is used for the other families via \texttt{scipy.optimize.least\_squares}. The objective is minimized from the prior midpoint and four random initializations, with up to 1,000 function evaluations per start, and we retain the fit with the smallest residual. That said, the method does not certify a global minimum. Projected SGD and posterior mean are calculated similarly as before. Figures~\ref{fig:exp-logistic-statistics}--\ref{fig:exp-stuart-statistics} shows these comparisons.

\begin{figure}[!tbp]
\centering
\includegraphics[width=\linewidth,trim=0bp 0.2bp 0bp 0bp,clip]{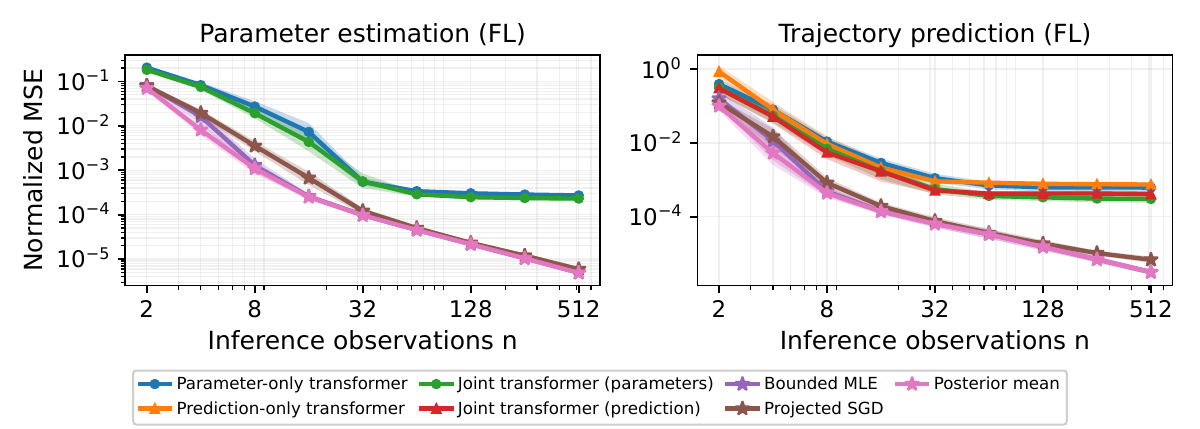}%
 \caption{Forced-linear statistical comparisons with neural models use $M=2^{19}$ and training $n=128$. Prediction baselines compose parameter estimates with the exact solution.}
\label{fig:exp-linear-statistics}
\end{figure}

\medskip

\FloatBarrier
\begin{figure}[!htbp]
 \centering
 \includegraphics[width=.9\linewidth,trim=0bp 5.5bp 0bp 0bp,clip]{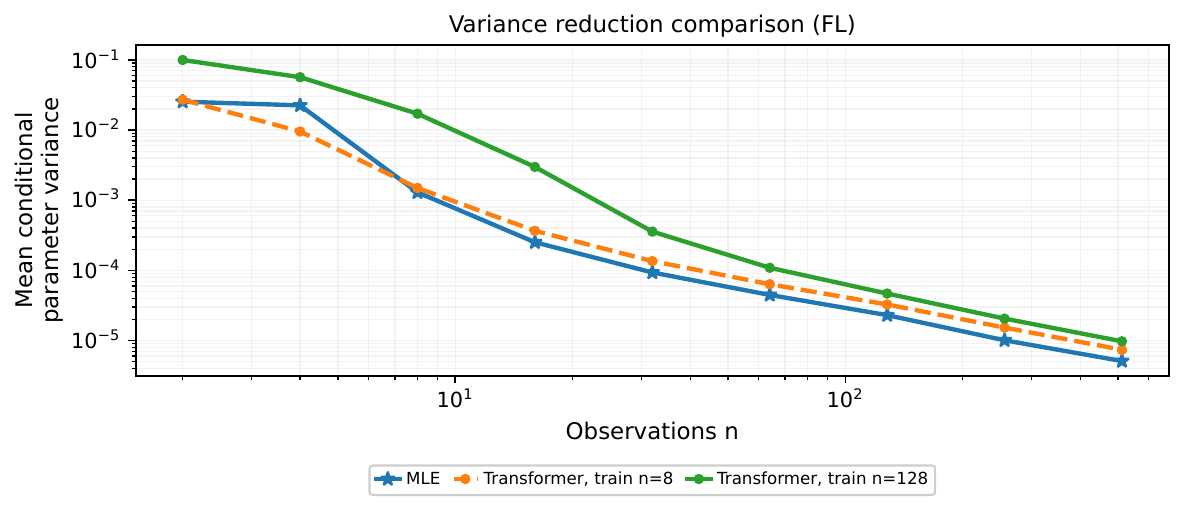}\par\smallskip
 \includegraphics[width=.9\linewidth,trim=0bp 5.5bp 0bp 0bp,clip]{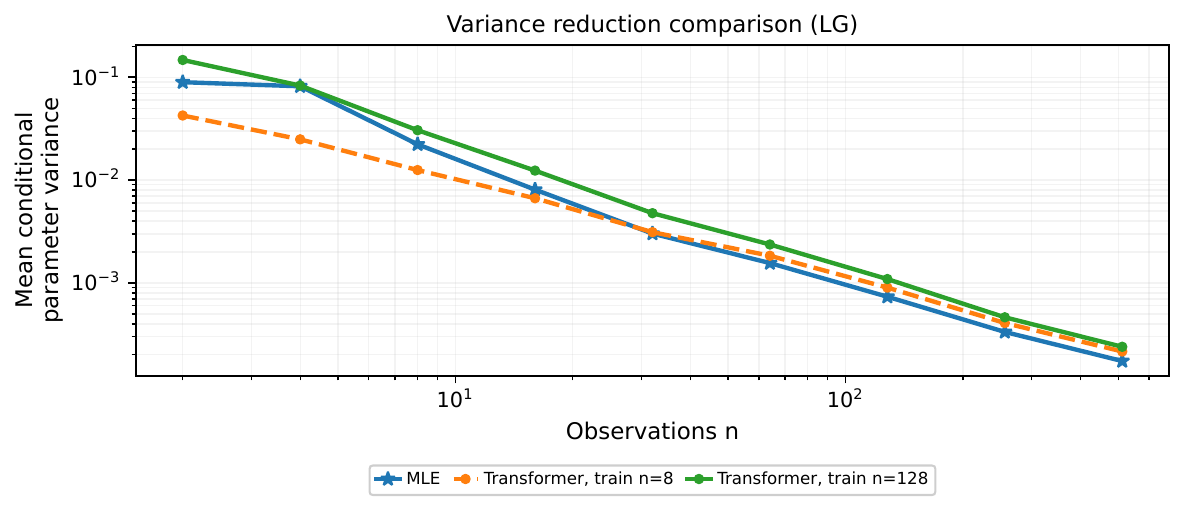}\par\smallskip
 \includegraphics[width=.9\linewidth,trim=0bp 5.5bp 0bp 0bp,clip]{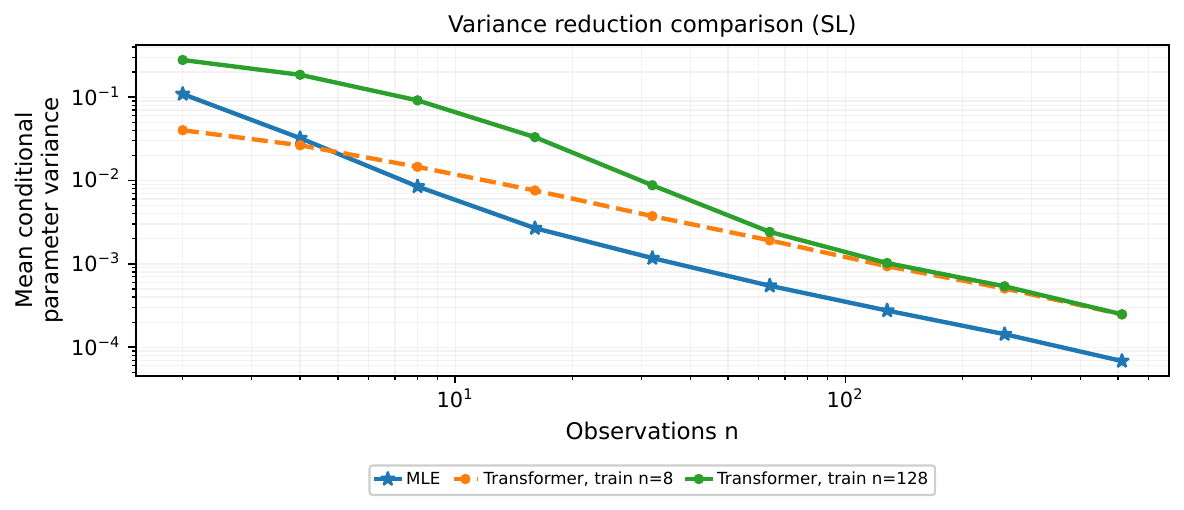}%
 \caption{Conditional parameter-estimate variance for forced-linear (top), logistic growth (middle), and Stuart--Landau (bottom), averaged over five fixed parameter values and normalized parameter coordinates. Each fixed parameter uses 128 independent datasets per $n$ and seed; methods receive identical observations.}
 \label{fig:exp-nonlinear-variance}
\end{figure}
\paragraph{Conditional variance across independent datasets}
We further examine the sampling variability of parameter estimates at fixed true parameters. For each of five interior parameter values, we generate 128 independent datasets at each inference context length by resampling observation times and noise. The MLE and two trained parameter transformers are evaluated on the same datasets. We compute the conditional sample variance of each estimated parameter coordinate and average over the coordinates and five parameter values, using normalized parameter scales and the mean token output for the transformers.
Figure~\ref{fig:exp-nonlinear-variance} shows the forced-linear conditional variance discussed in Section~\ref{sec:experiments} and repeats the fixed-parameter conditional variance reduction experiment for logistic growth and Stuart--Landau.
 \[
 \widehat V_n=\frac{1}{5p}\sum_{a=1}^{5}\sum_{j=1}^{p}
 \frac{1}{127}\sum_{r=1}^{128}
 \left(\widehat z_{a,n,j}^{(r)}-\overline z_{a,n,j}\right)^2.
\]
Variances are calculated separately at each fixed parameter before averaging, so variation between different true parameters is excluded.

\FloatBarrier
\subsection{Query-time dependence of joint parameter estimation}
\label{app:joint-query-sweep}
\begin{figure}[!htbp]
 \centering
 \includegraphics[width=\linewidth,trim=0bp 5.5bp 0bp 0bp,clip]{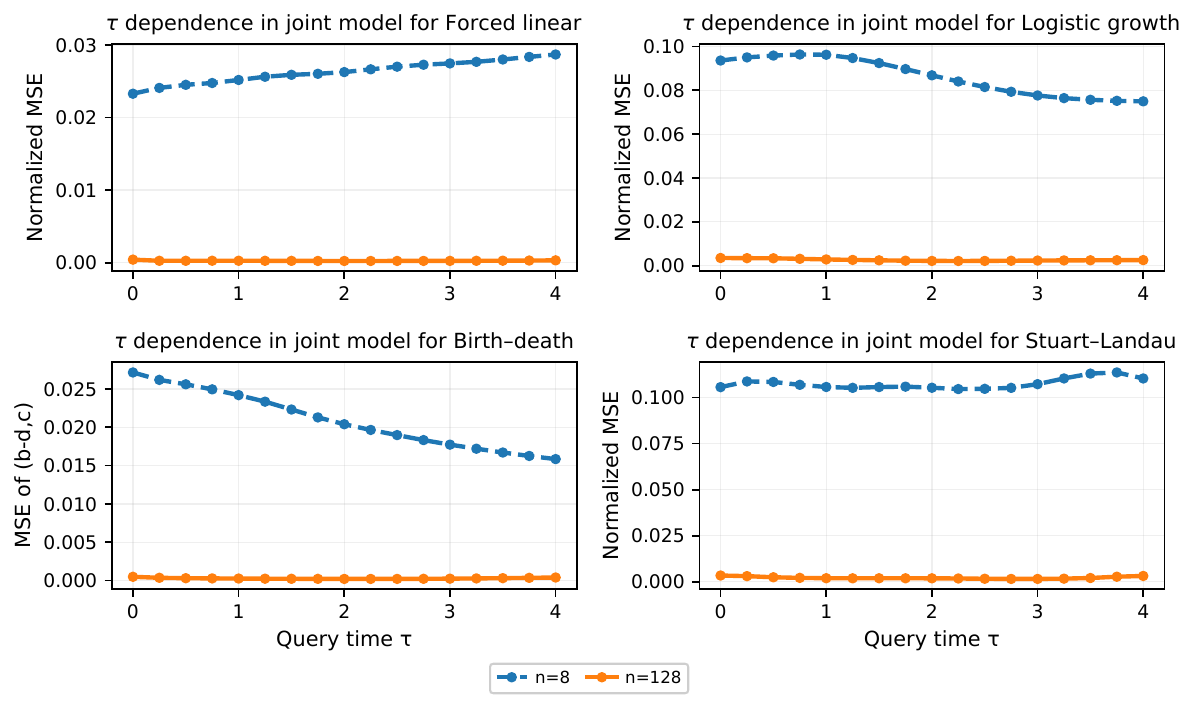}%
 \caption{Joint-model parameter MSE versus query time: forced linear (top left), logistic growth (top right), birth--death (bottom left), and Stuart--Landau (bottom right). Observation datasets remain fixed throughout each query sweep. Curves average both seeds at matched training and inference contexts $n=8,128$.}
 \label{fig:joint-query-sweep}
\end{figure}

We test whether the joint model's parameter-estimation accuracy depends on the broadcast query time (Figure~\ref{fig:joint-query-sweep}). For each family, we draw eight parameters uniformly from the original parameter box and generate 16 independent noisy observation datasets per parameter. Each dataset is held fixed while the query time varies over 17 equally spaced values in $[0,4]$. The models use $M=2^{19}$ and training contexts $n=8$ or $128$, with inference context equal to training context. We average normalized all-token parameter MSE over parameter coordinates, datasets, fixed parameters, and the two model seeds. The same diagnostic datasets are used across seeds. Here birth--death is evaluated on its individual parameter coordinates, rather than its identifiable combinations. This diagnostic measures query-dependent average accuracy, not equality of the parameter outputs across query times.

\FloatBarrier
\subsection{Initial-state recovery by querying at zero}
\label{app:initial-state-query}
\begin{figure}[!htbp]
 \centering
 \includegraphics[width=.7\linewidth,trim=0bp 5.6bp 0bp 0bp,clip]{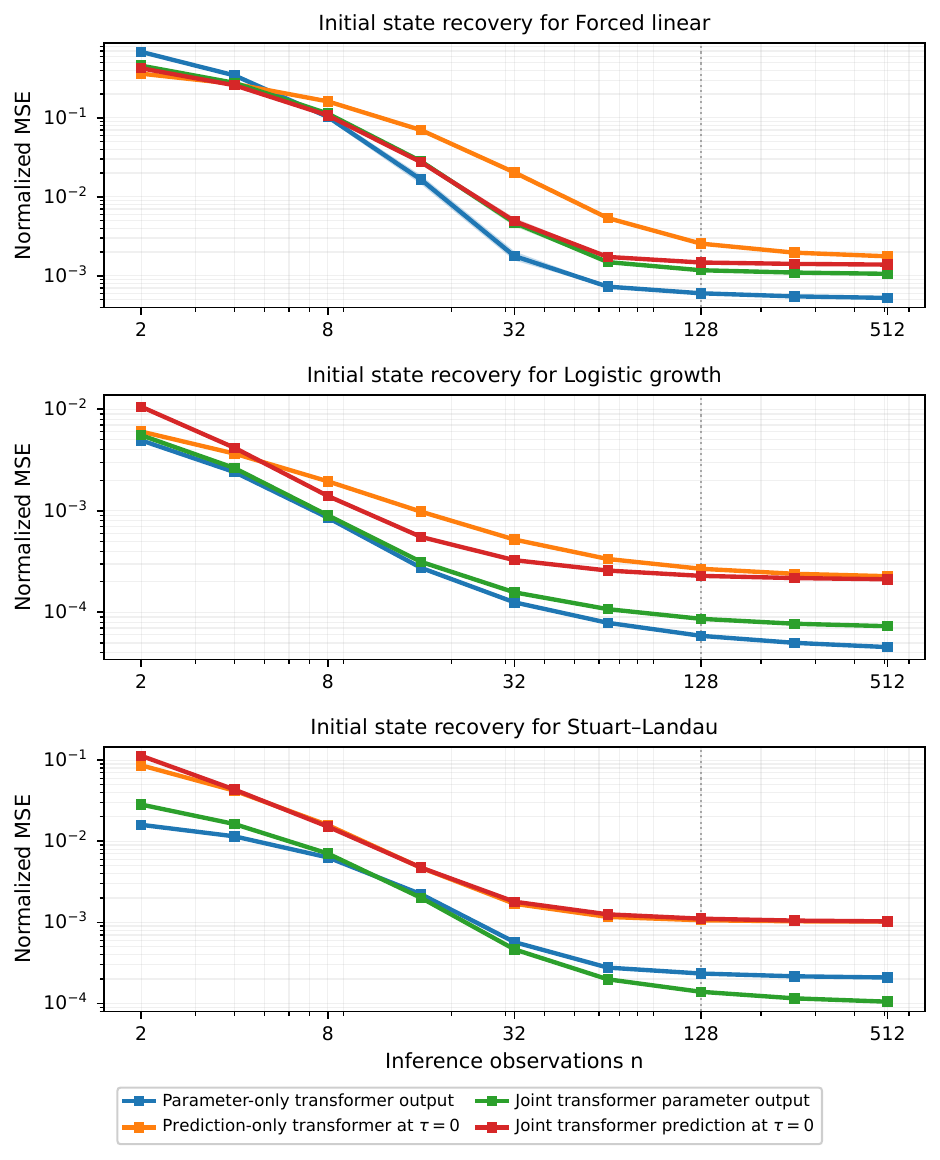}%
 \caption{Initial-state recovery: separate and joint parameter outputs versus direct predictions at $\tau=0$. Errors use normalized state units. Dotted lines mark training context $n=128$.}
 \label{fig:initial-state-forced-linear}
\end{figure}

In Figure~\ref{fig:initial-state-forced-linear}, we compare four estimates of the initial state: the initial-condition coordinates of the parameter transformer; the prediction transformer's output at $\tau=0$; and the joint transformer's initial-condition coordinates and trajectory output, both evaluated with query $\tau=0$. All models use $M=2^{19}$ and training context $n=128$. At each inference context, the methods receive the same 8,192 noisy observation datasets per seed. Errors are averaged over output tokens and state coordinates, using the family state scales for all four methods, including the decoded initial-condition estimates. Curves average the two seed-specific risks equally; bands are pointwise 95\% bootstrap confidence intervals conditional on the trained models. We include forced linear, logistic growth, and Stuart--Landau; birth--death is excluded because its initial condition is fixed.

\FloatBarrier
\subsection{Descriptive pretraining-risk scaling fits}
\label{app:pretraining-risk-fits}
\begin{figure}[htbp]
 \centering
 \includegraphics[width=\linewidth,trim=0bp 5.5bp 0bp 0bp,clip]{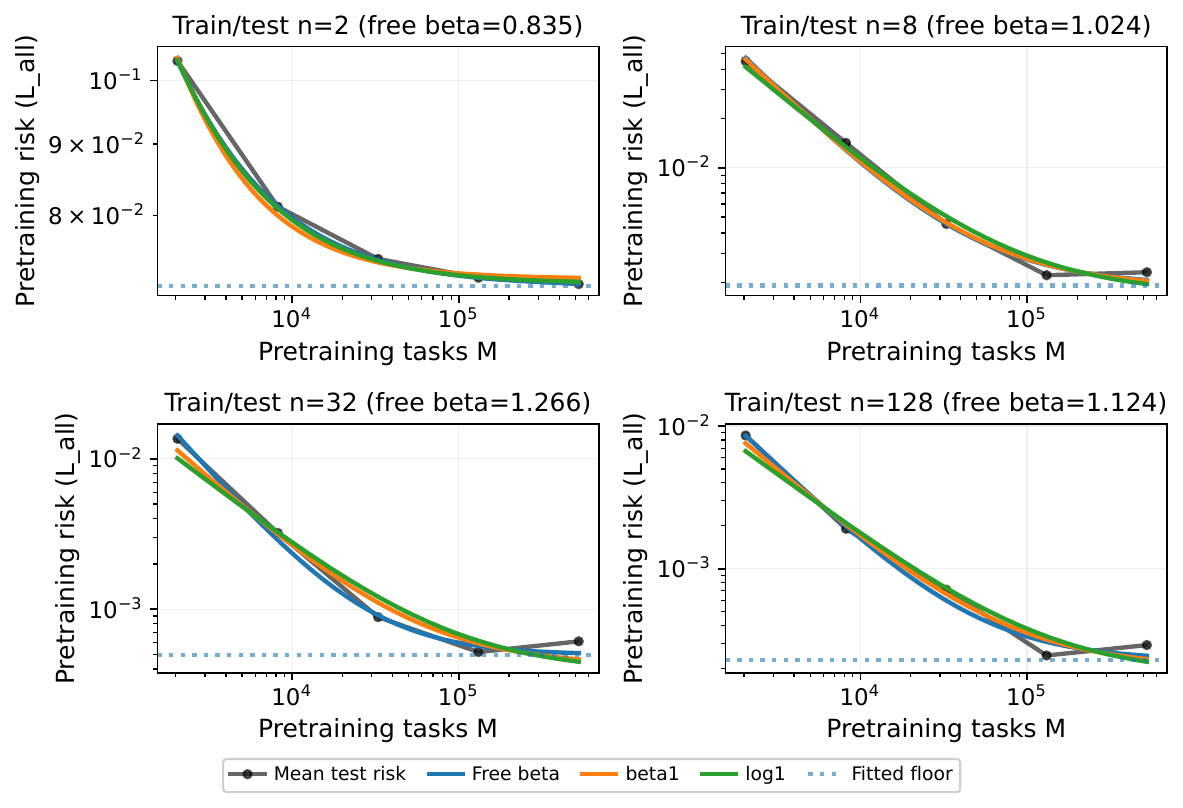}%
 \caption{Forced-linear held-out pretraining risk versus training-bank size $M$, with training and evaluation context fixed at $n=2,8,32,128$. Points average both seeds' normalized all-token parameter MSE. Fits minimize squared log-risk residuals across both seeds using $F+A(M/2^{11})^{-\beta}$, yielding $\beta=0.835,1.024,1.266,1.124$, respectively. Fixed-exponent and logarithm-corrected curves are shown for comparison; dotted lines indicate fitted floors. These are descriptive fits to total risk, not a verification of an exact $1/M$ rate.}
 \label{fig:pretraining-risk-fits}
\end{figure}

Figure~\ref{fig:pretraining-risk-fits} shows descriptive fits of the forced-linear held-out pretraining risk as a function of the number of pretraining tasks $M$.

\FloatBarrier

\end{document}